\documentclass[pmlr]{jmlr}

\RequirePackage{graphicx}
\usepackage{booktabs}
\usepackage{multirow}
\usepackage{minted}
\usepackage{xcolor}

\definecolor{skillbg}{HTML}{F7F7F5}

\setminted{
  breaklines=true,
  breakanywhere=true,
  fontsize=\scriptsize,
  xleftmargin=0pt,
  xrightmargin=0pt,
  framesep=0pt,
}
\makeatletter
\renewcommand{\minted@error}[1]{\PackageWarning{minted}{#1}}
\makeatother

\usepackage{enumitem}
\RestyleAlgo{ruled}

\SetKwComment{tcp}{// }{}
\makeatletter
\def\set@curr@file#1{\def\@curr@file{#1}} 
\makeatother
\usepackage[load-configurations=version-1]{siunitx} 

\theorembodyfont{\upshape}
\theoremheaderfont{\scshape}
\theorempostheader{:}
\theoremsep{\newline}

\jmlrproceedings{PMLR}{Proceedings of Machine Learning Research}
\jmlrvolume{340}
\jmlryear{2026}
\jmlrworkshop{Machine Learning for Healthcare}

\title[KMGen]{KMGen: A Skill-based Approach for Synthetic Individual Patient Data Generation}

\author{\Name{Jalen Jiang}
        \Email{jalenj4@illinois.edu}\\
       \addr Siebel School of Computing and Data Science\\
       University of Illinois Urbana-Champaign\\
       \AND
       \Name{Chufan Gao}
       \Email{chufan2@illinois.edu}\\
       \addr Siebel School of Computing and Data Science\\
       University of Illinois Urbana-Champaign\\
       \AND
       \Name{Ethan Rasmussen}
       \Email{ethanmr3@illinois.edu}\\
       \addr Siebel School of Computing and Data Science\\
       University of Illinois Urbana-Champaign\\
       \AND
       \Name{Stephen Z. Xie}
       \Email{xie.stephen@mayo.edu}\\
       \addr Mayo Clinic Alix School of Medicine\\
       \AND
       \Name{Jimeng Sun}
       \Email{jimeng@illinois.edu}\\
       \addr Siebel School of Computing and Data Science\\
       University of Illinois Urbana-Champaign
       }

\makeatletter
\long\def\@makecaption#1#2{%
   \vskip 0pt
   \setbox\@tempboxa\hbox{#1: #2}%
   \ifdim \wd\@tempboxa >\hsize
       \begin{list}{#1:}{%
       \settowidth{\labelwidth}{#1:}
       \setlength{\leftmargin}{\labelwidth}
       \addtolength{\leftmargin}{\labelsep}
       \setlength{\topsep}{0pt}
       \setlength{\partopsep}{0pt}
       \setlength{\itemsep}{0pt}
       \setlength{\parsep}{0pt}
        }\item #2 \end{list}\par
     \else
       \hbox to\hsize{\hfil\box\@tempboxa\hfil}
   \fi
   \def\@tempa{table}%
   \ifx\@captype\@tempa\vskip 8pt\fi}
\makeatother

\setlist{topsep=3pt, itemsep=1.5pt, parsep=1pt}

\begin{document}
\maketitle

\begin{abstract}
Individual patient data (IPD) from clinical trials is the substrate for survival modeling, meta-analysis, and safety research, yet IPD is rarely released. Prior work has addressed only half of this gap: reconstructing Kaplan--Meier (KM) curves from published plots --- typically requiring manual digitization or human-in-the-loop correction --- while offering no mechanism for generating the adverse-event (AE) streams that constitute the other half of a patient record. We introduce \textbf{KMGen}, the first end-to-end framework that (i)~fully automates KM curve extraction at accuracy competitive with human-guided tools, and (ii)~generates synthetic per-patient AE trajectories from public trial registry records. The extraction stage is a fully automated agentic pipeline --- an agent generates code to extract each step in the km curve --- achieving a mean Integrated Absolute Error (IAE) of \textbf{0.0151} on a 32-plot benchmark spanning clean, edge-case, and adversarial conditions. The IPD generation stage decouples \emph{patient archetype extraction} from \emph{statistical sampling}: an LLM distills the trial record into arm-specific statistics, adverse events, patient demographics, and risk multipliers. A mechanistic sampler generates patient events via clinical archetypes, bootstrap rank-correlation coupling to the empirical KM curve (preserving the marginal survival distribution exactly), and cycle-based AE scheduling with an induction/maintenance split. Across three held-out oncology trials spanning an order of magnitude in cohort size and 30 independent regenerations per trial, KMGen achieves mean integrated KM absolute difference $\Delta_{\text{KM}}\,{\leq}\,0.051$, sex/ECOG JSD ${\leq}\,0.013$ on 5 of 6 demographic slots, and recovers ${\geq}\,71\%$ of the top-15 AEs by exact MedDRA term under a single fixed parameter set. The pipeline is interpretable end-to-end and released as open source at \url{https://github.com/chufangao/kmgen}.

\end{abstract}

\section{Introduction}
Individual patient data (IPD) represents the gold standard for clinical research, enabling precise modeling of time-to-event outcomes, subgroup identification, and rigorous validation of statistical assumptions. However, access to IPD from Randomized Controlled Trials (RCTs) is often restricted due to privacy regulations, proprietary interests, and data governance constraints. Consequently, secondary analyses frequently rely on aggregated summary statistics, such as hazard ratios (HR) and median survival times. These summary statistics impose limitations on downstream applications that require granular event timing or censoring patterns, particularly in the development of machine learning models and meta-analyses.

Recent methodological advances have sought to mitigate these limitations by reconstructing IPD from published survival data. \citet{guyot2012} introduced an iterative numerical algorithm to invert Kaplan--Meier \cite{kaplan1958} equations, allowing for the recovery of sufficient statistics from published curves. Subsequent developments, such as IPDfromKM \cite{liu2021}, enhanced the stability of this reconstruction process through modified iterative algorithms and improved user interfaces. Most recently, KM-GPT \cite{zhao2025km} demonstrated the application of multi-modal LLMs to automate the extraction process, significantly reducing manual digitization requirements. While these reconstruction methods have substantially advanced the accessibility of historical trial data, their utility is inherently bounded by the existence of published figures.

To address these complementary constraints, we introduce \textbf{KMGen}, a unified framework for both the extraction and generation of synthetic IPD from clinical trial reports and KM curves. KMGen operates through a dual-purpose architecture that addresses the full spectrum of IPD accessibility needs: (1) an adaptive hybrid pipeline for extracting patient-level survival data from published KM plots, and (2) a semi-deterministic agentic harness for generating high-fidelity synthetic IPD when reconstruction is not feasible. This comprehensive approach builds upon the foundational work of prior reconstruction methods while extending their utility to scenarios where trial data must be simulated for methods development or secondary analyses.

The extraction component of KMGen employs a two-stage pipeline that combines computer vision with multimodal LLM reasoning. An LLM agent, using an adaptive markdown skill, analyzes the input image to select from a technique toolbox (color-based extraction, spatial cluster separation, tick-mark bounding box detection), while deterministic computer vision code handles pixel-level precision. This architecture addresses three core challenges in KM plot extraction: (1) axis calibration despite matplotlib rendering artifacts, (2) curve separation under visual anti-patterns including overlapping curves and black-and-white figures, and (3) validation against ground-truth synthetic benchmarks.

The generation component shifts the problem from reconstruction to creation. End-to-end LLM samplers can hallucinate clinically implausible event sequences and tend toward mode collapse on the long tail of rare \cite{davidson2026reasoning}. KMGen instead splits the work: an LLM agent maps the trial record into a structured, MedDRA-compliant configuration, and a deterministic sampler then realizes per-patient event streams from that configuration --- using bootstrap rank-correlation coupling to the empirical KM curve (preserving the marginal survival distribution by construction), patient archetypes with per-organ-system risk multipliers, and cycle-based AE scheduling with an induction--maintenance phase split. Every clinical prior is an explicit, citable parameter rather than a latent weight, which makes the resulting synthetic IPD auditable for downstream methods development and safety-monitoring work in settings where access to real IPD is restricted.

This paper contributes to the ecosystem of IPD tools by unifying extraction and generation capabilities within a single framework. Our primary contributions are:
\begin{enumerate}[leftmargin=*]
\item The development of KMGen, a dual-purpose framework that combines adaptive hybrid extraction with deterministic synthetic generation for comprehensive IPD accessibility.
\item An evaluation of extraction performance on a 32-plot synthetic benchmark spanning clean plots, edge cases, and adversarial stress tests, demonstrating a mean IAE of 0.0151---comparable to methods that require much more human guidance \cite{zhao2025km}.
\item An end-to-end evaluation of the generation pipeline on three real oncology trials with held-out IPD, reporting mean $\pm$ std over 30 independent regenerations: mean integrated KM absolute difference $\Delta_{\text{KM}} \leq 0.051$, top-15 AE overlap $\geq 71\%$, and faithful early-phase AE timing under a single fixed parameter set.
\item The release of an open-source implementation to facilitate reproducibility and further development within the research community.
\end{enumerate}

The remainder of this paper is organized as follows: Section~\ref{sec:km-methodology} presents the KM plot extraction pipeline and our synthetic benchmark setup. Section~\ref{sec:generation-methodology} details the mechanistic synthetic IPD generation methodology. Section~\ref{sec:km-results} evaluates the KM plot extraction and the generation pipeline against ground-truth IPD from three complex oncology trials, highlighting fidelity metrics for survival and safety profiles. Finally, Section~\ref{sec:paper-discussion} discusses the implications for healthcare machine learning and future directions in synthetic clinical data generation.

\subsection{Generalizable Insights about Machine Learning in the Context of Healthcare}
\label{sec:generalizable-insights}

The central design principle in KMGen is to assign interpretive tasks to the multimodal LLM and computational tasks to deterministic code, rather than letting either component perform both. The LLM is restricted to strategy selection and to populating structured fields from heterogeneous source material. This separation confines the LLM to contexts where approximation is tolerable, while providing exactness and reproducibility where the task requires it. This insight can be generalized to digitization and IPD generation tasks across clinical fields.

\begin{figure}[ht]
    \centering
    \includegraphics[width=\linewidth]{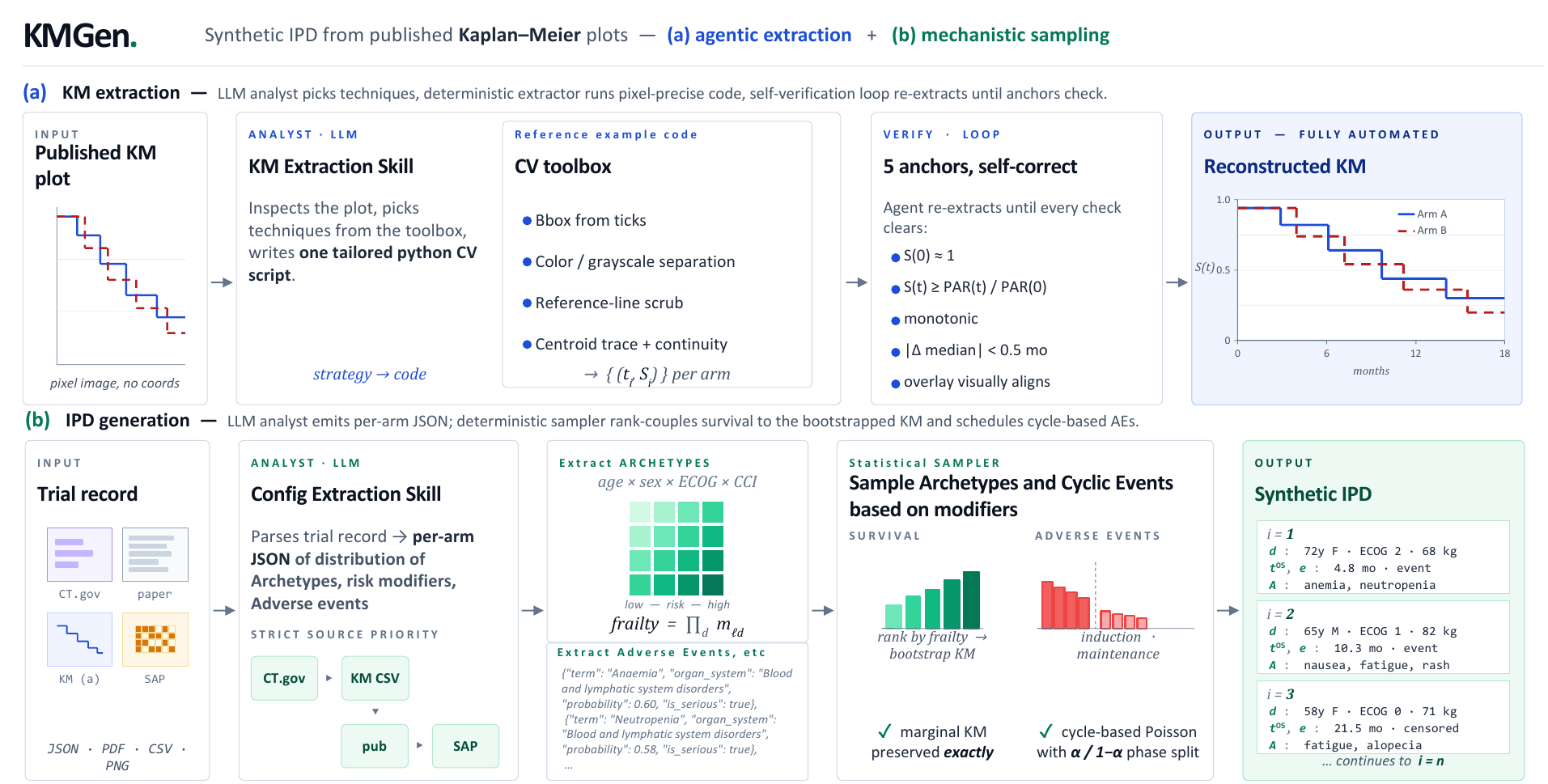}
    \caption{ KMGen pipeline. (a) Fully automated KM extraction  (b) Individual Patient Data generation from extracted KM plot, ClinicalTrial.gov data, publication, and statistical analysis plan.}
    \label{fig:hero}
\end{figure}
\section{Methods}

\subsection{KM Plot Extraction Pipeline}
\label{sec:km-methodology}
The extraction pipeline is a deliberate split between an \emph{analyst}
(multimodal LLM) and an \emph{extractor} (deterministic Python). The
analyst inspects the image, identifies challenges (curve count,
color scheme, reference lines, confidence-interval bands, patients-at-risk
table, annotation density), and selects applicable techniques from a toolbox.
The extractor then executes pixel-precise code tailored to that plot. For small residual artifacts, the agent may hardcode localized corrections over clean neighbors.
This plays to each tool's strengths: LLMs read context, strategize, and write code;
code counts pixels and extracts finer coordinates.

The technique toolbox consists of the following suggested methods:
\begin{itemize}[leftmargin=*]
    \item \textbf{Bounding-box detection from tick marks.} The outermost
        x- and y-axis tick marks define the plot rectangle. Tick spacing
        is verified by linear regression with $R^2 > 0.999$; frame
        borders (full-axis lines) are disambiguated from ticks (short
        perpendicular segments) by length.
    \item \textbf{Color separation.} HSL hue matching for hue-distinct arms,
  RGB thresholds when hues are close in color space, and grayscale intensity
  thresholds for black-and-white plots where arms appear at distinct luminance
  levels.
    \item \textbf{Reference-line scrubbing.} Any row or column whose
        dark-pixel coverage exceeds 30\% of the plot width or height is a
        reference line, not a curve; those rows are blanked (with a
        $\pm 2$~px margin) before tracing.
    \item \textbf{Curve tracing with continuity.} Each column selects the
        pixel cluster closest to the previous column's position; the
        centroid of the cluster (not the topmost pixel) becomes the
        reported point. Monotonicity is enforced post-trace.
    \item \textbf{Dashed-line gap bridging.} For dashed arms, gaps between
        dashes are carried forward up to a dash-period threshold, so that
        the tracer does not treat a dash gap as the curve terminating.
    \item \textbf{Terminal-drop detection.} At the last populated column,
        if the color mask spans more than 10~px vertically, the bottom
        of the span is appended as a final step (the terminal drop in the
        survival curve).
    \item \textbf{Pre-extraction cleanup.} In-plot text, legends, and
        patients-at-risk tables are masked as rectangular exclusion
        zones before tracing.
\end{itemize}

The extractor emits coordinates at configurable granularity (daily, monthly, or whatever the trial specifies. Final output is a
JSON document containing the detected bounding box, axis ranges, per-arm
label/color/coordinate list, and (when readable) a patients-at-risk
table.

Verification runs on five independent checks: (i) $S(0) \approx 1.0$ at the leftmost populated column (if not, the bounding box is wrong); (ii) $S(t) \geq \mathrm{PAR}(t)/\mathrm{PAR}(0)$ when the PAR table is available; (iii) zero monotonicity violations; (iv) any on-plot median annotation matches the extracted median to within 0.5~months; and (v) visual
  annotation overlays are rendered and fed back to the agent for inspection — the agent self-corrects and re-extracts until the overlays confirm the trace aligns with the source curves.

\subsubsection{Synthetic KM Plot Evaluation}
\label{sec:km-benchmark}
We evaluate the extraction component of KMGen on a 32-plot synthetic
benchmark spanning clean standard plots, plausible edge cases drawn from
published medical figures, and stress tests designed to probe specific
failure modes. The pipeline is adaptive: each plot is extracted by an LLM agent that writes tailored CV code, inspects the result against visual overlays and published anchors, and self-corrects when drift or anomalies are detected before committing final output.

Ground truth is generated by Weibull-sampling event times with random
censoring and rendering each curve as a
matplotlib step function at 4-decimal precision. The 32-plot benchmark
breaks into four categories:

\begin{itemize}[leftmargin=*]
    \item \textbf{Standard synthetic (5 plots.)} Two-arm plots with
        $x_{\max} \in \{12, 36, 36, 18, 12\}$~months, clean color
        separation, no in-plot decoration. These set the pipeline's
        floor accuracy.
    \item \textbf{Edge cases (7 plots.)} Plausible variations drawn
        from published medical figures: high-survival 144-month
        follow-up, cumulative-incidence (rising) curves, four-arm
        comparisons, confidence-interval shading, near-flat survival,
        small-dense step structure, and multi-panel layouts.
    \item \textbf{Single-degradation stress tests (12 plots.)} Each plot
        applies one adversarial condition to a standard 2-arm base:
        Gaussian blur, JPEG compression, black-and-white, horizontal or
        vertical stretch, tiny resolution, legend overlap, gridlines,
        three similarly-hued arms, annotation-heavy, clean multi-panel,
        and an 8-arm DIEP-like reconstruction plot.
    \item \textbf{Combination stress tests (8 plots.)} Stack multiple
        adversarial conditions: flat+overlap, B\&W+overlap,
        stretched+dense, tiny+B\&W, tiny+blurry, 4-arm+tiny,
        JPEG+blurry+dark, and grid+annotation+B\&W.
\end{itemize}

Axis ranges vary by plot ($x_{\max}$ between 12 and 144 months), so all
reported errors are normalised to $[0, 1]$ before integration.

\subsubsection{KM Plot Extraction Metrics}
\label{sec:km-metrics}

All metrics are computed by
comparing each extracted arm against the corresponding truth arm. The
primary metric is the \textbf{Integrated Absolute Error (IAE)}:

\begin{equation}
\mathrm{IAE} = \int_0^1 \bigl|\, S_{\text{ext}}(\tau) - S_{\text{truth}}(\tau) \,\bigr|\, d\tau,
\qquad \tau = t / x_{\max}
\label{eq:iae}
\end{equation}

where both step functions are interpolated onto the union of their
x-grids using left-step interpolation $S(t) = S(t_k)$ for
$t_k \leq t < t_{k+1}$. The integral
is computed by the trapezoidal rule on the merged grid. IAE lies in
$[0, 1]$; $\mathrm{IAE} = 0$ is a perfect match.

We report three secondary metrics:
\begin{itemize}[leftmargin=*]
    \item \textbf{Score} $= \max(0, 1 - \overline{\mathrm{IAE}})$, aggregated across arms.
    \item \textbf{Median AE}: the median of
        $|S_{\text{ext}}(t_i) - S_{\text{truth}}(t_i)|$ evaluated at 100
        evenly-spaced $t_i \in [0, x_{\max}]$
    \item \textbf{Median OS error}:
        $|\mathrm{med}_{\text{ext}} - \mathrm{med}_{\text{truth}}|$,
        where $\mathrm{med}$ is the first $t$ with $S(t) \leq 0.5$. Undefined when the curve
        never crosses $0.5$.
\end{itemize}

\subsection{IPD Generation Methodology}
\label{sec:generation-methodology}

Our pipeline converts a publicly available clinical trial registry record into a synthetic individual-patient-data (IPD) cohort in three stages. First, an LLM-driven \emph{extraction} step distills the trial registry record into a structured configuration. Second, a \emph{sampling} step generates per-patient event streams from that configuration. Third, an \emph{evaluation} step compares both the configuration and the synthetic cohort against real IPD held out from the extraction agent. We describe each stage below; the full implementation details appear in Appendix~\ref{app:implementation}.

\subsubsection{Trial Configuration Schema}
\label{sec:schema}

A trial configuration (JSON file) is emitted per arm and contains the following fields. Trial identity: trial ID, arm name, title, condition, treatment description. Enrollment: per-arm count $n$. Treatment timing: cycle length (days), induction cycle count $K_{\mathrm{ind}}$, induction AE fraction $\alpha$. Demographics: age mean, SD, min, max; sex ratio; ECOG distribution; weight/height mean and SD; race and region distributions. Survival: pointer to empirical KM CSV. Adverse events: list of records, each with MedDRA Preferred Term, SOC, incidence probability $p_a$ (pooled across ClinicalTrials.gov serious and other events), and seriousness flag (true when grade$\,{\geq}\,3$ events constitute ${\geq}50\%$ of any-grade occurrences).

Override dictionaries allow the extraction agent to adjust default risk multipliers when the drug class/paper warrants it. Overrides are serialized only when non-default.
An example trial config is shown in Appendix~\ref{app:skill_2}.

\subsubsection{Patient Archetype Extraction}

Given a trial registry record (e.g. NCT03041311), an LLM agent populates a per-arm configuration that the downstream sampler consumes. The agent follows a strict four-level source-priority hierarchy:
\begin{enumerate}[leftmargin=*]
    \item \textbf{ClinicalTrials.gov JSON} (primary): arm labels, baseline demographics, and adverse-event (AE) incidences from the structured results record. AE organ-system labels use the verbatim MedDRA System Organ Class (SOC) field.
    \item \textbf{Reconstructed overall-survival Kaplan--Meier CSV (from the first section)}: empirical $(t, e)$ pairs; no parametric fitting is applied.
    \item \textbf{Primary publication and appendix}: consulted only for fields that ClinicalTrials.gov leaves unresolved (finer ECOG bucketing, low-frequency AEs below the 5\% reporting threshold, subgroup survival).
    \item \textbf{Statistical analysis plan (SAP) and protocol}: last resort, for cycle length, assessment schedule, and class-effect AEs at the ${<}1\%$ level documented in drug labels but absent from ClinicalTrials.gov.
\end{enumerate}
The hierarchy is strict because each level introduces a noisier source of information than the one above: ClinicalTrials.gov is machine-parseable and closest to the sponsor's submission, while publications and SAPs involve PDFs and PNGs, respectively.

\paragraph{Channel separation.} Only the \emph{survival} channel originates from a plot image: the empirical $(t,e)$ pairs at level~2 are traced from the published figure by the extraction pipeline of Section~\ref{sec:km-methodology}. Demographics, MedDRA terms, and AE incidences enter at levels~1, 3, and 4 as structured or tabular values and are never read off a curve, so they do not inherit curve-tracing error. Extraction error therefore propagates into exactly one downstream channel; Section~\ref{sec:propagation} isolates and quantifies that contribution.

\subsubsection{Statistically Sampling for Synthetic IPD}

Given a configuration, the sampler emits $n$ synthetic patients, each a chronologically ordered event stream (treatment start, AE events, terminal death, or censoring). Generation proceeds in three stages (full pseudocode in Algorithm~\ref{alg:generation}, Appendix~\ref{app:algorithm}); each is detailed in the subsubsections that follow.

\paragraph{Patient Archetypes}
\label{sec:archetypes}

The configuration is expanded into a discrete set of \emph{archetypes}, each a cell in the cross-product of four risk dimensions: \textbf{Age}: $\{\text{younger}, \text{older}\}$, split at $\lfloor \bar{a} \rfloor$ (the reported mean age), \textbf{Sex}: $\{\text{male}, \text{female}\}$, weighted by the reported sex ratio, \textbf{ECOG}~\cite{oken1982}: collapsed into $\{\text{good}, \text{poor}\}$ at ECOG $\leq 1$ vs.\ $\geq 2$, \textbf{Comorbidity} (CCI~$\geq 2$~\cite{charlson1987}): $\{\text{low}, \text{high}\}$.
This yields 8--16 archetypes. For archetype $\ell = (\ell_{\text{age}}, \ell_{\text{sex}}, \ell_{\text{ecog}}, \ell_{\text{com}})$, we normalize the weights $w_\ell = \prod_d p(\ell_d)$ (sums to one) for sampling purposes. Survival and AE risk modifiers compose multiplicatively across factors:
\begin{align}
    m^{\text{OS}}_\ell &= \prod_{d \in \{\text{age},\text{ecog},\text{com}\}} m^{\text{OS}}_{\ell_d}, &
    m^{\text{AE}}_\ell(s) &= \prod_{d \in \{\text{age},\text{sex},\text{ecog},\text{com}\}} m^{\text{AE}}_{\ell_d}(s),
\end{align}
where $m^{\text{AE}}_\ell(s)$ is the risk multiplier for SOC $s$. This multiplicative structure follows the standard proportional-hazards assumption~\cite{cox1972} and keeps per-axis multipliers interpretable.

Default risk multipliers (Table~\ref{tab:risk-multipliers}) are clinical estimates informed by a targeted literature review of geriatric oncology and chemotherapy toxicity studies. The cited studies provide directional evidence and overall effect sizes but not per-SOC relative risks; the per-organ multipliers are therefore calibrated estimates, not direct transcriptions (see Appendix~\ref{app:multipliers} for the detailed evidence base).

\paragraph{Survival Assignment via Rank-Correlation Coupling}
\label{sec:survival}

The synthetic cohort must reproduce the reconstructed Kaplan--Meier~\cite{kaplan1958} curve \emph{exactly} as a marginal distribution while inducing within-cohort risk stratification. Parametric fits discard empirical curve shape; marginal KM sampling followed by independent archetype assignment destroys the frailty--survival correlation.

We resolve this tension with a bootstrap-plus-rank-correlation coupling, drawing on the bootstrap for censored data~\cite{efron1981} and the Iman--Conover method~\cite{iman1982}. Let $\{(t_k, e_k)\}_{k=1}^{K}$ denote the reconstructed empirical KM pairs. We bootstrap $n$ pairs with replacement and sort by time. Each patient $i$ with archetype $\ell(i)$ receives a frailty score:
\begin{equation}
    r_i = -\log\!\bigl(m^{\text{OS}}_{\ell(i)}\bigr) + \varepsilon_i, \qquad \varepsilon_i \sim \mathcal{N}(0, \sigma^2),
\end{equation}
where $\sigma$ controls the coupling strength. We assign the shortest bootstrap time to the highest-risk patient:
\begin{equation}
    (t_i^{\text{OS}}, e_i) = \bigl(t_{(n - \pi(i) + 1)},\; e_{(n - \pi(i) + 1)}\bigr), \quad \pi(i) = \text{rank of } r_i.
\end{equation}
Because the coupling is a permutation of bootstrap draws, the marginal KM distribution is preserved exactly.

\paragraph{Cohort size.} The permutation argument holds at any $n$: no cohort size is required for the marginal to be unbiased, and small cohorts only coarsen the resolution of the realized step function. What does degrade with $n$ is the \emph{conditional} structure --- with fewer patients than archetypes, the rank coupling cannot populate the frailty ordering densely enough for stratified survival to be estimable. We therefore recommend a floor of a few dozen patients per arm (the smallest arm evaluated here is $n_r{=}37$), below which the cohort remains marginally faithful but should not be used for subgroup analysis.

\paragraph{Demographics Sampling}

For each patient $i$ conditional on archetype $\ell(i)$:
\begin{align*}
    \text{age}_i &\sim \text{Uniform}\{a_{\min}^{(\ell)}, \ldots, a_{\max}^{(\ell)}\}, \\
    \text{weight}_i &\sim \mathcal{N}(\mu_w^{(\ell)}, \sigma_w^2) \text{ clipped to } [40, 160]\,\text{kg}, \\
    \text{height}_i &\sim \mathcal{N}(\mu_h^{(\ell)}, \sigma_h^2) \text{ clipped to } [140, 200]\,\text{cm}.
\end{align*}
Per-archetype means are scaled from trial-level means by sex-specific multipliers. Race and region are drawn from the reported categorical distributions. Truncation bounds are physiologically motivated for adult oncology populations; supplementary Figure~\ref{fig:supp-demo} shows full distributions.

\paragraph{Adverse Event Sampling}
\label{sec:ae-sampling}

AEs are generated on a cycle-based time grid to match the visit-driven observation mechanism of real trials. Let $c$ be the cycle length in months, $N_i = \max(1, \lfloor t_i^{\text{OS}} / c \rfloor)$ the number of realized cycles, and $N_i^{\text{ind}} = \min(K_{\mathrm{ind}}, N_i)$ the induction-phase length. A patient-level propensity modifier $\eta_i \sim \text{LogNormal}(0, \sigma_\eta^2)$ captures inter-patient overdispersion~\cite{vaupel1979,hougaard2000}.

For AE $a$ with base cumulative incidence parameter $p_a$, SOC $s_a$, and cycle $k$, the per-cycle Poisson rate is:
\begin{equation}
    \lambda_{i,a,k} = p_a \cdot m^{\text{AE}}_{\ell(i)}(s_a) \cdot \eta_i \cdot \frac{\phi(k)}{N_i^{\text{phase}(k)}},
\end{equation}
where the phase function $\phi(k) = \alpha$ during induction ($k \leq N_i^{\text{ind}}$) and $\phi(k) = 1-\alpha$ during maintenance. The induction fraction $\alpha$ reflects that most treatment-emergent events cluster in early cycles when dose intensity is highest~\cite{crawford2008}. Occurrences $M_{i,a,k} \sim \text{Poisson}(\lambda_{i,a,k})$ are assigned onset times uniformly in the first third of each cycle window, modeling assessment-driven detection.

\paragraph{Dependence structure.} The single scalar $\eta_i$ multiplies \emph{every} rate for patient $i$, so the model is a shared-frailty~\cite{vaupel1979,hougaard2000} construction rather than a fully independent one: a high-propensity patient is more likely to experience many events across all organ systems, which induces positive within-patient correlation among AEs. What the model does not represent is \emph{term-specific} co-occurrence beyond this shared scalar (e.g., the pairing of febrile neutropenia with neutropenia specifically, over and above both patients' general toxicity level). A copula or hierarchical per-SOC frailty layer is the natural extension, but the joint AE statistics needed to fit one are not reported in registry records --- ClinicalTrials.gov publishes marginal incidences only --- so we treat the shared scalar as a deliberate identifiability-driven tradeoff.

\subsubsection{Synthetic IPD Evaluation}
\label{sec:ipd_evaluation}

The pipeline is subject to two types of errors --- \emph{extraction error} (does the configuration capture the published trial?) and \emph{generative error} (does the sampler realize the configuration faithfully?) --- and we evaluate each in isolation.

\paragraph{Stage A: Extraction quality.} The extracted configuration is compared directly against parameters computed from real IPD that the extraction agent never sees. Scalar demographics are scored by relative error; distributions by Jensen--Shannon divergence (JSD). AE incidence is scored by mean absolute error (MAE), root mean square error (RMSE), JSD, and term-set overlap counts.

\paragraph{Stage B: End-to-end fidelity.} The synthetic cohort is compared against real IPD across six axes: (1)~\emph{demographics} (JSD for discrete variables, Kolmogorov--Smirnov (KS) for continuous); (2)~\emph{overall survival} (median OS error, KS, Mann--Whitney $U$, and integrated absolute difference $\Delta_{\text{KM}} = t_{\max}^{-1}\!\int_0^{t_{\max}}\!|S_{\text{real}}(t) - S_{\text{synth}}(t)|\,dt$); (3)~\emph{AE frequency} (JSD, cosine similarity, top-15 overlap); (4)~\emph{AE timing} (onset KS); (5)~\emph{AE burden per patient} (burden KS); and (6)~\emph{organ-system distribution} (SOC-level JSD and cosine similarity). JSD is preferred over KL divergence for its symmetry, boundedness, and tolerance of non-overlapping support; $\Delta_{\text{KM}}$ avoids the proportional-hazards assumption required by log-rank or Cox methods.

\section{Results}

\label{sec:km-results}
\subsection{KM Extraction Results}

Table~\ref{tab:km-aggregate} summarizes IAE by plot category;
Table~\ref{tab:km-perplot} (Appendix~\ref{app:km-gallery}) lists every
plot sorted best-to-worst. The pipeline achieves mean IAE $0.0151$ across the benchmark, with a heavy-tailed distribution: median IAE is $0.0080$ (about half the mean), and 22 of 32 plots score below $0.015$. The tail is driven almost entirely by combination stress tests (mean IAE $0.0283$) and a handful of outliers, rather than by broad under-performance.

\begin{table}[htb]
\centering
\caption{Extraction accuracy by plot category. The synthetic benchmark (top block, 32 plots) measures IAE against programmatically-rendered ground truth. The real-trial block compares the extracted published KM curve against the empirical KM curve reconstructed from sponsor-delivered IPD for three oncology trials (one arm per trial). IAE is the Integrated Absolute Error in $[0,1]$; lower is better.}
\label{tab:km-aggregate}
\resizebox{\linewidth}{!}{%
\begin{tabular}{lrrrrrr}
\toprule
Category & $n$ & Mean IAE & Median IAE & Min & Max & Std \\
\midrule
\multicolumn{7}{l}{\textit{Synthetic benchmark (ground truth: rendered step function)}} \\
Standard & 5 & 0.0051 & 0.0051 & 0.0033 & 0.0068 & 0.0015 \\
Edge & 7 & 0.0089 & 0.0070 & 0.0012 & 0.0237 & 0.0080 \\
Single stress & 12 & 0.0140 & 0.0087 & 0.0051 & 0.0540 & 0.0136 \\
Combo stress & 8 & 0.0283 & 0.0234 & 0.0012 & 0.0744 & 0.0231 \\
\textbf{Synthetic (all)} & \textbf{32} & \textbf{0.0151} & \textbf{0.0080} & \textbf{0.0012} & \textbf{0.0744} & \textbf{0.0164} \\
\midrule
\multicolumn{7}{l}{\textit{Real trials (ground truth: KM reconstructed from sponsor IPD)}} \\
NCT03041311 (Placebo, $n{=}53$)         &  1 & 0.0309 & 0.0309 & 0.0309 & 0.0309 & --- \\
NCT02499770 (E/P, $n{=}37$)             &  1 & 0.0345 & 0.0345 & 0.0345 & 0.0345 & --- \\
NCT00844649 (Gemcitabine, $n{=}430$)    &  1 & 0.0133 & 0.0133 & 0.0133 & 0.0133 & --- \\
\textbf{Real trials (all)} & \textbf{3} & \textbf{0.0262} & \textbf{0.0309} & \textbf{0.0133} & \textbf{0.0345} & --- \\
\bottomrule
\end{tabular}%
}
\end{table}

\begin{figure}[ht]
\centering
\includegraphics[width=\linewidth]{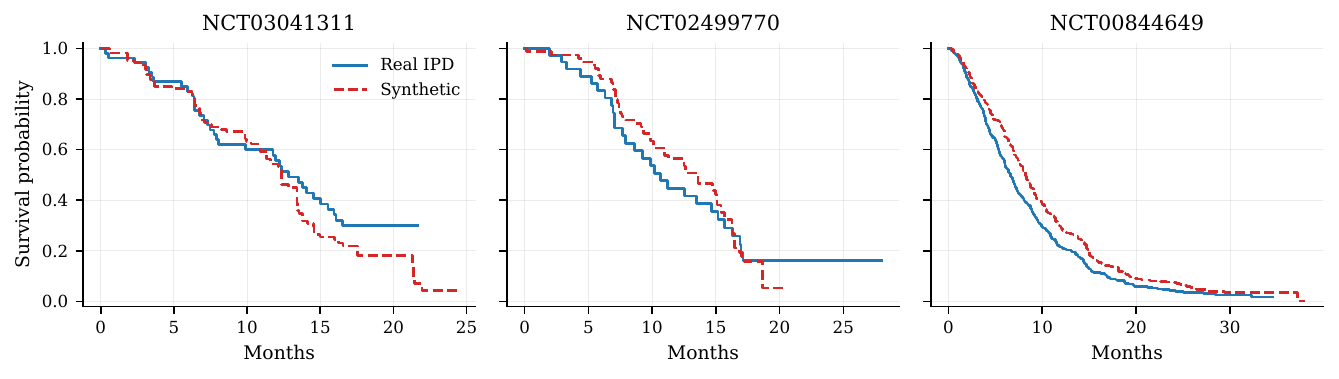}
\caption{Real (solid) vs.\ synthetic (dashed) KM curves. Mean $\Delta_{\text{KM}} \leq 0.051$ on all trials; Mann--Whitney $U$ fails to reject the null on both smaller trials (mean $p \geq 0.6$) but rejects on NCT00844649 in 29 of 30 seeds, driven by a 1.3-month median shift at $n_r{=}430$.}
\label{fig:km}
\end{figure}

To probe whether the synthetic numbers transfer to real figures, we also
evaluate the pipeline on the three oncology trials for which we have
sponsor-delivered individual patient data (IPD):
NCT03041311 (trilaciclib, SCLC), NCT02499770 (T/E/P combination), and
NCT00844649 (nab-paclitaxel + gemcitabine, pancreatic). For each trial
we extract the overall-survival curve from the published KM figure and
compare against the empirical KM reconstructed from the real IPD arm
that matches the extracted arm label, using the same IAE definition
(Eq.~\ref{eq:iae}). Mean IAE is $0.0262$ across the three arms, with
the large-cohort Gemcitabine arm ($n{=}430$) achieving $0.0133$ ---
within the synthetic benchmark's median. The two small-cohort
comparisons (NCT03041311 Placebo $n{=}53$: $0.0309$; NCT02499770 E/P
$n{=}37$: $0.0345$) are larger, which is expected: the real IPD KM at
$n{\sim}40$ is itself a noisy estimator of the underlying survival
function, so the residual folds in both extraction error and
finite-sample noise in the reference, rather than extraction error
alone.

\paragraph{Relation to KM-GPT.} These numbers are in the same range as
those reported by KM-GPT~\cite{zhao2025km}, obtained here without
human-in-the-loop correction and on a benchmark that deliberately
includes adversarial degradations (Section~\ref{sec:km-benchmark}). We
do not report a head-to-head run: KM-GPT is released as a hosted
interface rather than as code, which precludes batch evaluation on our
32-plot benchmark, so the comparison is against published numbers on
different plots and should be read as indicative rather than
controlled. The substantive distinction is scope --- extraction-only
work stops at the curve, whereas the contribution evaluated in the
remainder of this section is per-patient AE generation on top of it.

\subsection{IPD Generation Results}
\label{sec:results}

\paragraph{Novelty.} To our knowledge, this is the first synthetic-IPD pipeline that generates adverse events as an integral part of full individual patient data reconstruction. Prior IPD reconstruction work~\cite{guyot2012,wei2017,liu2021} targets survival curves only; synthetic health-record generators~\cite{goncalves2020} model demographics and diagnoses but not trial-specific treatment-emergent AEs with phase-stratified timing and organ-system risk stratification.

\paragraph{Experimental setup.} We evaluate on three trials with real IPD available to the authors: \textbf{NCT03041311} (trilaciclib vs.\ placebo, small-cell lung cancer, $n_r{=}53$), \textbf{NCT02499770} (targeted--chemo combination, $n_r{=}37$), and \textbf{NCT00844649} (nab-paclitaxel + gemcitabine, pancreatic cancer, $n_r{=}430$). These span an order of magnitude in sample size and three drug classes. All sampler parameters use a single set of defaults (grounded in literature and LLM priors) with no per-trial tuning. Real IPD are firewalled from the extraction agent (Appendix~\ref{app:implementation}).

\paragraph{Headline results.} Table~\ref{tab:results} reports all metrics. Stage~A extraction-quality scalar errors are ${\leq}2\%$ for demographic moments and AE shared-term MAE is below 0.035. End-to-end (Stage~B), mean $\Delta_{\text{KM}} \leq 0.051$ and top-15 AE overlap is ${\geq}71\%$ on every trial. Stage~B numbers are mean~$\pm$~std over 30 independent regenerations of the synthetic cohort (seeds $0,\dots,29$); Stage~A is deterministic from the extracted configs.

\begin{table}[t]
\centering
\small
\setlength{\tabcolsep}{4pt}
\caption{End-to-end fidelity and extraction quality. Synthetic cohorts at $2{\times}$ real enrollment. $\downarrow$ = lower is better, $\uparrow$ = higher is better. JSD $\in [0, \log 2]$, KS $\in [0, 1]$. Stage~B values are reported as mean~$\pm$~std across 30 independent regenerations of the synthetic cohort (seeds $0,\dots,29$); Stage~A metrics are deterministic functions of the extracted configs and have no resampling variability.}
\label{tab:results}
\resizebox{\linewidth}{!}{%
\begin{tabular}{lccc}
\toprule
 & \textbf{NCT03041311} & \textbf{NCT02499770} & \textbf{NCT00844649} \\
 & \scriptsize{$n_r{=}53,\, n_s{=}107$} & \scriptsize{$n_r{=}37,\, n_s{=}75$} & \scriptsize{$n_r{=}430,\, n_s{=}861$} \\
\midrule
\multicolumn{4}{l}{\textit{Demographics (Stage B, mean $\pm$ std over 30 regens)}} \\
\quad Sex JSD $\downarrow$                    & 0.0054 $\pm$ 0.0064 & 0.0033 $\pm$ 0.0060 & 0.0005 $\pm$ 0.0006 \\
\quad ECOG JSD $\downarrow$                   & 0.272 $\pm$ 0.048   & 0.013 $\pm$ 0.012   & 0.003 $\pm$ 0.001   \\
\quad Weight KS $\downarrow$                  & 0.201 $\pm$ 0.029   & 0.202 $\pm$ 0.050   & 0.129 $\pm$ 0.013   \\
\quad Height KS $\downarrow$                  & 0.217 $\pm$ 0.043   & 0.166 $\pm$ 0.041   & 0.102 $\pm$ 0.014   \\
\quad BMI KS $\downarrow$                     & 0.161 $\pm$ 0.026   & 0.176 $\pm$ 0.034   & 0.150 $\pm$ 0.016   \\
\midrule
\multicolumn{4}{l}{\textit{Overall survival (Stage B, mean $\pm$ std over 30 regens)}} \\
\quad Median OS $|\Delta|$ (mo.) $\downarrow$ & 0.84 $\pm$ 0.76     & 1.12 $\pm$ 0.83     & 1.28 $\pm$ 0.34     \\
\quad Median OS rel.\ error $\downarrow$      & $7.0 \pm 6.3$\%     & $11.3 \pm 8.4$\%    & $19.4 \pm 5.1$\%    \\
\quad OS KS $\downarrow$                      & 0.173 $\pm$ 0.055   & 0.156 $\pm$ 0.039   & 0.112 $\pm$ 0.018   \\
\quad $\Delta_{\text{KM}}$ $\downarrow$       & 0.049 $\pm$ 0.018   & 0.051 $\pm$ 0.018   & 0.040 $\pm$ 0.011   \\
\quad Mann--Whitney $U$ $p$                   & 0.63 $\pm$ 0.29     & 0.60 $\pm$ 0.28     & 0.008 $\pm$ 0.031   \\
\midrule
\multicolumn{4}{l}{\textit{Adverse event frequency (Stage B, mean $\pm$ std over 30 regens)}} \\
\quad AE JSD $\downarrow$                     & 0.186 $\pm$ 0.010   & 0.208 $\pm$ 0.010   & 0.144 $\pm$ 0.002   \\
\quad AE cosine similarity $\uparrow$         & 0.852 $\pm$ 0.018   & 0.842 $\pm$ 0.026   & \textbf{0.940 $\pm$ 0.003} \\
\quad Top-15 overlap $\uparrow$               & 11.7 $\pm$ 0.8 / 15 & 10.7 $\pm$ 0.8 / 15 & \textbf{12.0 $\pm$ 0.2 / 15} \\
\quad Mean per-AE $|\Delta|$ $\downarrow$     & 0.106 $\pm$ 0.005   & 0.110 $\pm$ 0.004   & \textbf{0.027 $\pm$ 0.001} \\
\midrule
\multicolumn{4}{l}{\textit{AE timing and burden (Stage B, mean $\pm$ std over 30 regens)}} \\
\quad Mean onset $|\Delta|$ (mo.) $\downarrow$& 0.71 $\pm$ 0.13     & 0.28 $\pm$ 0.12     & 0.77 $\pm$ 0.04     \\
\quad Onset KS $\downarrow$                   & 0.153 $\pm$ 0.021   & 0.270 $\pm$ 0.017   & 0.253 $\pm$ 0.004   \\
\quad AE burden (real $\to$ synth)            & 15.1 $\to$ 10.1 $\pm$ 0.5 & 12.8 $\to$ 6.9 $\pm$ 0.6 & 15.4 $\to$ 11.4 $\pm$ 0.2 \\
\quad Burden KS $\downarrow$                  & 0.277 $\pm$ 0.029   & 0.457 $\pm$ 0.057   & 0.185 $\pm$ 0.010   \\
\midrule
\multicolumn{4}{l}{\textit{Extraction quality (Stage A)}} \\
\quad Age mean rel.\ error $\downarrow$       & 0.9\%  & 0.9\%  & 1.3\%  \\
\quad Sex ratio abs.\ error $\downarrow$      & 0.059  & 0.001  & 0.015 \\
\quad Height mean rel.\ error $\downarrow$    & 0.7\%  & 0.8\%  & $<$0.1\% \\
\quad ECOG dist.\ JSD $\downarrow$            & 0.269  & 0.005  & 0.002  \\
\quad AE shared-term MAE $\downarrow$         & 0.035  & 0.030  & \textbf{0.017} \\
\quad AE shared-term RMSE $\downarrow$        & 0.045  & 0.044  & 0.033  \\
\quad AE full-term JSD $\downarrow$           & 0.154  & 0.198  & 0.156  \\
\quad Shared / real-only / ext-only           & 71/67/27 & 57/70/9 & 190/377/59 \\
\bottomrule
\end{tabular}%
}
\end{table}

\subsection{Extraction Quality (Stage~A)}

Scalar demographics are captured with relative errors ${\leq}1.3\%$ across all trials for age, height, and weight means, confirming faithful parsing of ClinicalTrials.gov baseline tables. The noisiest scalar is the sex ratio on NCT03041311 (5.9~pp error), driven by small cohort size rather than systematic failure. AE shared-term MAE ranges from 1.7--3.5~pp and shrinks with trial size, as expected given ClinicalTrials.gov's non-serious event 5\% reporting threshold. The shared/real-only/extracted-only confirms the expected asymmetry: the real-only tail (67--377 terms) is always larger, because real IPD records every event while ClinicalTrials.gov truncates below 5\%. Race and region JSD$\,{=}\,1.0$ on all trials due to non-overlapping label taxonomies (e.g., ``White'' vs.\ ``Caucasian''); this is a known label-alignment limitation.

\subsection{End-to-End Fidelity (Stage~B)}

\paragraph{Demographics.} Sex JSD is ${\leq}0.006$ on every trial and ECOG JSD is ${\leq}0.013$ on the two trials whose IPD reports the full $\{0,1,2\}$ ECOG support. The exception is ECOG on NCT03041311 (JSD${=}0.272{\pm}0.048$), where real IPD reports only ECOG~$\{1,2\}$ while the extracted config permits ECOG~0; this is a support mismatch, not a shape error. Anthropometric KS means are ${\leq}0.22$ across all three trials (Figure~\ref{fig:supp-demo}); the statistically significant rejections on NCT00844649 reflect power at $n_r{=}430$, not large effect size.

\paragraph{Overall survival.} Figure~\ref{fig:km} shows nearly overlapping curve pairs. Mean $\Delta_{\text{KM}}$ is $0.049$, $0.051$, and $0.040$ respectively, all below $0.06$. NCT03041311 achieves median OS relative error of $7.0{\pm}6.3\%$ with Mann--Whitney $p{=}0.63{\pm}0.29$ (failing to reject in 28 of 30 seeds). The larger median overshoots on the remaining two trials ($11{\pm}8\%$ on NCT02499770; $19{\pm}5\%$ on NCT00844649) inherit slightly longer right tails from upstream KM reconstruction --- the sampler is faithful to its input. The Mann--Whitney rejection on NCT00844649 ($p{=}0.008{\pm}0.031$, rejecting in 29 of 30 seeds) is driven by the $n_r{=}430$ power on a $1.3$-month median shift, not a structural curve mismatch.

\paragraph{Extraction-error propagation.}
\label{sec:propagation}
Because the survival channel is the only one traced from a plot image (Section~\ref{sec:generation-methodology}), the effect of extraction error on the end-to-end result can be measured directly by swapping the input curve. For the one arm per trial with matched real IPD, we regenerate twice under identical seeds: once from the \emph{extracted} published KM, and once from an \emph{oracle} KM reconstructed from the held-out IPD itself. The difference between the two $\Delta_{\text{KM}}$ values is the portion of generative error attributable to extraction. Table~\ref{tab:propagation} (Appendix~\ref{app:additional}) reports the result: the extracted-input $\Delta_{\text{KM}}$ exceeds the oracle-input value by $0.016$, $0.007$, and $0.006$ on the three trials --- at most about half of the corresponding extraction IAE, and a minority (12--32\%) of total $\Delta_{\text{KM}}$ in every case. Extraction error is therefore bounded and does not amplify through the sampler; the residual is dominated by finite-sample noise in the reference KM, which is why the gap is largest on the smallest cohort.

\begin{figure}[t]
\centering
\includegraphics[width=\linewidth]{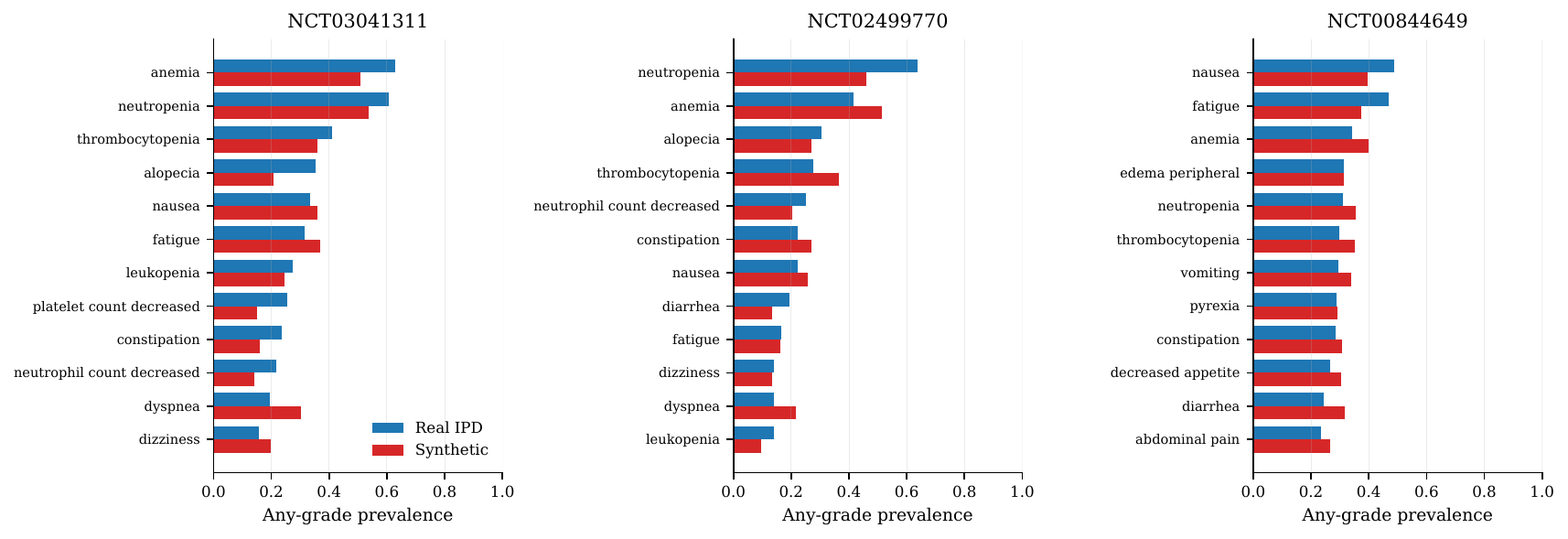}
\caption{Top-12 shared AE terms per trial, ranked by real-IPD prevalence. Top-15 overlap is $71$--$78\%$ on the small trials and $80\%$ on NCT00844649 (means over 30 regenerations).}
\label{fig:ae-freq}
\end{figure}

\paragraph{AE frequency.} Figure~\ref{fig:ae-freq} shows that rank ordering agrees on the dominant hematologic and GI terms across all trials. Cosine similarity reaches $0.940{\pm}0.003$ on the largest (pancreatic) trial, and mean per-AE incidence difference shrinks from $0.106$ on $n_r{=}53$ to $0.027$ on $n_r{=}430$ --- AE fidelity scales with source-table completeness rather than an intrinsic pipeline ceiling.

\begin{figure}[t]
\centering
\includegraphics[width=\linewidth]{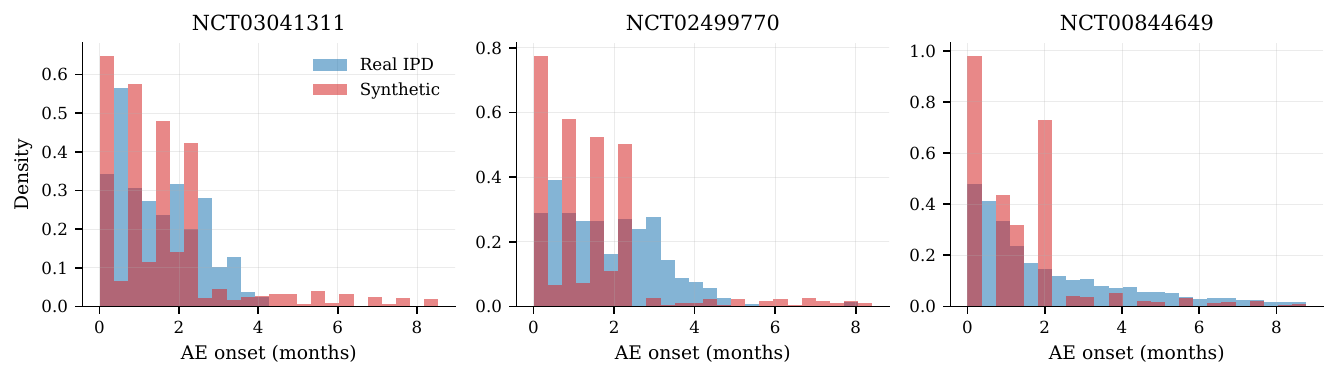}
\caption{AE onset density. Synthetic onsets show cycle-length striping (a direct consequence of cycle-based sampling); real onsets are smoother due to continuous-time recording. Mean onset error is ${\leq}0.71$ months.}
\label{fig:ae-onset}
\end{figure}

\paragraph{AE timing.} Figure~\ref{fig:ae-onset} shows the expected periodic striping in synthetic onsets at cycle-length multiples (${\approx}0.7$ months for a 21-day cycle). Part of this periodicity is a property of the data-generating process rather than an artifact: trials ascertain adverse events at scheduled on-treatment visits, so real onset times are themselves visit-quantized, and a continuous-onset model would misstate when events are \emph{recorded}. What is artifactual is the degree of concentration --- our onsets fall in the first third of each cycle window, which is tighter than real ascertainment. Despite this, mean-onset error stays below $0.8$ months on every trial and the dominant early-phase mass is preserved; onset KS is ${\leq}0.27$. Spreading onsets uniformly across the full cycle, or convolving with a detection-delay kernel, removes the visual striping without altering any component of the model.

\paragraph{AE burden.} Synthetic burden systematically undershoots real burden by $4$--$6$ events per patient across all trials (Figure~\ref{fig:supp-burden}) at the default $\sigma_\eta{=}0.4$. The lognormal propensity reproduces within-cohort dispersion, but the overall AE budget is under-calibrated. Because survival is assigned independently of AE sampling (Section~\ref{sec:survival}), this gap can be closed post hoc without disturbing the other channels: Table~\ref{tab:sigma-sweep} (Appendix~\ref{app:additional}) re-runs the full 30-regeneration protocol at $\sigma_\eta \in \{0.4, 0.6, 0.8, 1.0\}$ and shows burden converging on real burden while $\Delta_{\text{KM}}$ stays numerically identical and AE-frequency metrics move within noise. At $\sigma_\eta{=}0.8$ the mean burden gap narrows from $4$--$6$ to $0.7$--$3.3$ events per patient and burden KS improves on all three trials (by 31\%, 25\%, and 66\% relative), with the largest gain on the largest cohort. We therefore recommend $\sigma_\eta{=}0.8$ as the default and report the corresponding full Stage-B table as Table~\ref{tab:results-sigma08}. Table~\ref{tab:results} retains $\sigma_\eta{=}0.4$ so that the headline results reflect the single untuned parameter set. Appendix~\ref{sec:ablations} gives the aggregate ablation.

\begin{figure}[t]
\centering
\includegraphics[width=\linewidth]{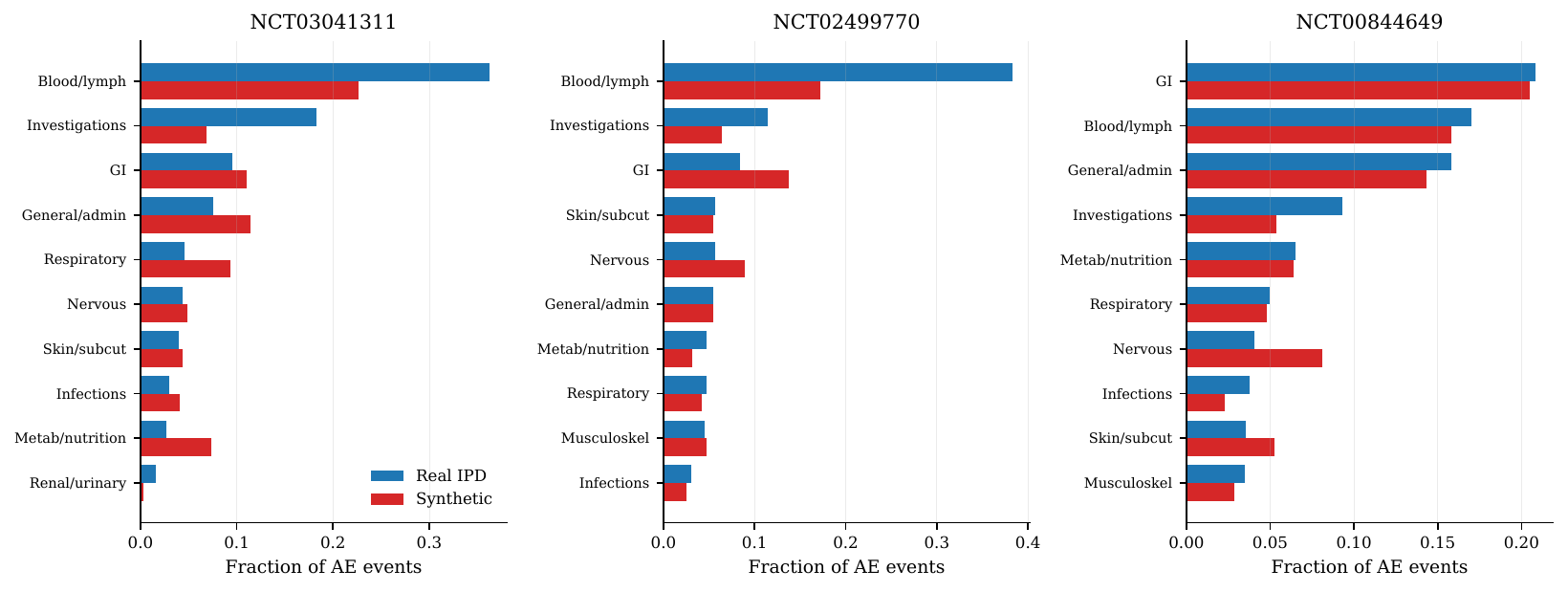}
\caption{Top-10 SOC distributions. The dominant real SOC is the top synthetic SOC in every trial, confirming correct organ-system targeting by the archetype risk multipliers.}
\label{fig:soc}
\end{figure}

\paragraph{Organ-system distribution.} Figure~\ref{fig:soc} confirms correct SOC rank ordering on all trials without per-trial tuning.

\section{Discussion}
\label{sec:paper-discussion}

Access to individual patient data (IPD) remains a critical bottleneck in clinical research, forcing downstream analyses to rely on aggregated summary statistics that obscure granular
event timing and censoring patterns. We introduce KMGen, the first end-to-end framework that (i) fully automates KM curve extraction at accuracy competitive with human-guided tools, and (ii) generates synthetic per-patient AE trajectories from public trial registry records. The extraction component achieves a mean IAE of 0.0151 on a 32-plot synthetic benchmark and 0.0262 on three published oncology trials. Using a single fixed parameter set across trials spanning an order of magnitude in cohort size (n = 37 to 430) and three distinct drug classes, KMGen produced synthetic cohorts matching real IPD across key metrics: $\Delta_{\text{KM}} \leq 0.051$, sex/ECOG JSD $\leq 0.013$ on five of six demographic slots, and recovery of $\geq 71\%$ of top-15 adverse events by exact MedDRA term.

Decoupling patient archetype extraction from statistical generation provides three properties relevant to method development. First, the pipeline is interpretable end-to-end—clinical priors such as the induction–maintenance phase split and archetype risk multipliers are explicit, parameterized components rather than learned weights. Second, bootstrap rank-correlation coupling preserves the marginal survival distribution exactly by construction, independent of the rank-noise parameter $\sigma$ (confirmed in ablation studies). Third, the extraction agent operates without access to real IPD, eliminating the possibility of inadvertent memorization or data leakage from protected health information.

\paragraph{Scope of the oncology tuning.} All three evaluation trials are oncology trials, but the oncology-specific content of the framework is a small, overridable set of priors rather than an architectural commitment. Only the archetype risk multipliers (Appendix~\ref{app:multipliers}) and the induction fraction $\alpha$ are tuned to cytotoxic oncology; the configuration schema, the source-priority hierarchy, the rank-correlation coupling, and the evaluation suite make no assumption about therapeutic area. Adaptation has two knobs. For a different drug class, the override dictionaries redirect risk to the relevant organ systems (Appendix~\ref{app:skill_2} lists the immunotherapy, platinum, and TKI cases). For a non-cyclic or continuously monitored design, the cycle length $c$ is set to the assessment interval and $\alpha$ is lowered toward $0.5$, which recovers a near-uniform onset process; the sampler itself is unchanged. The three trials do span three drug classes --- a CDK4/6 inhibitor, a targeted--chemotherapy combination, and a cytotoxic doublet --- over a $10{\times}$ cohort range under one fixed parameter set, but we make a claim of architectural generality, not of validated non-oncology performance, and scope the latter as future work.

\paragraph{Sensitivity to the underlying LLM.} Both agents run on Claude Opus~4.6. The LLM's role is deliberately narrow: in the extraction stage it selects techniques from the toolbox and writes the calling code, while the pixel-level coordinates are produced by deterministic computer-vision routines behind the automatic self-checks of Section~\ref{sec:km-methodology}; in the generation stage it populates a typed configuration whose fields are then consumed by a deterministic sampler. Model changes can therefore alter \emph{strategy selection} but cannot silently perturb the numerics, and a mis-traced curve surfaces in the overlay and anchor checks rather than propagating unnoticed. We have not run a controlled cross-model comparison, so we do not claim invariance to the backbone; stronger models would plausibly improve technique selection on the adversarial tail of the benchmark, which is where our residual error concentrates.

The synthetic IPD produced by KMGen is intended for methodological applications where access to real IPD is restricted, including stress-testing survival models under varying censoring patterns and validating safety-monitoring algorithms. These cohorts are not substitutes for randomized controlled trials and do not support regulatory claims. The framework models structural correlation between adverse events and survival—shorter survival mechanistically limits event opportunity through the realized cycle count $N_i$—but does not simulate causal pathways by which treatment toxicity induces mortality.

\paragraph{Conclusion}
KMGen provides a unified framework for extracting survival data from published Kaplan--Meier plots and generating synthetic individual patient data from clinical trial registries. The mechanistic sampling approach—decoupling LLM-driven patient archetype extraction from deterministic statistical generation—ensures interpretability and marginal distribution preservation by construction. Evaluated on three oncology trials spanning 37 to 430 patients over 30 independent regenerations per trial, the framework achieves a mean $\Delta_{\text{KM}} \leq 0.051$ and top-15 AE overlap $\geq 71\%$. The resulting synthetic IPD is suitable for methodological development in settings where access to real patient-level data is restricted. The framework is released as open source to enable community validation and integration into existing workflows for secondary analysis of clinical trial data.

\bibliography{sample}

\newpage
\appendix

\section{Generation Algorithm}
\label{app:algorithm}

\begin{algorithm}[!ht]
\hrule height 0.8pt
\vspace{2pt}
\caption{Synthetic patient generation. Notation: $n$ = cohort size; $(t_k, e_k)$ = empirical KM time--event pairs, $t_k$ = start time of cycle $k$; $p_a, s_a$ = base AE rate and SOC for AE term $a$; $c$ = cycle length (months); $K_{\mathrm{ind}}$ = induction cycles; $\alpha$ = induction AE fraction; $\mathcal{L}$ = archetype set; $w_\ell$ = archetype weight; $m^{\mathrm{OS}}_\ell, m^{\mathrm{AE}}_\ell(s)$ = survival and per-SOC AE risk multipliers; $r_i$ = frailty score; $\sigma$ = rank-noise strength; $\eta_i$ = patient propensity; $\sigma_\eta$ = propensity dispersion; $\phi_k$ = phase weight; $\mathbf{d}_i$ = demographics; $\mathcal{A}_i$ = AE event list. $N_i^{\text{phase}(k)}$ = cycles in phase $k$ for patient $i$}
\label{alg:generation}
\vspace{2pt}
\hrule
\vspace{2pt}

\KwIn{Trial configuration $\mathcal{C}$}
\KwOut{Synthetic cohort $\{(\mathbf{d}_i,\, t_i^{\mathrm{OS}},\, e_i,\, \mathcal{A}_i)\}_{i=1}^{n}$}

\tcc{Stage 1 --- Archetype construction (\S\ref{sec:archetypes})}
$\mathcal{L} \gets \{\text{age}\} \times \{\text{sex}\} \times \{\text{ECOG}\} \times \{\text{comorbidity}\}$\;
\ForEach{archetype $\ell \in \mathcal{L}$}{
    $w_\ell \gets \prod_d p(\ell_d)$; \quad $m^{\mathrm{OS}}_\ell \gets \prod_d m^{\mathrm{OS}}_{\ell_d}$; \quad $m^{\mathrm{AE}}_\ell(s) \gets \prod_d m^{\mathrm{AE}}_{\ell_d}(s)$\;
}
\BlankLine

\tcc{Stage 2 --- Patient instantiation \& survival coupling (\S\ref{sec:survival})}
\For{$i \gets 1$ \KwTo $n$}{
    $\ell(i) \sim \mathrm{Categorical}(\{w_\ell\})$; \quad sample $\mathbf{d}_i \mid \ell(i)$\;
    $r_i \gets -\log(m^{\mathrm{OS}}_{\ell(i)}) + \varepsilon_i$, \quad $\varepsilon_i \sim \mathcal{N}(0, \sigma^2)$\;
}
Bootstrap $n$ pairs $\{(\tilde{t}_j, \tilde{e}_j)\}$ from KM; sort by time\;

$\pi \gets \mathrm{argsort}(\mathbf{r})$; 

\quad $(t_i^{\mathrm{OS}}, e_i) \gets (\tilde{t}_{(n - \pi(i) + 1)},\, \tilde{e}_{(n - \pi(i) + 1)})$\;

\BlankLine

\tcc{Stage 3 --- Adverse event sampling (\S\ref{sec:ae-sampling})}
\For{$i \gets 1$ \KwTo $n$}{
    $\eta_i \sim \mathrm{LogNormal}(0, \sigma_\eta^2)$; 
    
    \quad $N_i \gets \max(1, \lfloor t_i^{\mathrm{OS}} / c \rfloor)$\;
    
    \For{cycle $k \gets 1$ \KwTo $N_i$}{
        $\phi_k \gets \alpha\,\mathbf{1}[k \leq K_{\mathrm{ind}}] + (1{-}\alpha)\,\mathbf{1}[k > K_{\mathrm{ind}}]$\;
        \ForEach{AE $a$}{
            $\lambda \gets p_a \cdot m^{\mathrm{AE}}_{\ell(i)}(s_a) \cdot \eta_i \cdot \phi_k / N_i^{\mathrm{phase}(k)}$\;
            
            $M \sim \mathrm{Poisson}(\lambda)$; \quad onsets $\sim \mathrm{Uniform}(t_k,\, t_k {+} c/3)$\;
            
            $\mathcal{A}_i \gets \mathcal{A}_i \cup \{(\text{onset}, a)\}^M$\;
        }
    }
}
\Return{$\{(\mathbf{d}_i,\, t_i^{\mathrm{OS}},\, e_i,\, \mathcal{A}_i)\}_{i=1}^{n}$}
\vspace{2pt}
\hrule height 0.8pt
\end{algorithm}

\section{Risk-Multiplier Evidence Base}
\label{app:multipliers}

\begin{table}[t]
\centering
\small
\setlength{\tabcolsep}{4pt}
\caption{Default AE risk multipliers by patient factor and organ system. Per-SOC values are clinical estimates informed by the overall effect sizes in the cited studies (Appendix~\ref{app:multipliers}). All multipliers are relative to a baseline of 1.0.}
\label{tab:risk-multipliers}
\resizebox{\linewidth}{!}{%
\begin{tabular}{llll}
\toprule
\textbf{Factor} & \textbf{Key SOCs affected} & \textbf{Default} & \textbf{Evidence} \\
\midrule
\multicolumn{4}{l}{\textit{Age $\geq$ 65 (``older'')}} \\
 & Blood \& lymphatic      & 1.5 & \cite{lyman2003,hurria2011} \\
 & Cardiac                 & 1.8 & \cite{swain2003} \\
 & Infections              & 1.6 & \cite{kuderer2006} \\
 & Nervous system          & 1.3 & \cite{argyriou2012} \\
 & Renal \& urinary        & 1.4 & \cite{launay2007} \\
\midrule
\multicolumn{4}{l}{\textit{Female sex}} \\
 & Gastrointestinal        & 1.3 & \cite{unger2022} \\
 & Blood \& lymphatic      & 1.25 & \cite{unger2022} \\
 & Nervous system          & 1.2 & \cite{schmetzer2012} \\
\midrule
\multicolumn{4}{l}{\textit{Poor ECOG ($\geq 2$)}} \\
 & Blood \& lymphatic      & 1.6 & \cite{hurria2011,lyman2011} \\
 & Infections              & 2.0 & \cite{lyman2011} \\
 & Cardiac                 & 1.6 & \cite{hurria2011,feliu2020} \\
 & Gastrointestinal        & 1.4 & \cite{sargent2009} \\
\midrule
\multicolumn{4}{l}{\textit{High comorbidity (CCI $\geq 2$)}} \\
 & Infections              & 1.5 & \cite{gross2007,kuderer2006} \\
 & Cardiac                 & 1.6 & \cite{sogaard2013} \\
 & Vascular                & 1.5 & \cite{sogaard2013} \\
\midrule
\multicolumn{4}{l}{\textit{OS multipliers by factor}} \\
 & Older / younger         & 0.88 / 1.12 & \cite{hurria2011,extermann2012} \\
 & Poor ECOG               & 0.62 & \cite{sargent2009} \\
 & High comorbidity        & 0.72 & \cite{gross2007,sogaard2013} \\
\bottomrule
\end{tabular}%
}
\end{table}

The per-SOC multipliers in Table~\ref{tab:risk-multipliers} are clinical estimates informed by the following studies. None report per-organ relative risks directly; the multipliers are calibrated to overall effect sizes.

\paragraph{Age.} Hurria et al.~\cite{hurria2011} report OR~1.85 (95\% CI 1.22--2.82) for grade~3--5 toxicity in patients aged ${\geq}72$, without organ-system stratification. Swain et al.~\cite{swain2003} document age-dependent cardiac risk from doxorubicin. Argyriou et al.~\cite{argyriou2012} review CIPN risk factors, though subsequent work found age was not a clear independent predictor. Launay-Vacher et al.~\cite{launay2007} report high prevalence of renal insufficiency in elderly oncology patients.

\paragraph{Sex.} Unger et al.~\cite{unger2022} report OR~1.34 (95\% CI 1.27--1.42) for grade~3+ AEs in women vs.\ men, with hematologic OR$\,{=}\,1.30$ and symptomatic OR$\,{=}\,1.33$, but without finer SOC breakdowns. Schmetzer \& Fl\"orcken~\cite{schmetzer2012} review sex-based PK and toxicity differences.

\paragraph{ECOG.} Sargent et al.~\cite{sargent2009} report absolute GI toxicity rates by performance status (grade~${\geq}3$ nausea: 16.4\% PS\,2 vs.\ 8.5\% PS\,0--1). Feliu et al.~\cite{feliu2020} found ECOG was not independently significant (OR$\,{=}\,1.30$, $p{=}0.236$). Lyman et al.~\cite{lyman2011} model neutropenia risk; ECOG was collected but absent from the final multivariable model.

\paragraph{Comorbidity.} Gross et al.~\cite{gross2007} find heart failure OR$\,{=}\,0.49$ for chemotherapy receipt (high-comorbidity patients underrepresented). S{\o}gaard et al.~\cite{sogaard2013} review comorbidity effects on cancer survival. Kuderer et al.~\cite{kuderer2006} report comorbidity-dependent febrile neutropenia mortality.

\paragraph{OS multipliers.} The OS multipliers for age and comorbidity are inferred via the toxicity--survival association, not from reported survival ratios directly. Both Hurria et al.~\cite{hurria2011} and Extermann et al.~\cite{extermann2012} predict chemotherapy toxicity rather than OS.

\section{Implementation Details}
\label{app:implementation}

\paragraph{Agent backbone.} Both agents --- the KM-extraction agent of Appendix~\ref{app:skill_1} and the configuration-extraction agent of Appendix~\ref{app:skill_2} --- run on Claude Opus~4.6 with default decoding settings and no sampling-parameter overrides. Neither agent produces numeric outputs directly: the extraction agent selects techniques and emits Python that a deterministic computer-vision routine executes, and the configuration agent fills a typed schema that the sampler consumes. All results in this paper are from a single backbone; we did not sweep models.

\paragraph{Data-leakage firewall.} The extraction agent never reads files under the IPD directory. The KM CSV discovery glob explicitly filters out paths containing the IPD segment, ensuring real patient data is used only for evaluation.

\paragraph{Real IPD ingestion.} Trial-specific loaders read CDISC SDTM-AD files for three supported trials. NCT03041311 and NCT02499770 use standard ADSL/ADAE/ADTTE domains. NCT00844649 uses a non-standard structure: demographics from DM, ECOG derived from Karnofsky performance status via the standard mapping (KPS$\,{\geq}\,90 \to 0$, $70$--$89 \to 1$, ${<}70 \to 2$), and survival assembled from follow-up and disposition domains.

\paragraph{AE term normalization.} A canonicalization function lowercases, collapses whitespace, strips punctuation, and rewrites British to American MedDRA spellings (e.g., ``anaemia'' $\to$ ``anemia''). Applied at both extraction and evaluation time to prevent cosmetic spelling differences from inflating apparent mismatch.

\paragraph{Weibull fallback.} When no empirical KM CSV is available, the sampler falls back to a parametric Weibull model~\cite{carroll2003}. This branch is not exercised in the main experiments.

\section{Ablation Studies}
\label{sec:ablations}

We ablate key design choices and conduct a sample-size sensitivity analysis on NCT03041311. Each ablation targets a specific methodological claim; we state the motivation before the result.

\paragraph{Induction fraction $\alpha$.}
The induction/maintenance split is the main structural assumption in the AE timing model. 

\begin{figure}[ht]
\centering
\includegraphics[width=\linewidth]{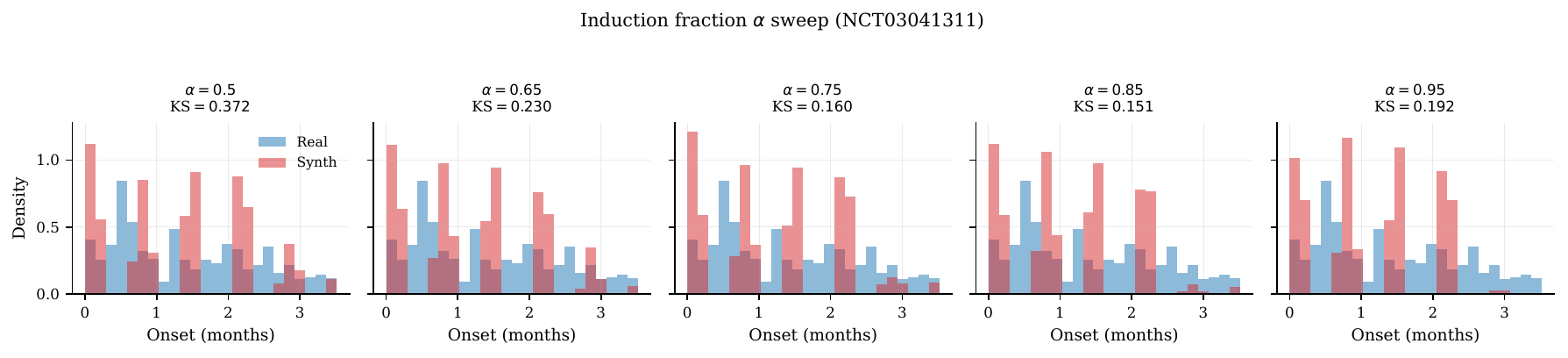}
\caption{Induction fraction sweep. The onset KS statistic is minimized at $\alpha{=}0.85$ (KS$\,{=}\,0.151$), with sharp degradation at low $\alpha$, validating the induction--maintenance phase split.}
\label{fig:ablation-induction}
\end{figure}

Figure~\ref{fig:ablation-induction} sweeps $\alpha \in \{0.50, 0.65, 0.75, 0.85, 0.95\}$. The onset KS statistic drops sharply from 0.372 at $\alpha{=}0.50$ (uniform onset, too much late-phase mass) to a minimum of 0.151 at $\alpha{=}0.85$, then rises to 0.192 at $\alpha{=}0.95$ (over-concentrated early onset). This U-shaped curve validates the induction--maintenance split and confirms the default $\alpha{=}0.80$--$0.85$ for cytotoxic regimens, consistent with clinical evidence that early cycles carry the highest AE burden~\cite{crawford2008}.

\begin{figure}[ht]
    \centering
    \includegraphics[width=0.5\linewidth]{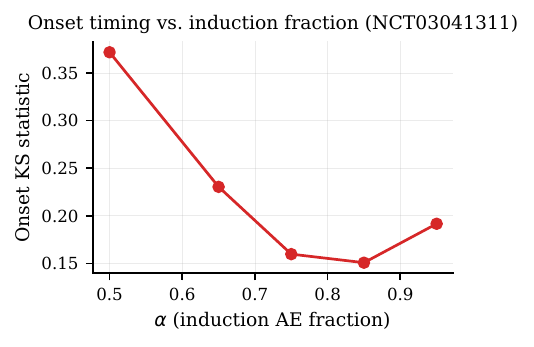}
    \caption{Onset KS as a function of induction fraction $\alpha$, showing a clear minimum near $\alpha{=}0.85$.}
    \label{fig:ablation_induction_summary}
\end{figure}

\paragraph{Patient propensity $\sigma_\eta$.}
The AE burden undershoot (4--5 events/patient) is the pipeline's largest weakness. This ablation directly investigates whether increasing inter-patient overdispersion closes the gap, and whether the lognormal frailty is load-bearing at all.

\begin{figure}[ht]
\centering
\includegraphics[width=\linewidth]{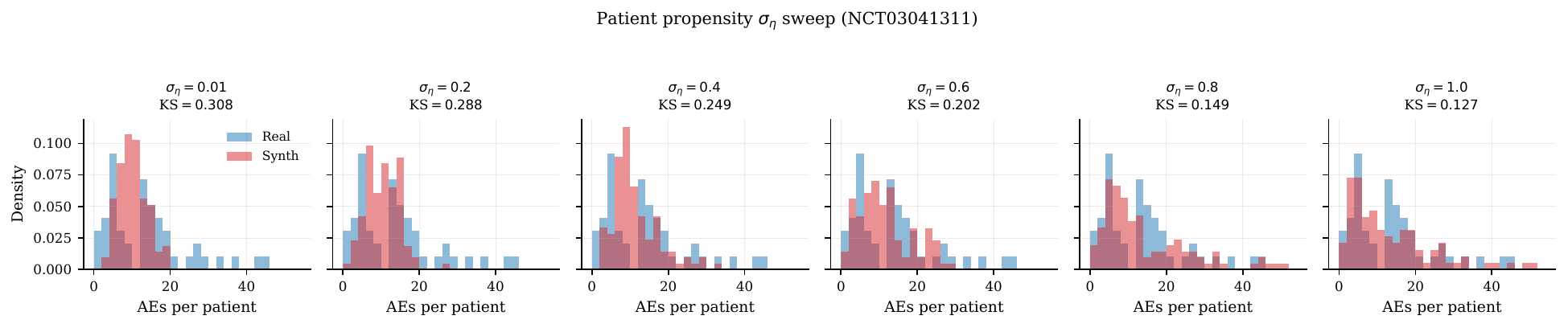}
\caption{Patient propensity sweep. Burden KS decreases monotonically, confirming the lognormal frailty is necessary. The default $\sigma_\eta{=}0.4$ is conservative; higher values further reduce the burden gap.}
\label{fig:ablation-propensity}
\end{figure}

Figure~\ref{fig:ablation-propensity} sweeps $\sigma_\eta \in \{0.01, 0.2, 0.4, 0.6, 0.8, 1.0\}$. Burden KS decreases monotonically from 0.308 (no frailty) to 0.127 ($\sigma_\eta{=}1.0$), confirming that the lognormal frailty is load-bearing: without it, the synthetic burden distribution is far too narrow. The monotonic improvement suggests real per-patient AE counts are heavily right-tailed, and even $\sigma_\eta{=}1.0$ does not over-disperse. The default $\sigma_\eta{=}0.4$ (KS$\,{=}\,0.249$) is a conservative choice that balances dispersion realism against numerical stability in small cohorts. Table~\ref{tab:sigma-sweep} repeats this sweep per trial under the full 30-regeneration protocol and confirms that the improvement is not purchased at the expense of the other channels: $\sigma_\eta{=}0.8$ closes most of the burden gap on all three trials while $\Delta_{\text{KM}}$ is unchanged to four decimal places and AE cosine similarity moves by ${<}0.01$. We therefore recommend $\sigma_\eta{=}0.8$ as the operating default.

\begin{figure}[ht]
\centering
\includegraphics[width=0.49\linewidth]{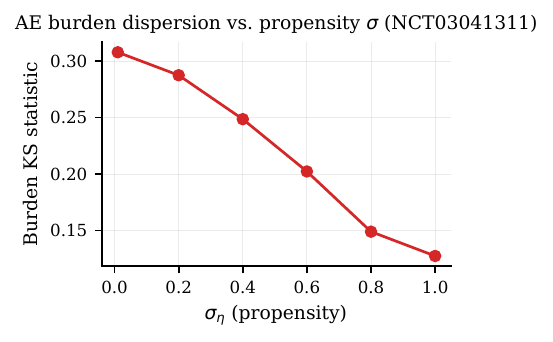}
\caption{Burden KS as a function of patient propensity $\sigma_\eta$. KS decreases monotonically; the default $\sigma_\eta{=}0.4$ is conservative --- higher values further reduce the gap.}
\label{fig:ablation-propensity-summary}
\end{figure}

\paragraph{Archetype risk multipliers.}
The archetype machinery (cross-product of age $\times$ sex $\times$ ECOG $\times$ comorbidity with per-SOC multipliers) is the most complex component of the sampler. This ``nuclear option'' ablation removes the entire archetype system by setting all multipliers to 1.0, testing whether it contributes measurably or is dead weight.

\begin{figure}[ht]
\centering
\includegraphics[width=\linewidth]{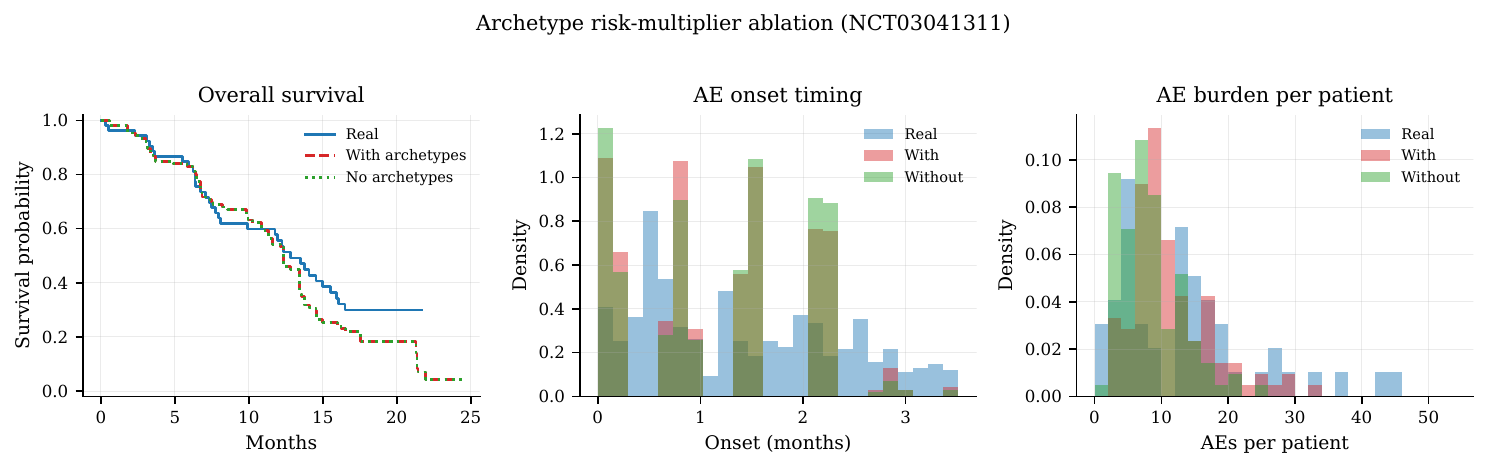}
\caption{Archetype ablation. Removing risk multipliers preserves marginal survival but degrades AE timing and burden realism.}
\label{fig:ablation-arch}
\end{figure}

Figure~\ref{fig:ablation-arch} compares the full model against a flat-multiplier ablation (all $m^{\text{AE}}_{\ell_d}(s) = 1$, all $m^{\text{OS}}_{\ell_d} = 1$). Removing archetypes preserves marginal KM fidelity (by construction, since bootstrap coupling is unchanged) but eliminates within-cohort heterogeneity in AE onset and burden. The full model's AE burden distribution is closer to real IPD, confirming that organ-system-level risk stratification contributes to realistic per-patient event profiles.

\paragraph{Sample size sensitivity.}
We generate synthetic cohorts at $2{\times}$ real enrollment ``to stabilize distributional metrics''. We sweep the multiplier from $0.5{\times}$ to $5{\times}$.

\begin{figure}[ht]
\centering
\includegraphics[width=\linewidth]{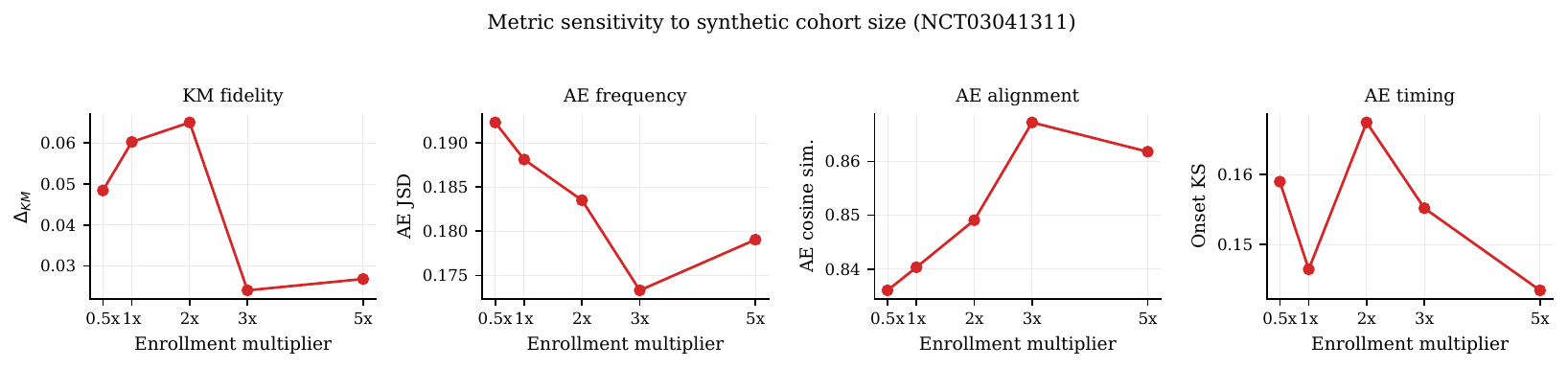}
\caption{Metric sensitivity to synthetic cohort size. AE frequency metrics improve with sample size; KM fidelity and onset timing are stable, exhibiting bootstrap-level variance.}
\label{fig:ablation-ss}
\end{figure}

Figure~\ref{fig:ablation-ss} sweeps the enrollment multiplier from 0.5$\times$ to 5$\times$. AE JSD decreases from 0.192 to 0.173 and cosine similarity increases from 0.84 to 0.86 as sample size grows, confirming that frequency-based metrics benefit from larger synthetic cohorts. $\Delta_{\text{KM}}$ and onset KS show bootstrap-level variance across multipliers rather than a monotone trend, indicating that metrics are generally stable above 1$\times$, supporting the 2$\times$ default.

\clearpage

\section{Additional Analyses}
\label{app:additional}

This appendix reports two analyses referenced from the main text: an isolation of how much KM-extraction error propagates into the generated cohort (Table~\ref{tab:propagation}), and a per-trial recalibration of the AE-burden channel through the propensity dispersion $\sigma_\eta$ (Tables~\ref{tab:sigma-sweep} and~\ref{tab:results-sigma08}). Both use the same evaluation code path, seeds, and $2{\times}$ enrollment as Table~\ref{tab:results}.

\subsection{Extraction-Error Propagation}
\label{app:propagation}

The survival channel is the only one traced from a plot image, so its contribution to end-to-end error can be isolated by swapping the input curve while holding everything else fixed. For the one arm per trial that matches a held-out real-IPD arm, we regenerate 30 cohorts (seeds $0,\dots,29$) from the extracted published KM, and 30 more from an \emph{oracle} KM reconstructed directly from the held-out IPD. The difference in $\Delta_{\text{KM}}$ is the propagated extraction error.

\begin{table}[ht]
\centering
\small
\caption{Extraction-error propagation (EXP1). \emph{Extracted} regenerates from the curve traced off the published figure; \emph{oracle} regenerates from the KM reconstructed from held-out real IPD. Propagated $\Delta$ is the difference in mean $\Delta_{\text{KM}}$. Values are mean~$\pm$~std over 30 regenerations. Unlike Table~\ref{tab:results}, which pools all arms of a trial, these numbers are for the single matched arm, so the extracted-input column is not identical to the headline $\Delta_{\text{KM}}$.}
\label{tab:propagation}
\resizebox{\linewidth}{!}{%
\begin{tabular}{lcccccc}
\toprule
 & & \multicolumn{2}{c}{$\Delta_{\text{KM}}$ $\downarrow$} & \multicolumn{2}{c}{OS KS $\downarrow$} & \\
\cmidrule(lr){3-4}\cmidrule(lr){5-6}
Trial (arm, $n_r$) & Extraction IAE & Extracted & Oracle & Extracted & Oracle & Propagated $\Delta$ \\
\midrule
NCT03041311 (Placebo, 53)     & 0.0309 & 0.055 $\pm$ 0.018 & 0.040 $\pm$ 0.022 & 0.238 $\pm$ 0.065 & 0.114 $\pm$ 0.035 & 0.015 \\
NCT02499770 (E/P, 37)         & 0.0345 & 0.056 $\pm$ 0.018 & 0.049 $\pm$ 0.018 & 0.156 $\pm$ 0.035 & 0.147 $\pm$ 0.035 & 0.007 \\
NCT00844649 (Gemcitabine, 430)& 0.0133 & 0.018 $\pm$ 0.005 & 0.012 $\pm$ 0.006 & 0.063 $\pm$ 0.014 & 0.043 $\pm$ 0.014 & 0.006 \\
\bottomrule
\end{tabular}%
}
\end{table}

The propagated component is $0.006$--$0.015$, i.e.\ at most about half the extraction IAE and 12--32\% of the total $\Delta_{\text{KM}}$; the sampler does not amplify upstream error. The largest propagated value occurs on the smallest cohort, consistent with the reference KM at $n_r{\sim}50$ being itself a noisy estimator. Oracle-input $\Delta_{\text{KM}}$ is bounded away from zero on every trial, which is the finite-sample floor: even a perfect curve yields a bootstrap-resampled cohort that differs from the reference realization.

\subsection{AE-Burden Recalibration via $\sigma_\eta$}
\label{app:sigma-sweep}

Section~\ref{sec:ipd_evaluation} identifies AE burden as the largest quantitative gap. Because survival is assigned by a permutation of bootstrap draws that is independent of AE sampling (Section~\ref{sec:survival}), the burden channel can be recalibrated without disturbing survival. Table~\ref{tab:sigma-sweep} re-runs the full protocol at four values of $\sigma_\eta$ on all three trials.

\begin{table}[ht]
\centering
\small
\caption{Propensity dispersion sweep (EXP2), 30 regenerations per cell. Burden approaches the real value monotonically in $\sigma_\eta$ while $\Delta_{\text{KM}}$ is invariant (identical to four decimal places, as survival assignment does not depend on $\sigma_\eta$) and AE frequency moves within noise. Best burden KS per trial in bold.}
\label{tab:sigma-sweep}
\resizebox{\linewidth}{!}{%
\begin{tabular}{llcccc}
\toprule
Trial & Metric & $\sigma_\eta{=}0.4$ & $\sigma_\eta{=}0.6$ & $\sigma_\eta{=}0.8$ & $\sigma_\eta{=}1.0$ \\
\midrule
\multirow{4}{*}{NCT03041311}
 & AEs/patient (real: 15.1) & 10.1 $\pm$ 0.5 & 11.3 $\pm$ 0.8 & 13.0 $\pm$ 1.5 & 17.2 $\pm$ 2.3 \\
 & Burden KS $\downarrow$ & 0.277 $\pm$ 0.029 & 0.218 $\pm$ 0.047 & 0.190 $\pm$ 0.045 & \textbf{0.165 $\pm$ 0.041} \\
 & $\Delta_{\text{KM}}$ $\downarrow$ & 0.049 & 0.049 & 0.049 & 0.049 \\
 & AE cosine $\uparrow$ & 0.852 & 0.850 & 0.852 & 0.861 \\
\midrule
\multirow{4}{*}{NCT02499770}
 & AEs/patient (real: 12.8) & 6.9 $\pm$ 0.6 & 7.7 $\pm$ 0.6 & 9.5 $\pm$ 1.2 & 10.5 $\pm$ 1.2 \\
 & Burden KS $\downarrow$ & 0.457 $\pm$ 0.057 & 0.407 $\pm$ 0.044 & \textbf{0.344 $\pm$ 0.057} & 0.352 $\pm$ 0.057 \\
 & $\Delta_{\text{KM}}$ $\downarrow$ & 0.051 & 0.051 & 0.051 & 0.051 \\
 & AE cosine $\uparrow$ & 0.842 & 0.841 & 0.844 & 0.846 \\
\midrule
\multirow{4}{*}{NCT00844649}
 & AEs/patient (real: 15.4) & 11.4 $\pm$ 0.2 & 12.6 $\pm$ 0.4 & 14.8 $\pm$ 0.5 & 17.9 $\pm$ 0.9 \\
 & Burden KS $\downarrow$ & 0.185 $\pm$ 0.010 & 0.122 $\pm$ 0.014 & 0.062 $\pm$ 0.015 & \textbf{0.060 $\pm$ 0.011} \\
 & $\Delta_{\text{KM}}$ $\downarrow$ & 0.040 & 0.040 & 0.040 & 0.040 \\
 & AE cosine $\uparrow$ & 0.940 & 0.940 & 0.940 & 0.938 \\
\bottomrule
\end{tabular}%
}
\end{table}

Burden KS is minimized at $\sigma_\eta{=}1.0$ on two of the three trials, but we do not adopt that value: the mean burden \emph{overshoots} real burden there ($17.2$ vs.\ $15.1$; $17.9$ vs.\ $15.4$), and the KS improvement over $\sigma_\eta{=}0.8$ is within one standard deviation on both. At $\sigma_\eta{=}0.8$ the synthetic mean approaches real burden from below on all three trials ($13.0$ vs.\ $15.1$; $9.5$ vs.\ $12.8$; $14.8$ vs.\ $15.4$) while improving burden KS by 31\%, 25\%, and 66\% relative to the default. NCT02499770 is the exception in degree rather than direction --- it is the smallest cohort and the one whose registry AE table is most heavily truncated by the 5\% reporting threshold, so a dispersion parameter alone cannot recover its missing tail; $\sigma_\eta{=}1.0$ brings its mean closer ($10.5$) but its KS slightly worse ($0.352$ vs.\ $0.344$). We therefore recommend $\sigma_\eta{=}0.8$ rather than the KS-minimizing value, and report the complete Stage-B table at that setting in Table~\ref{tab:results-sigma08}. Across all four columns $\Delta_{\text{KM}}$ is invariant and the AE-frequency metrics move by less than $0.01$, which is the behavior the decoupled design predicts: the burden channel is tunable in isolation.

\begin{table}[ht]
\centering
\small
\setlength{\tabcolsep}{4pt}
\caption{Full Stage-B fidelity at the recommended $\sigma_\eta{=}0.8$ (compare Table~\ref{tab:results}, which uses $\sigma_\eta{=}0.4$). Mean~$\pm$~std over 30 independent regenerations, synthetic cohorts at $2{\times}$ real enrollment. Only the burden rows change materially; demographic, survival, and AE-frequency rows are within resampling noise of the $\sigma_\eta{=}0.4$ values.}
\label{tab:results-sigma08}
\resizebox{\linewidth}{!}{%
\begin{tabular}{lccc}
\toprule
 & \textbf{NCT03041311} & \textbf{NCT02499770} & \textbf{NCT00844649} \\
 & \scriptsize{$n_r{=}53,\, n_s{=}107$} & \scriptsize{$n_r{=}37,\, n_s{=}75$} & \scriptsize{$n_r{=}430,\, n_s{=}861$} \\
\midrule
\multicolumn{4}{l}{\textit{Demographics}} \\
\quad Sex JSD $\downarrow$                    & 0.0054 $\pm$ 0.0064 & 0.0033 $\pm$ 0.0060 & 0.0005 $\pm$ 0.0006 \\
\quad ECOG JSD $\downarrow$                   & 0.272 $\pm$ 0.048   & 0.013 $\pm$ 0.012   & 0.003 $\pm$ 0.001   \\
\quad Weight KS $\downarrow$                  & 0.204 $\pm$ 0.039   & 0.200 $\pm$ 0.056   & 0.125 $\pm$ 0.017   \\
\quad Height KS $\downarrow$                  & 0.207 $\pm$ 0.029   & 0.179 $\pm$ 0.042   & 0.103 $\pm$ 0.013   \\
\quad BMI KS $\downarrow$                     & 0.171 $\pm$ 0.038   & 0.191 $\pm$ 0.030   & 0.151 $\pm$ 0.016   \\
\midrule
\multicolumn{4}{l}{\textit{Overall survival}} \\
\quad Median OS $|\Delta|$ (mo.) $\downarrow$ & 0.84 $\pm$ 0.76     & 1.12 $\pm$ 0.83     & 1.28 $\pm$ 0.34     \\
\quad Median OS rel.\ error $\downarrow$      & $7.0 \pm 6.4$\%     & $11.3 \pm 8.4$\%    & $19.4 \pm 5.1$\%    \\
\quad OS KS $\downarrow$                      & 0.173 $\pm$ 0.055   & 0.156 $\pm$ 0.039   & 0.112 $\pm$ 0.018   \\
\quad $\Delta_{\text{KM}}$ $\downarrow$       & 0.049 $\pm$ 0.018   & 0.051 $\pm$ 0.018   & 0.040 $\pm$ 0.011   \\
\quad Mann--Whitney $U$ $p$                   & 0.63 $\pm$ 0.29     & 0.60 $\pm$ 0.28     & 0.008 $\pm$ 0.031   \\
\midrule
\multicolumn{4}{l}{\textit{Adverse event frequency}} \\
\quad AE JSD $\downarrow$                     & 0.185 $\pm$ 0.006   & 0.204 $\pm$ 0.011   & 0.142 $\pm$ 0.002   \\
\quad AE cosine similarity $\uparrow$         & 0.852 $\pm$ 0.015   & 0.844 $\pm$ 0.030   & \textbf{0.940 $\pm$ 0.004} \\
\quad Top-15 overlap $\uparrow$               & 11.3 $\pm$ 0.9 / 15 & 10.8 $\pm$ 1.0 / 15 & \textbf{12.0 $\pm$ 0.0 / 15} \\
\quad Mean per-AE $|\Delta|$ $\downarrow$     & 0.112 $\pm$ 0.007   & 0.111 $\pm$ 0.010   & \textbf{0.030 $\pm$ 0.001} \\
\midrule
\multicolumn{4}{l}{\textit{AE timing and burden}} \\
\quad Mean onset $|\Delta|$ (mo.) $\downarrow$& 0.72 $\pm$ 0.13     & 0.27 $\pm$ 0.12     & 0.77 $\pm$ 0.06     \\
\quad Onset KS $\downarrow$                   & 0.155 $\pm$ 0.018   & 0.269 $\pm$ 0.014   & 0.253 $\pm$ 0.004   \\
\quad AE burden (real $\to$ synth)            & 15.1 $\to$ \textbf{13.0 $\pm$ 1.5} & 12.8 $\to$ \textbf{9.5 $\pm$ 1.2} & 15.4 $\to$ \textbf{14.8 $\pm$ 0.5} \\
\quad Burden KS $\downarrow$                  & \textbf{0.190 $\pm$ 0.045} & \textbf{0.344 $\pm$ 0.057} & \textbf{0.062 $\pm$ 0.015} \\
\bottomrule
\end{tabular}%
}
\end{table}

\clearpage

\section{Supplementary Figures}
\label{app:supplementary}

\begin{figure}[ht]
\centering
\includegraphics[width=\linewidth]{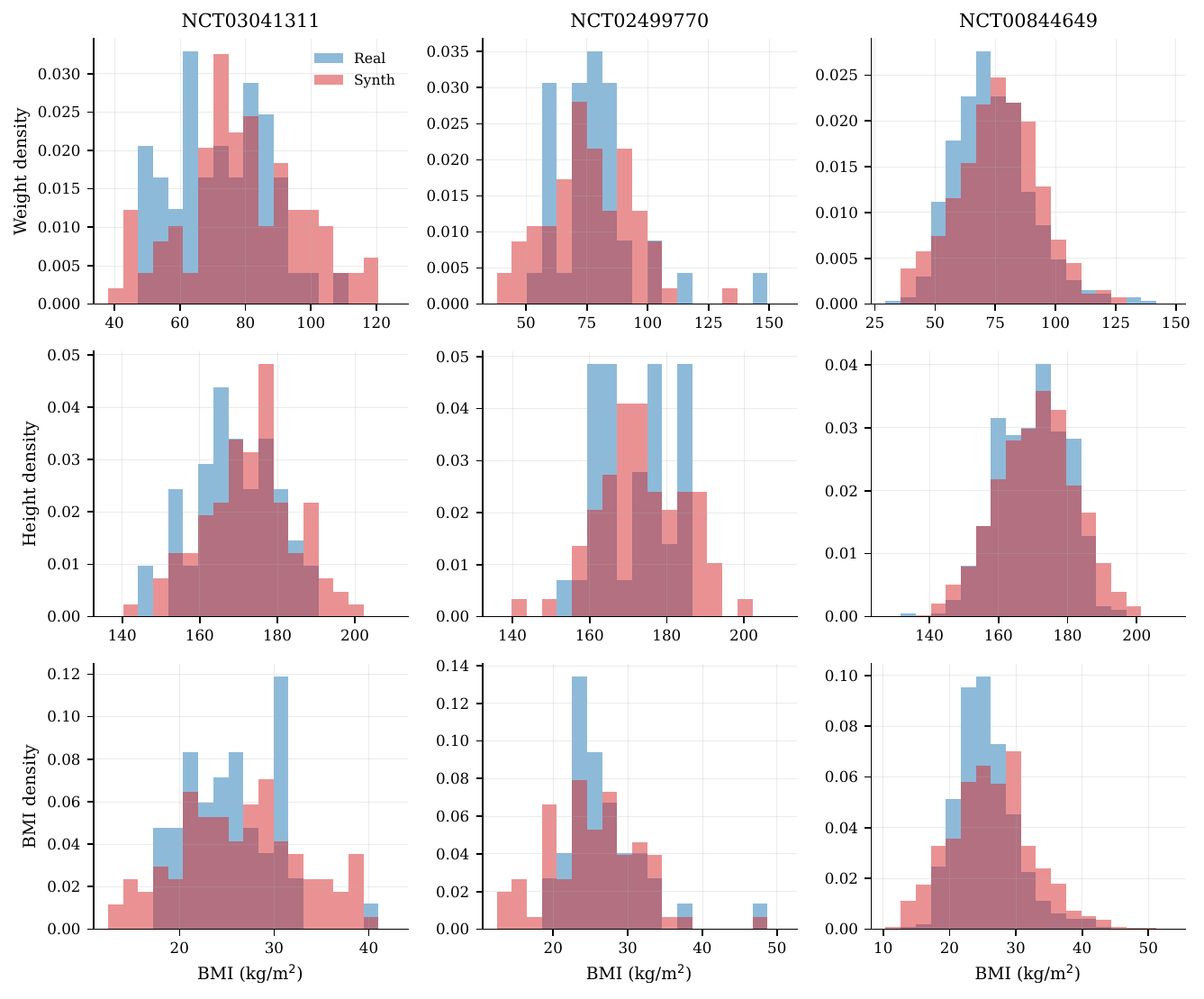}
\caption{Weight, height, and BMI distributions (real vs.\ synthetic) for all three trials. KS statistics are ${\leq}0.24$ in all cases; the truncated-Gaussian sampling reproduces the overall shape of real anthropometric distributions.}
\label{fig:supp-demo}
\end{figure}

\begin{figure}[ht]
\centering
\includegraphics[width=\linewidth]{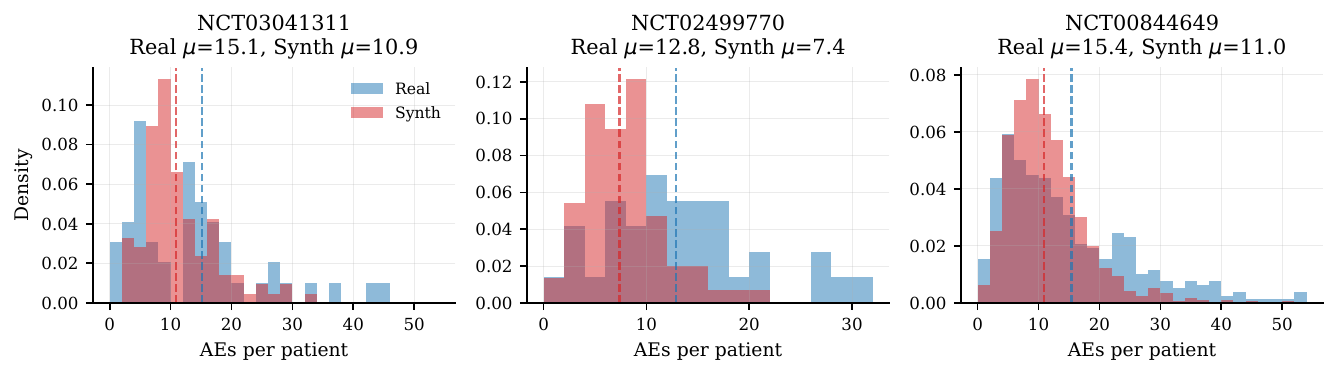}
\caption{AE burden per patient. Dashed lines show means.}
\label{fig:supp-burden}
\end{figure}
\clearpage

\section{Full Extraction Gallery}
\label{app:km-gallery}

This appendix shows a paired extraction diagnostic for every plot in
the 32-plot benchmark (Section~\ref{sec:km-benchmark}). Each figure
has two panels. The \emph{left panel} is the source KM plot with the
pipeline's extracted step coordinates overlaid as small rings
(\texttt{annotate\_image()} output); visible drift between rings and
curve indicates extraction error, aligned rings indicate a faithful
trace. The \emph{right panel} overlays the ground-truth step function
(solid) and the extracted step function (dashed) on a common axis, with
the absolute-difference region shaded to make the IAE integrand
visually legible. The IAE is printed in the right-panel title.

Captions report IAE, median AE, median OS error (undefined when the
curve does not cross $S = 0.5$ within the shared time range), and arm
count. Per-plot observations comment on what succeeded or failed
visibly in each extraction. Plots within each category are ordered by
ascending IAE.

Source plots were selected for their uniquely challenging KM plot features, such as overlapping shading from the wide confidence intervals of multiple lines, closely clustered lines with partial overlap and multiple crossovers, lines with dense ticks for adverse events, overlapping colored lines producing new colors, and overlapping dotted and solid lines resulting in a consistently solid line during the overlap.

Per-plot numeric details are collected in Table~\ref{tab:km-perplot}.

\begin{table}[ht]
\centering
\small
\setlength{\tabcolsep}{4pt}
\caption{Per-plot attempt-1 extraction metrics, sorted by IAE (best first).}
\label{tab:km-perplot}
\resizebox{\linewidth}{!}{%
\begin{tabular}{@{}lllrrrr@{}}
\toprule
\# & Plot & Category & IAE & Med.\ AE & Med.\ OS err (mo) & Arms \\
\midrule
1 & \texttt{stress\_combo\_flat\_overlap} & Combo stress & 0.0012 & 0.0005 & 0.000 & 2 \\
2 & \texttt{edge\_high\_survival} & Edge & 0.0012 & 0.0007 & 0.000 & 2 \\
3 & \texttt{edge\_near\_flat} & Edge & 0.0029 & 0.0016 & 0.032 & 2 \\
4 & \texttt{synthetic\_002} & Standard & 0.0033 & 0.0016 & 0.095 & 2 \\
5 & \texttt{edge\_ci\_shading} & Edge & 0.0035 & 0.0019 & 0.105 & 2 \\
6 & \texttt{synthetic\_003} & Standard & 0.0040 & 0.0017 & 0.232 & 2 \\
7 & \texttt{stress\_gridlines} & Single stress & 0.0051 & 0.0029 & 0.244 & 2 \\
8 & \texttt{synthetic\_001} & Standard & 0.0051 & 0.0028 & 0.104 & 2 \\
9 & \texttt{stress\_annotation\_heavy} & Single stress & 0.0052 & 0.0029 & 0.250 & 2 \\
10 & \texttt{stress\_three\_similar} & Single stress & 0.0060 & 0.0005 & 0.048 & 3 \\
11 & \texttt{synthetic\_004} & Standard & 0.0062 & 0.0042 & 0.065 & 2 \\
12 & \texttt{stress\_multi\_panel\_clean} & Single stress & 0.0064 & --- & --- & 4 \\
13 & \texttt{synthetic\_005} & Standard & 0.0068 & 0.0024 & 0.046 & 2 \\
14 & \texttt{edge\_four\_arms} & Edge & 0.0070 & 0.0054 & 0.123 & 4 \\
15 & \texttt{stress\_jpeg\_artifact} & Single stress & 0.0074 & 0.0052 & 0.269 & 2 \\
16 & \texttt{stress\_combo\_bw\_overlap} & Combo stress & 0.0075 & 0.0037 & 0.100 & 2 \\
17 & \texttt{stress\_legend\_overlap} & Single stress & 0.0085 & 0.0030 & 0.382 & 2 \\
18 & \texttt{stress\_bw} & Single stress & 0.0088 & 0.0054 & 0.290 & 2 \\
19 & \texttt{edge\_small\_dense} & Edge & 0.0096 & 0.0096 & 0.219 & 2 \\
20 & \texttt{edge\_cumulative\_incidence} & Edge & 0.0145 & 0.0102 & 0.000 & 3 \\
21 & \texttt{stress\_tiny} & Single stress & 0.0149 & 0.0129 & 0.210 & 2 \\
22 & \texttt{stress\_stretched\_tall} & Single stress & 0.0149 & 0.0135 & 0.285 & 2 \\
23 & \texttt{stress\_blurry} & Single stress & 0.0151 & 0.0134 & 0.278 & 2 \\
24 & \texttt{stress\_combo\_4arm\_tiny} & Combo stress & 0.0175 & 0.0183 & 0.407 & 4 \\
25 & \texttt{stress\_combo\_tiny\_bw} & Combo stress & 0.0198 & 0.0161 & 0.231 & 2 \\
26 & \texttt{stress\_stretched\_wide} & Single stress & 0.0215 & 0.0187 & 0.537 & 2 \\
27 & \texttt{edge\_multi\_panel} & Edge & 0.0237 & --- & --- & 4 \\
28 & \texttt{stress\_combo\_tiny\_blurry} & Combo stress & 0.0271 & 0.0212 & 0.200 & 2 \\
29 & \texttt{stress\_combo\_stretched\_dense} & Combo stress & 0.0378 & 0.0375 & 1.703 & 2 \\
30 & \texttt{stress\_combo\_jpeg\_blurry\_dark} & Combo stress & 0.0412 & 0.0256 & 1.071 & 2 \\
31 & \texttt{stress\_diep\_like} & Single stress & 0.0540 & --- & --- & 8 \\
32 & \texttt{stress\_combo\_grid\_annotation\_bw} & Combo stress & 0.0744 & 0.0058 & 4.888 & 2 \\
\bottomrule
\end{tabular}%
}
\end{table}
\clearpage

\subsection{Standard Synthetic (5)}
\begin{figure}[htb]
\centering
\begin{minipage}[t]{\linewidth}
\centering
\includegraphics[width=\linewidth]{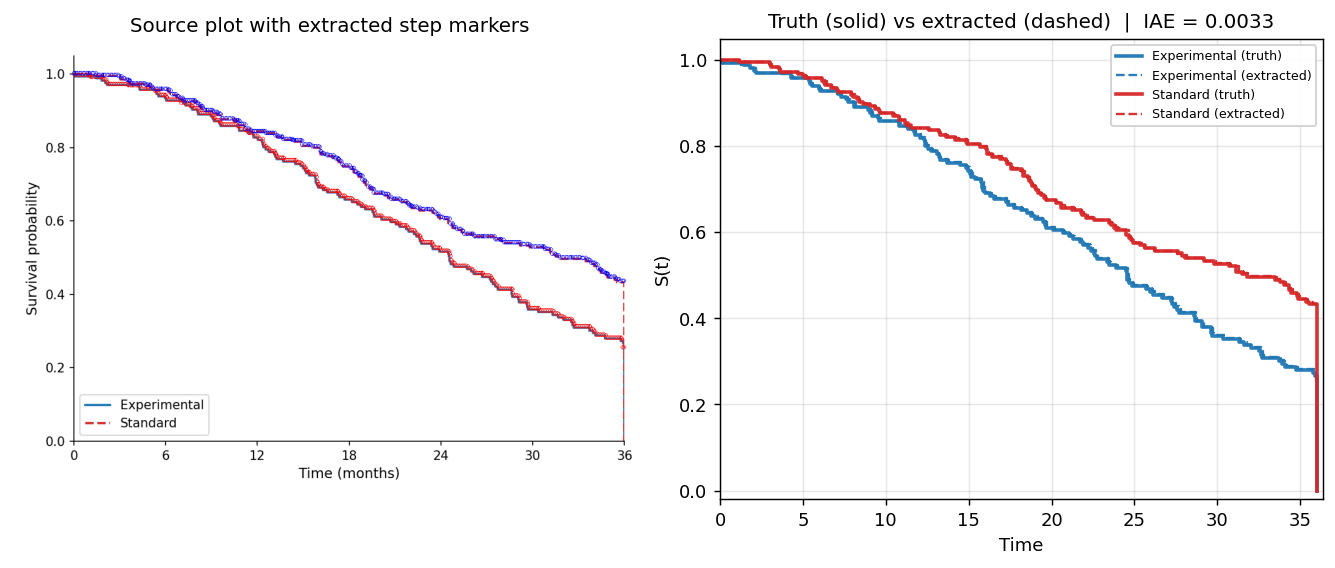}\\
\small \texttt{synthetic\_002} (IAE=0.0033, Med.\ AE=0.0016, Med.\ OS err=0.10~mo, 2 arms)
\end{minipage}
\hfill
\begin{minipage}[t]{\linewidth}
\centering
\includegraphics[width=\linewidth]{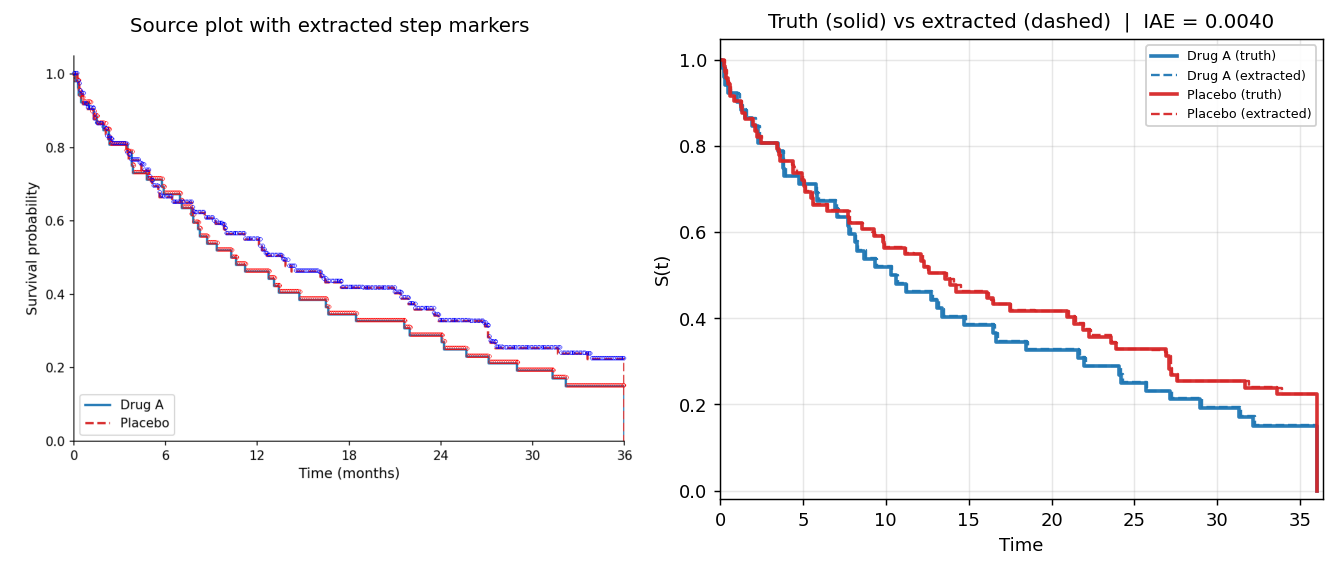}\\
\small \texttt{synthetic\_003} (IAE=0.0040, Med.\ AE=0.0017, Med.\ OS err=0.23~mo, 2 arms)
\end{minipage}
\caption{\textbf{synthetic\_002}: Standard 2-arm plot over 36~mo. Clean color separation; steps tracked precisely. \quad \textbf{synthetic\_003}: Standard 2-arm plot over 36~mo with tighter step spacing.}
\end{figure}
\begin{figure}[htb]
\centering
\begin{minipage}[t]{\linewidth}
\centering
\includegraphics[width=\linewidth]{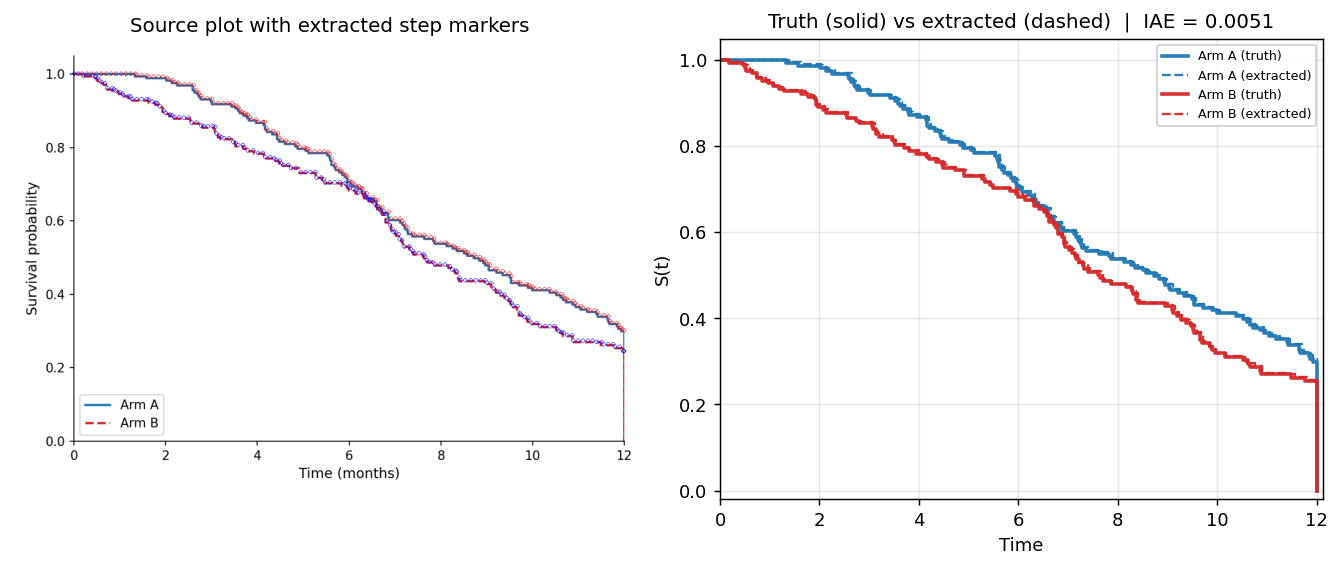}\\
\small \texttt{synthetic\_001} (IAE=0.0051, Med.\ AE=0.0028, Med.\ OS err=0.10~mo, 2 arms)
\end{minipage}
\hfill
\begin{minipage}[t]{\linewidth}
\centering
\includegraphics[width=\linewidth]{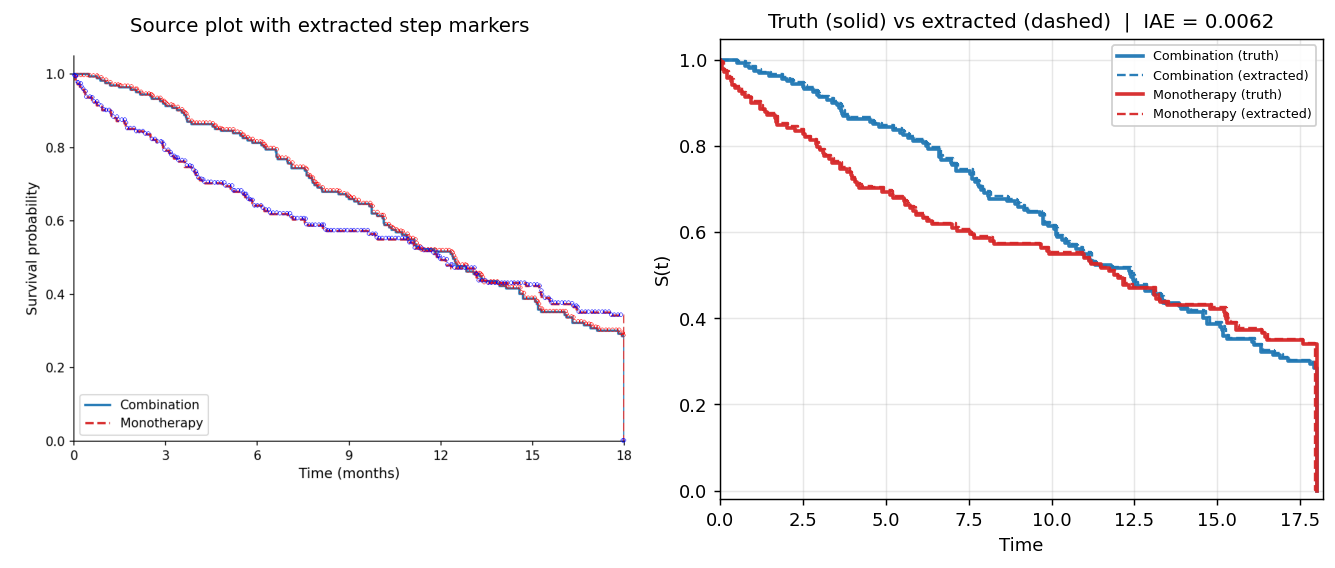}\\
\small \texttt{synthetic\_004} (IAE=0.0062, Med.\ AE=0.0042, Med.\ OS err=0.06~mo, 2 arms)
\end{minipage}
\caption{\textbf{synthetic\_001}: Two well-separated arms over 12~mo with clear step structure. Extraction markers match the steps visually with no systematic drift. \quad \textbf{synthetic\_004}: Standard 2-arm, 18~mo. Accurate step tracking.}
\end{figure}
\begin{figure}[htb]
\centering
\begin{minipage}[t]{\linewidth}
\centering
\includegraphics[width=\linewidth]{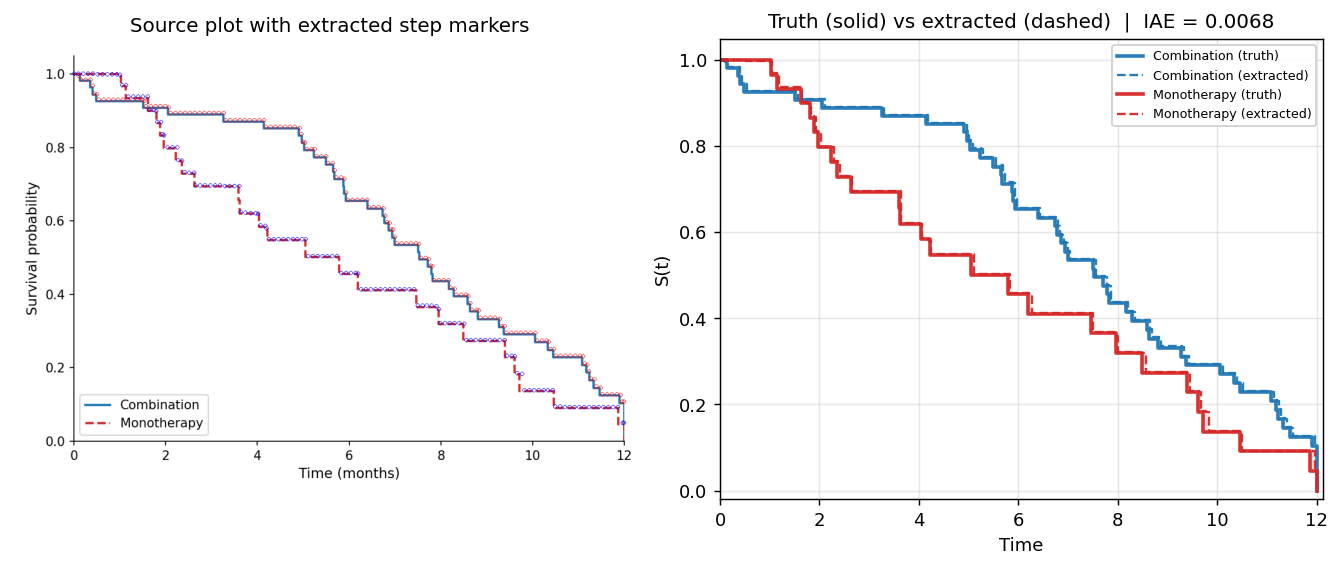}\\
\small \texttt{synthetic\_005} (IAE=0.0068, Med.\ AE=0.0024, Med.\ OS err=0.05~mo, 2 arms)
\end{minipage}
\caption{\textbf{synthetic\_005}: Standard 2-arm, 12~mo with steep drop. Vertical-drop detection picks up the final step.}
\end{figure}
\clearpage

\subsection{Edge Cases (7)}
\begin{figure}[htb]
\centering
\begin{minipage}[t]{\linewidth}
\centering
\includegraphics[width=\linewidth]{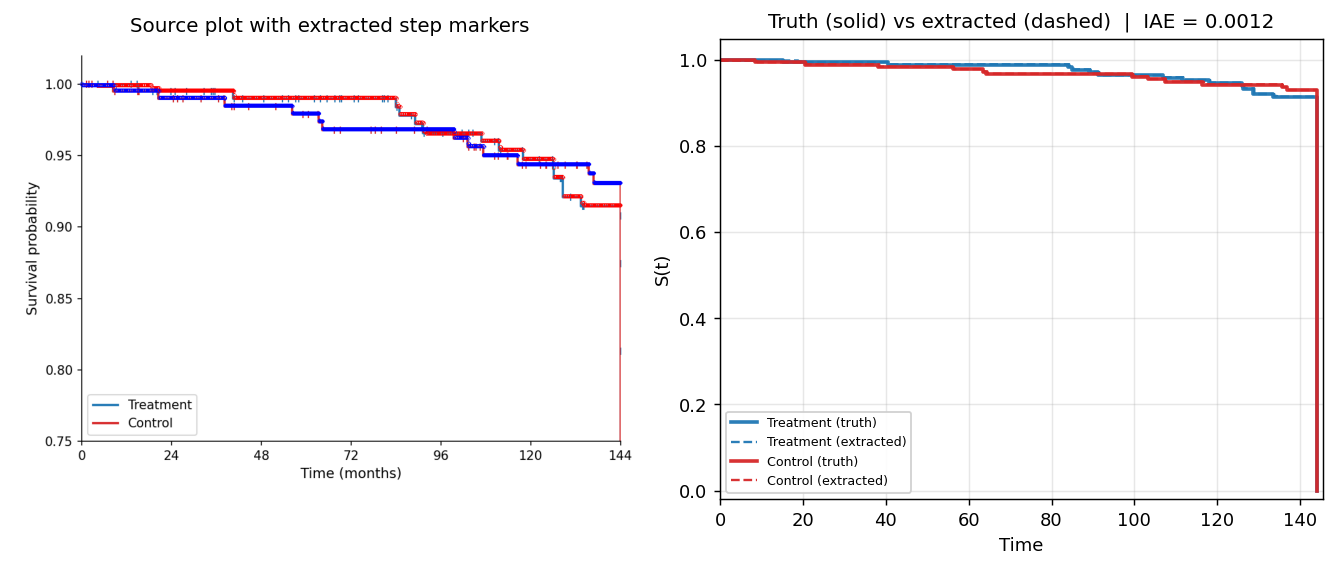}\\
\small \texttt{edge\_high\_survival} (IAE=0.0012, Med.\ AE=0.0007, Med.\ OS err=0.00~mo, 2 arms)
\end{minipage}
\hfill
\begin{minipage}[t]{\linewidth}
\centering
\includegraphics[width=\linewidth]{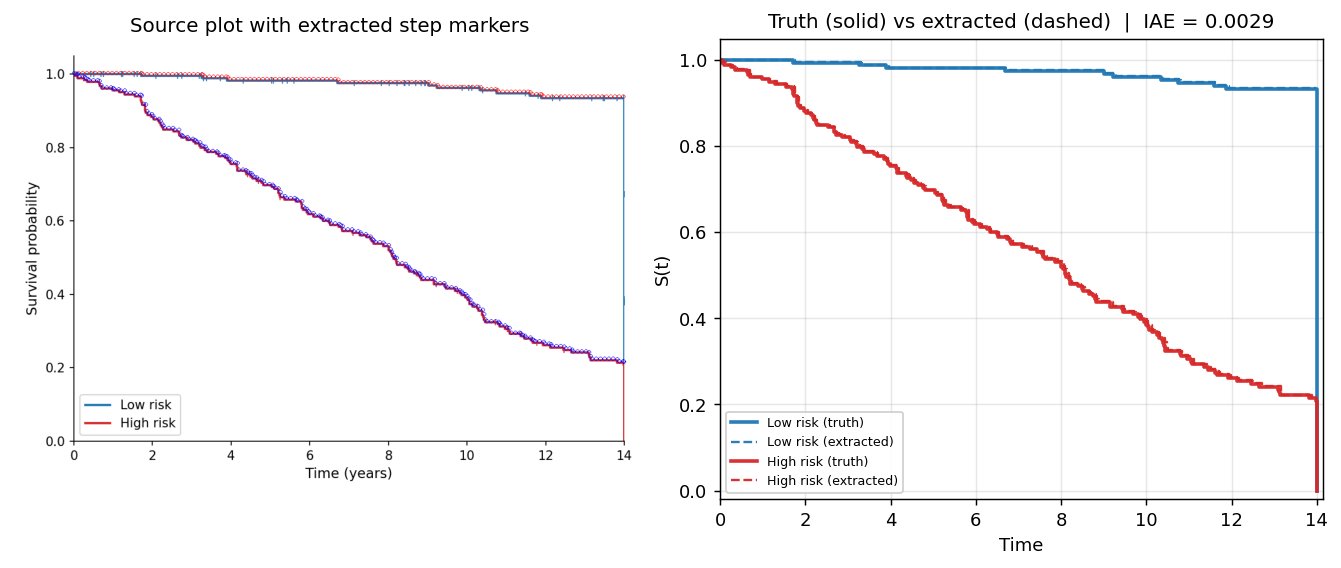}\\
\small \texttt{edge\_near\_flat} (IAE=0.0029, Med.\ AE=0.0016, Med.\ OS err=0.03~mo, 2 arms)
\end{minipage}
\caption{\textbf{edge\_high\_survival}: Minimal events over 144~mo follow-up. Red and blue ring markers track both arms cleanly through the entire range, including closely-overlapping segments near 96--120~mo. \quad \textbf{edge\_near\_flat}: Near-flat mortality profile. Ring density is high and follows the tight step structure closely.}
\end{figure}
\begin{figure}[htb]
\centering
\begin{minipage}[t]{\linewidth}
\centering
\includegraphics[width=\linewidth]{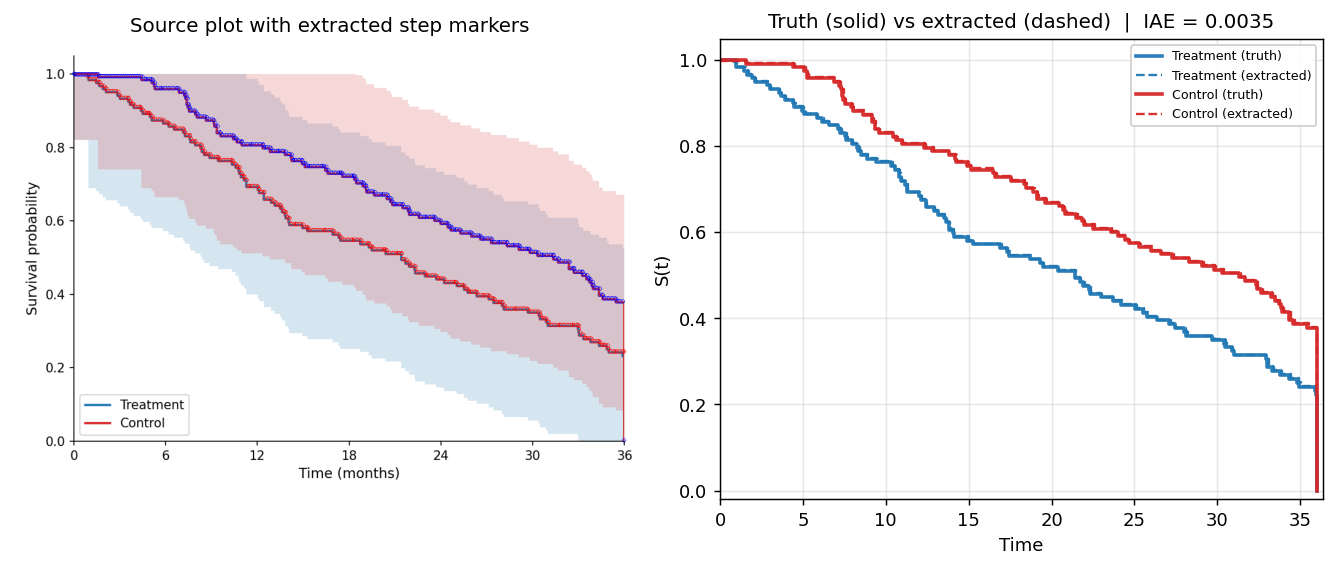}\\
\small \texttt{edge\_ci\_shading} (IAE=0.0035, Med.\ AE=0.0019, Med.\ OS err=0.10~mo, 2 arms)
\end{minipage}
\hfill
\begin{minipage}[t]{\linewidth}
\centering
\includegraphics[width=\linewidth]{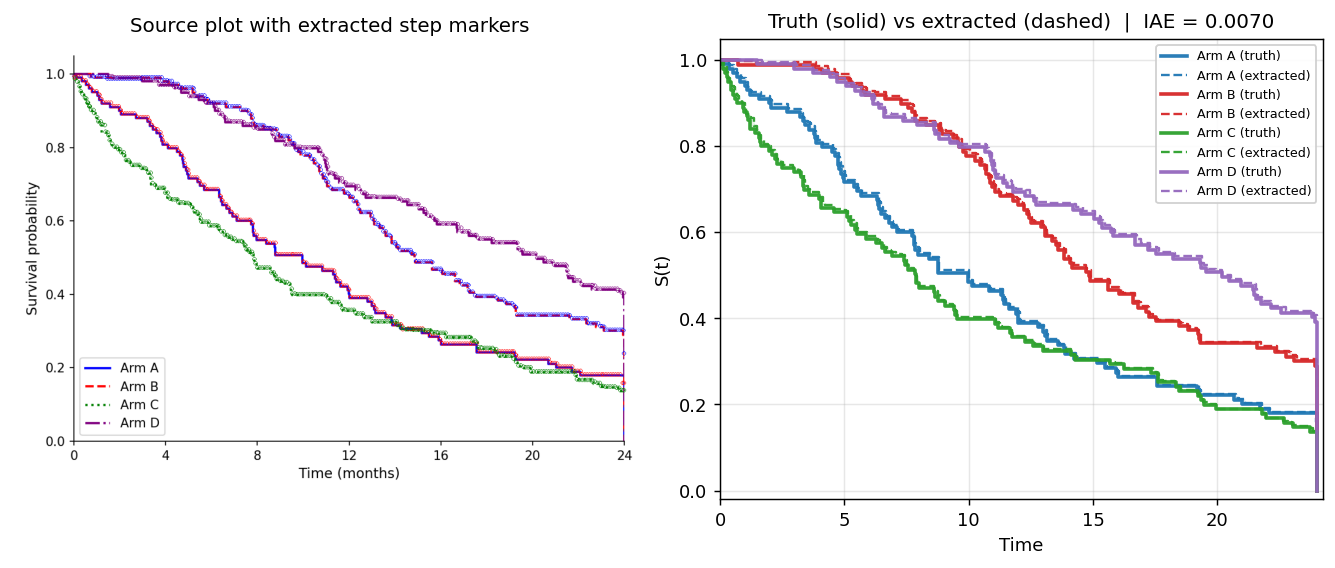}\\
\small \texttt{edge\_four\_arms} (IAE=0.0070, Med.\ AE=0.0054, Med.\ OS err=0.12~mo, 4 arms)
\end{minipage}
\caption{\textbf{edge\_ci\_shading}: Confidence-interval shading suppressed by the color mask; curves tracked cleanly. \quad \textbf{edge\_four\_arms}: 4-arm comparison; color separation and continuity tracking both hold.}
\end{figure}
\begin{figure}[htb]
\centering
\begin{minipage}[t]{\linewidth}
\centering
\includegraphics[width=\linewidth]{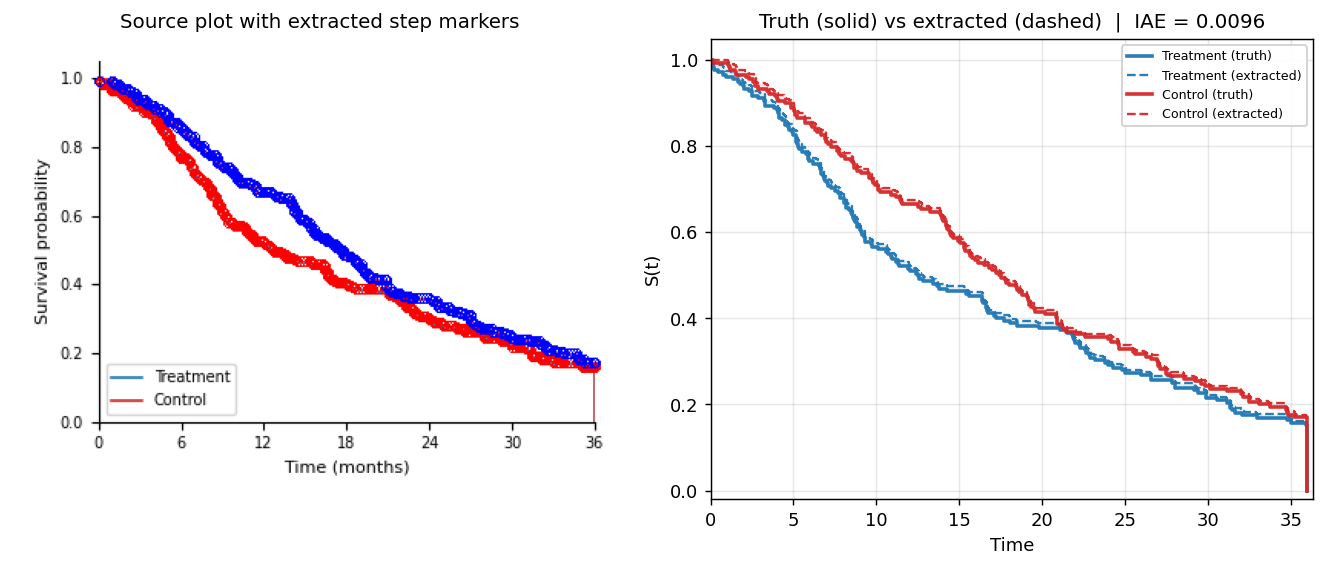}\\
\small \texttt{edge\_small\_dense} (IAE=0.0096, Med.\ AE=0.0096, Med.\ OS err=0.22~mo, 2 arms)
\end{minipage}
\hfill
\begin{minipage}[t]{\linewidth}
\centering
\includegraphics[width=\linewidth]{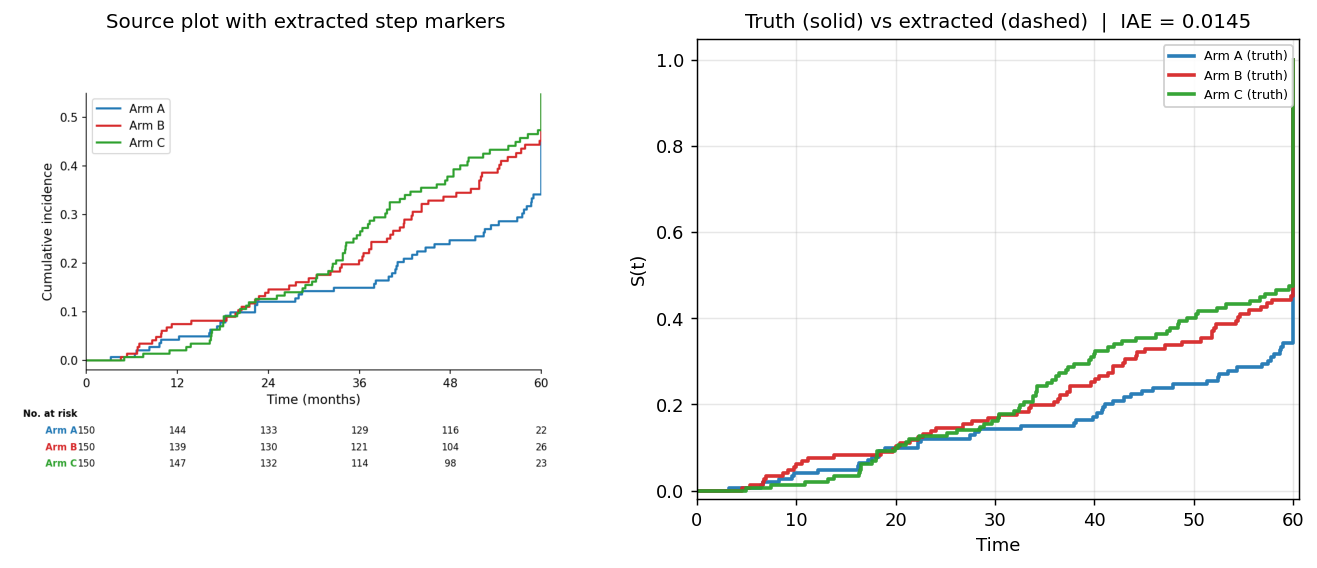}\\
\small \texttt{edge\_cumulative\_incidence} (IAE=0.0145, Med.\ AE=0.0102, Med.\ OS err=0.00~mo, 3 arms)
\end{minipage}
\caption{\textbf{edge\_small\_dense}: Dense step structure on a 36-month scale; fine sampling required. \quad \textbf{edge\_cumulative\_incidence}: Cumulative-incidence plot (curves rising). Monotonicity constraint inverted; rising structure recovered.}
\end{figure}
\begin{figure}[htb]
\centering
\begin{minipage}[t]{\linewidth}
\centering
\includegraphics[width=\linewidth]{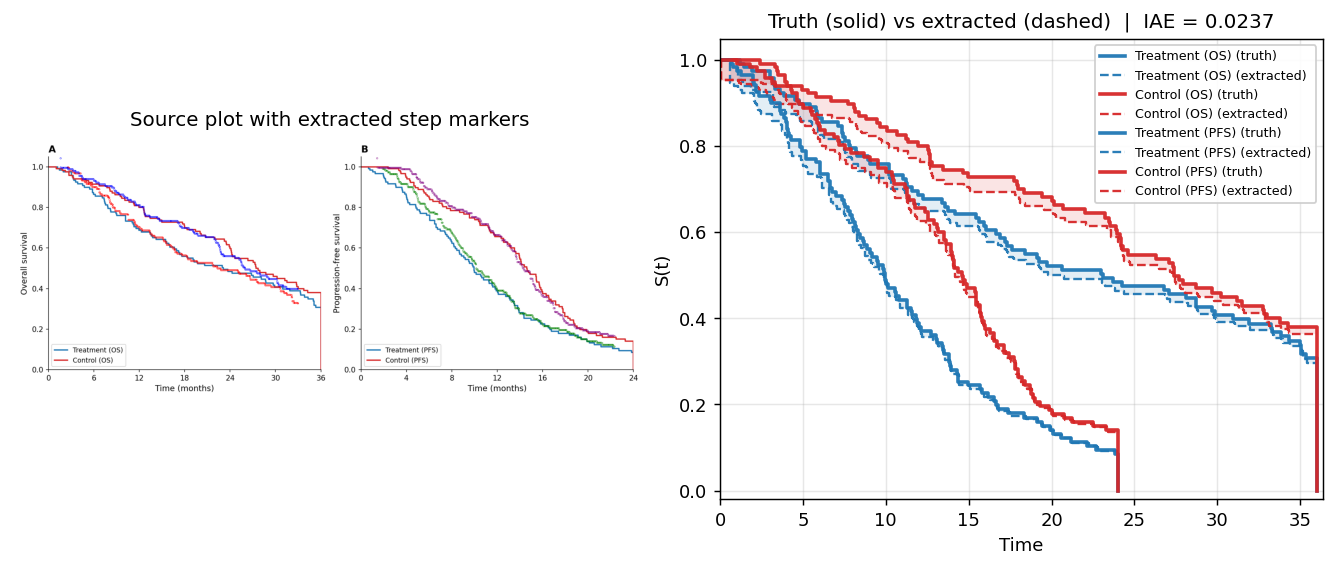}\\
\small \texttt{edge\_multi\_panel} (IAE=0.0237, Med.\ AE=---, Med.\ OS err=---~mo, 4 arms)
\end{minipage}
\caption{\textbf{edge\_multi\_panel}: Multi-panel layout (4 arms across subplots). Per-panel $x_\text{max}$ differs, inflating the aggregate IAE relative to what any single panel would show.}
\end{figure}
\clearpage

\subsection{Single-Degradation Stress Tests (12)}
\begin{figure}[htb]
\centering
\begin{minipage}[t]{\linewidth}
\centering
\includegraphics[width=\linewidth]{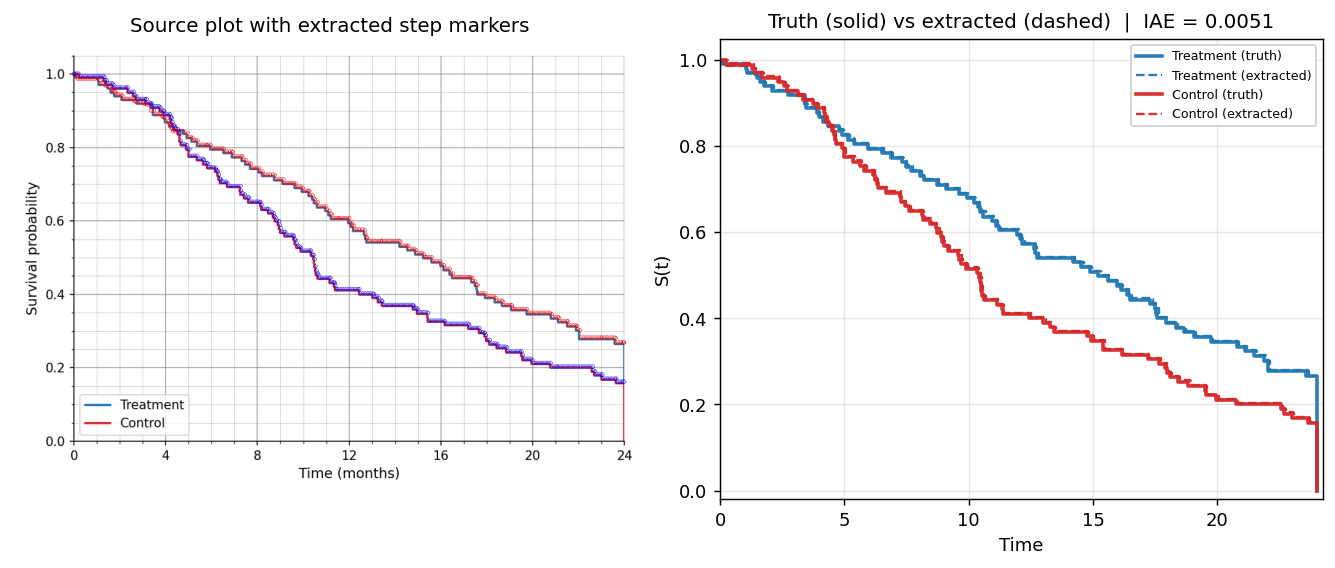}\\
\small \texttt{stress\_gridlines} (IAE=0.0051, Med.\ AE=0.0029, Med.\ OS err=0.24~mo, 2 arms)
\end{minipage}
\hfill
\begin{minipage}[t]{\linewidth}
\centering
\includegraphics[width=\linewidth]{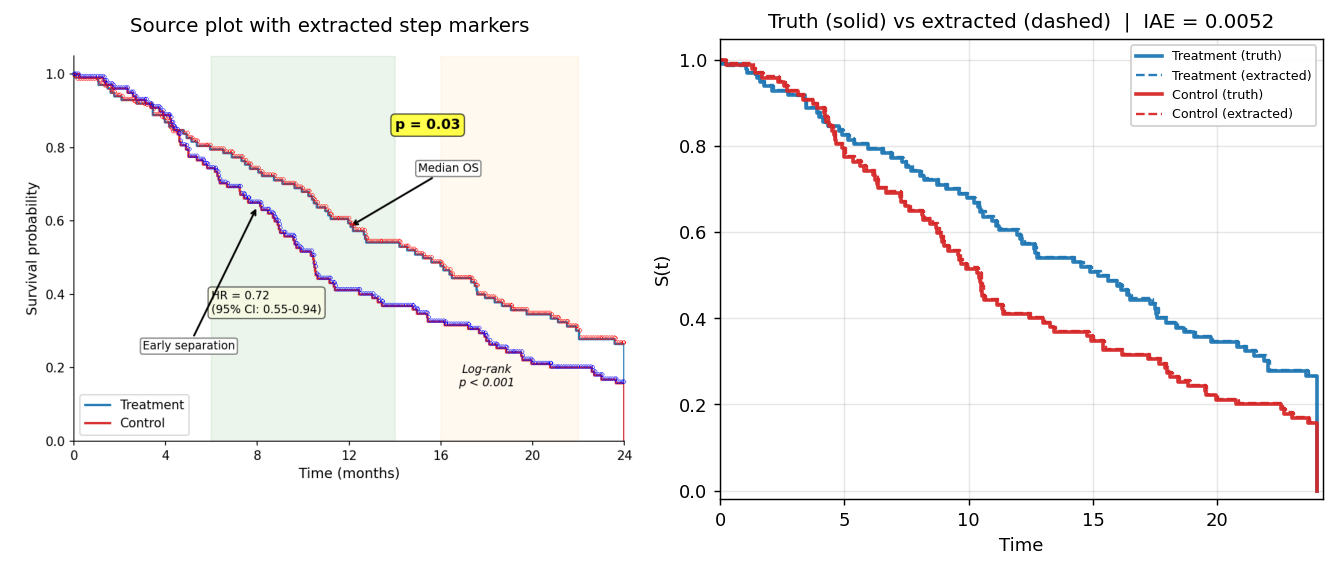}\\
\small \texttt{stress\_annotation\_heavy} (IAE=0.0052, Med.\ AE=0.0029, Med.\ OS err=0.25~mo, 2 arms)
\end{minipage}
\caption{\textbf{stress\_gridlines}: Light gridlines scrubbed by the $>$30\% row/column coverage rule before tracing. \quad \textbf{stress\_annotation\_heavy}: Multiple in-plot annotations masked before tracing.}
\end{figure}
\begin{figure}[htb]
\centering
\begin{minipage}[t]{\linewidth}
\centering
\includegraphics[width=\linewidth]{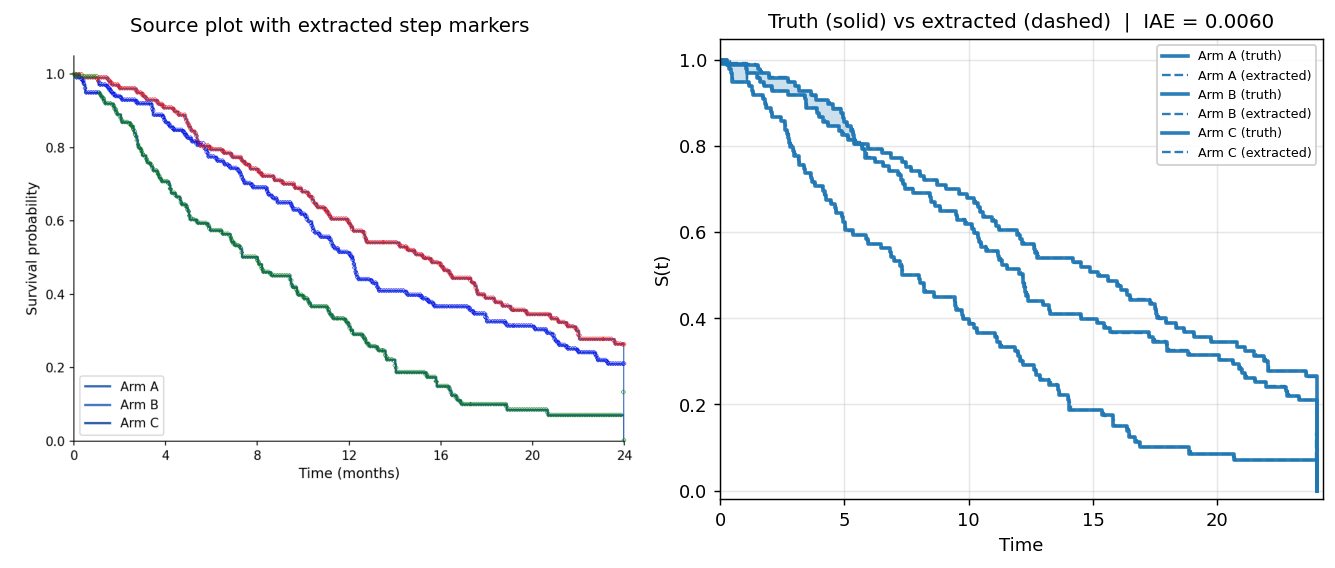}\\
\small \texttt{stress\_three\_similar} (IAE=0.0060, Med.\ AE=0.0005, Med.\ OS err=0.05~mo, 3 arms)
\end{minipage}
\hfill
\begin{minipage}[t]{\linewidth}
\centering
\includegraphics[width=\linewidth]{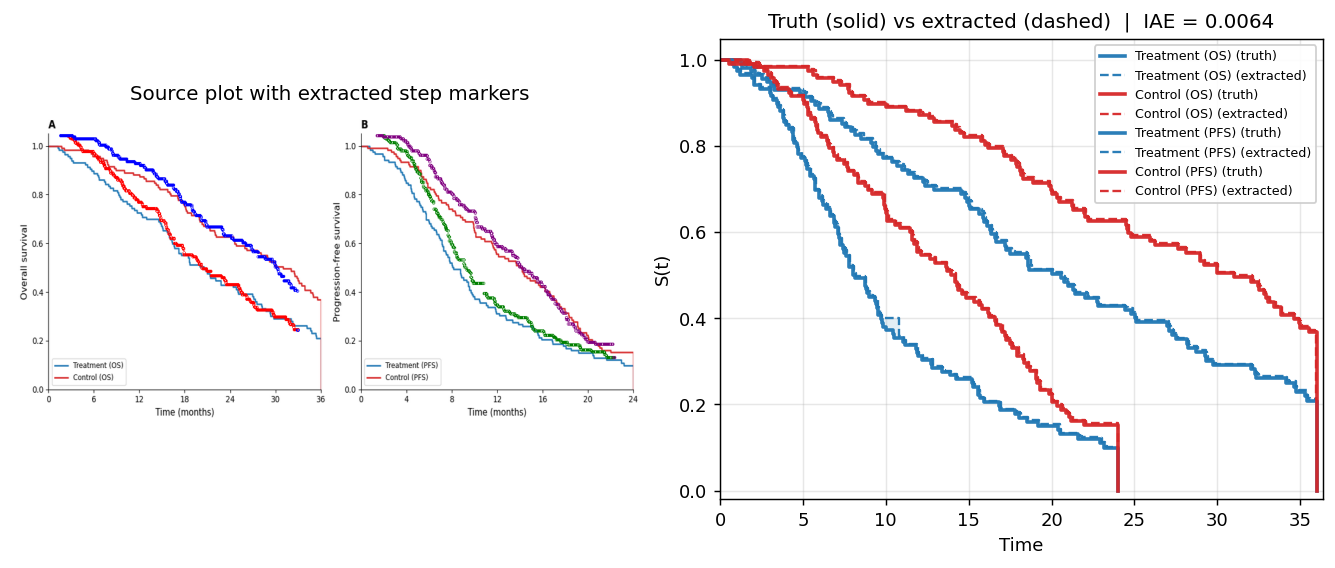}\\
\small \texttt{stress\_multi\_panel\_clean} (IAE=0.0064, Med.\ AE=---, Med.\ OS err=---~mo, 4 arms)
\end{minipage}
\caption{\textbf{stress\_three\_similar}: Three similarly-hued arms. HSL mask separates them with a narrow margin. \quad \textbf{stress\_multi\_panel\_clean}: Clean multi-panel layout; per-panel $x_\text{max}$ causes minor ($<0.002$) mismatch between saved and recomputed IAE.}
\end{figure}
\begin{figure}[htb]
\centering
\begin{minipage}[t]{\linewidth}
\centering
\includegraphics[width=\linewidth]{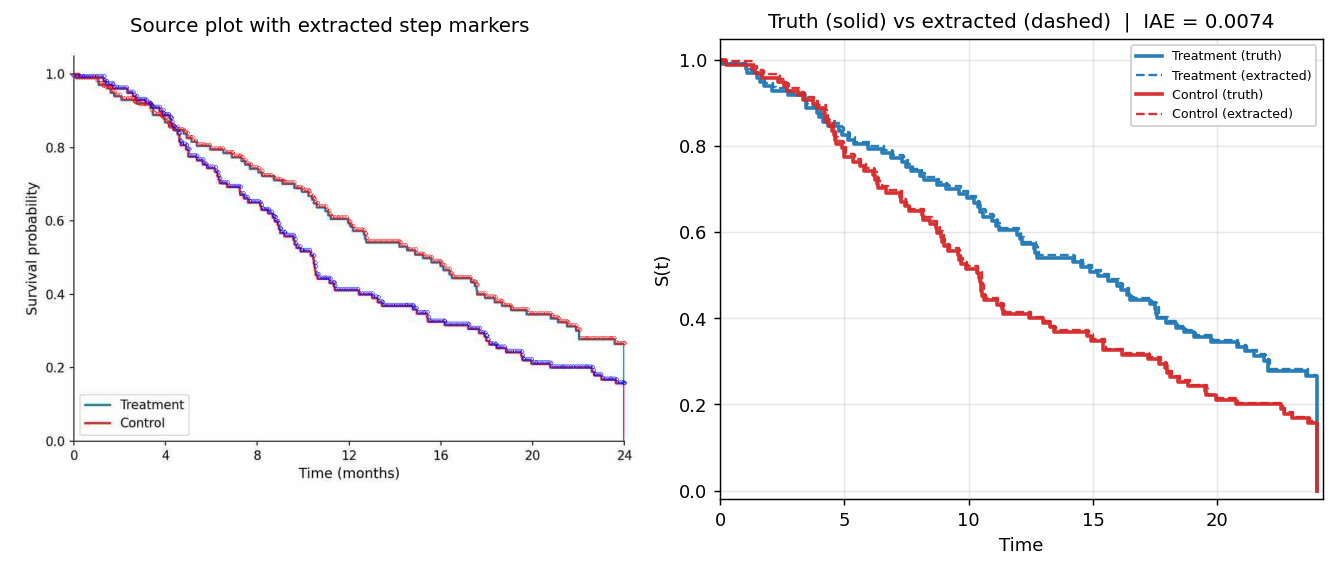}\\
\small \texttt{stress\_jpeg\_artifact} (IAE=0.0074, Med.\ AE=0.0052, Med.\ OS err=0.27~mo, 2 arms)
\end{minipage}
\hfill
\begin{minipage}[t]{\linewidth}
\centering
\includegraphics[width=\linewidth]{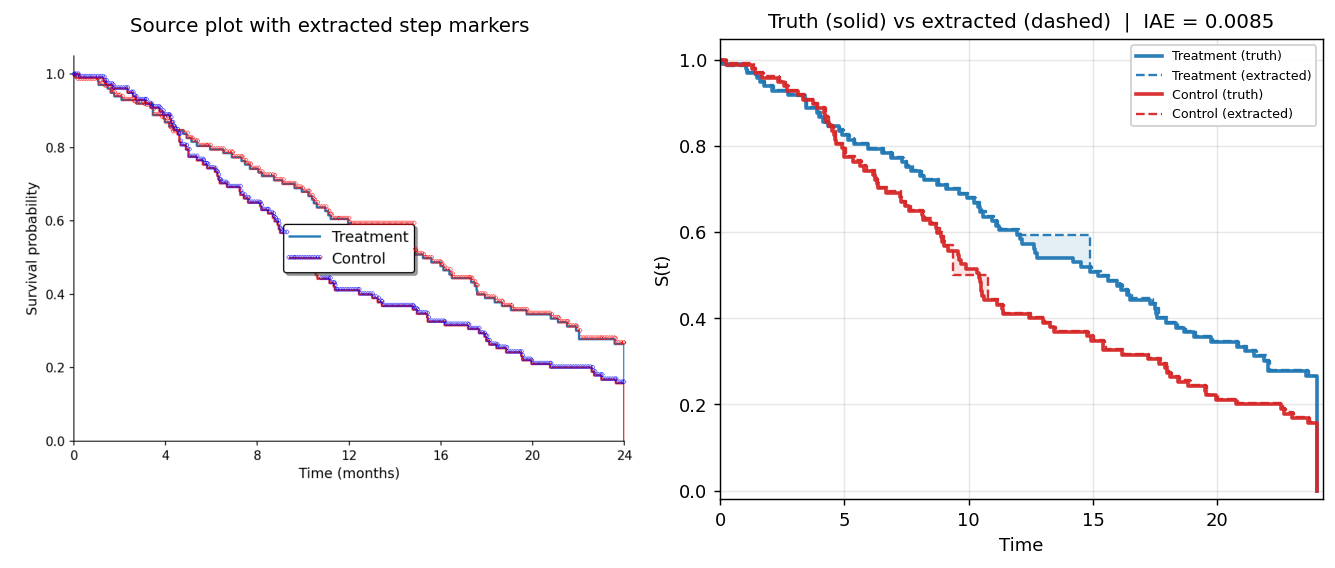}\\
\small \texttt{stress\_legend\_overlap} (IAE=0.0085, Med.\ AE=0.0030, Med.\ OS err=0.38~mo, 2 arms)
\end{minipage}
\caption{\textbf{stress\_jpeg\_artifact}: JPEG compression smears the edge; extraction stays within 1--2 px. \quad \textbf{stress\_legend\_overlap}: Legend overlaps the plot area; masked as a rectangular exclusion zone before tracing.}
\end{figure}
\begin{figure}[htb]
\centering
\begin{minipage}[t]{\linewidth}
\centering
\includegraphics[width=\linewidth]{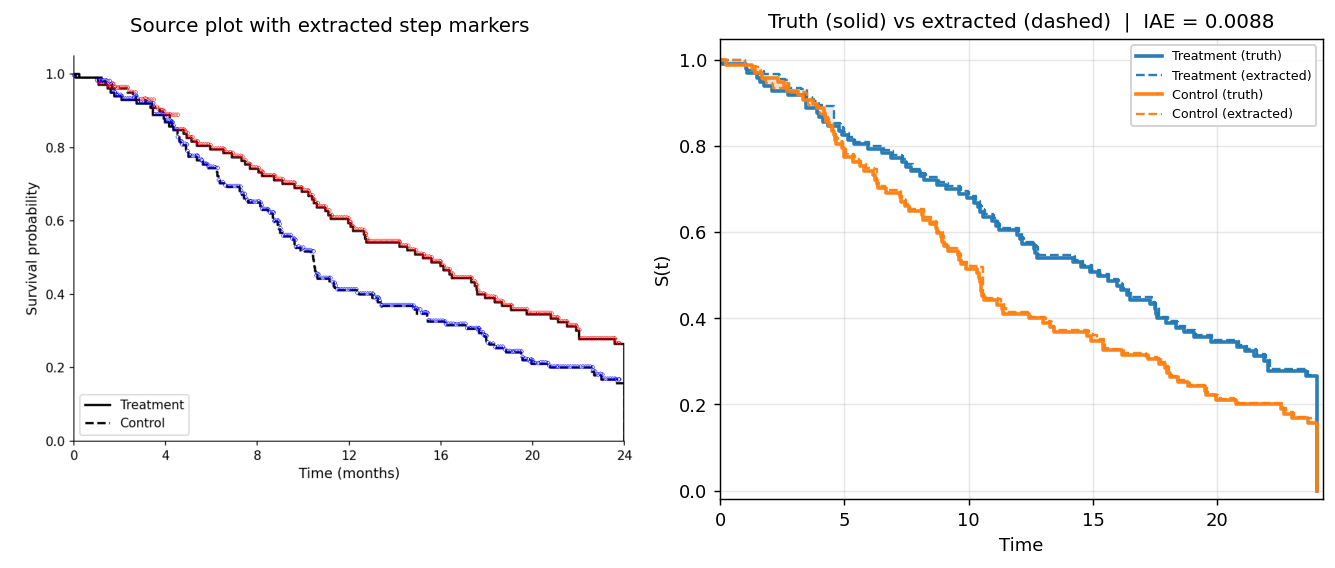}\\
\small \texttt{stress\_bw} (IAE=0.0088, Med.\ AE=0.0054, Med.\ OS err=0.29~mo, 2 arms)
\end{minipage}
\hfill
\begin{minipage}[t]{\linewidth}
\centering
\includegraphics[width=\linewidth]{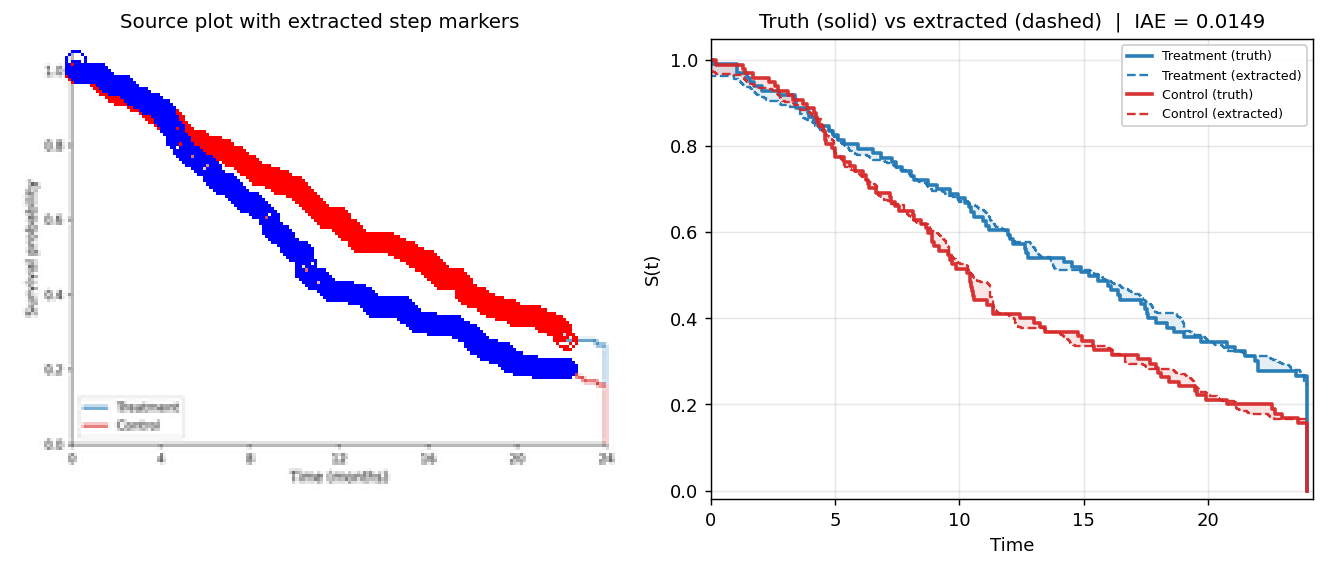}\\
\small \texttt{stress\_tiny} (IAE=0.0149, Med.\ AE=0.0129, Med.\ OS err=0.21~mo, 2 arms)
\end{minipage}
\caption{\textbf{stress\_bw}: Black \& white with solid vs dashed distinction. Grayscale separation holds; dashed gaps bridged. \quad \textbf{stress\_tiny}: Low pixel density; small mask errors translate to larger absolute errors in $S(t)$.}
\end{figure}
\begin{figure}[htb]
\centering
\begin{minipage}[t]{\linewidth}
\centering
\includegraphics[width=\linewidth]{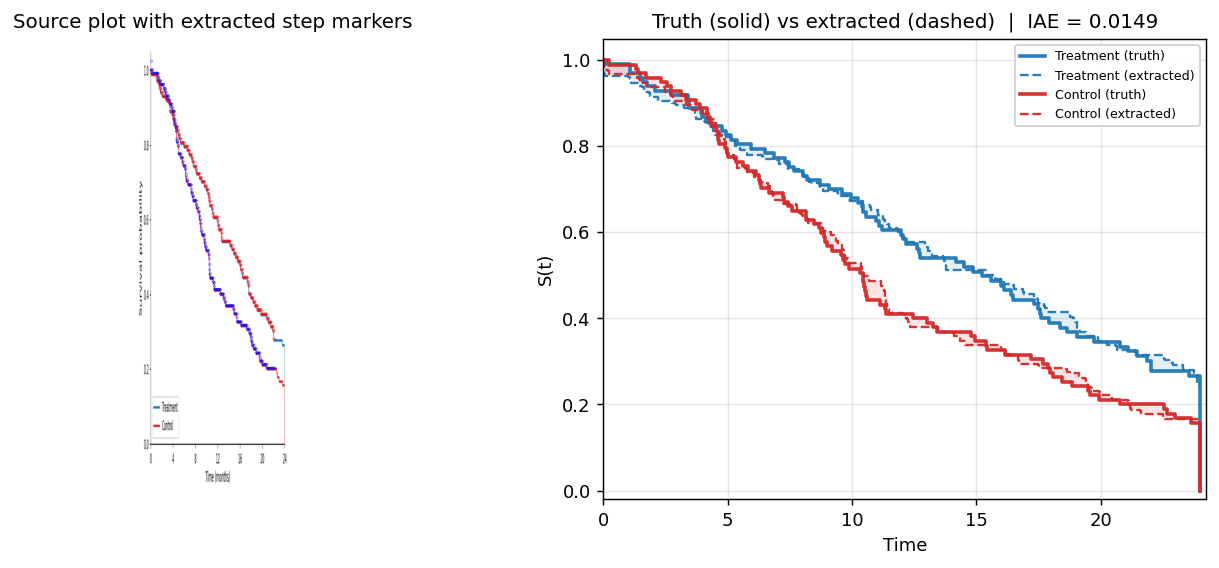}\\
\small \texttt{stress\_stretched\_tall} (IAE=0.0149, Med.\ AE=0.0135, Med.\ OS err=0.28~mo, 2 arms)
\end{minipage}
\hfill
\begin{minipage}[t]{\linewidth}
\centering
\includegraphics[width=\linewidth]{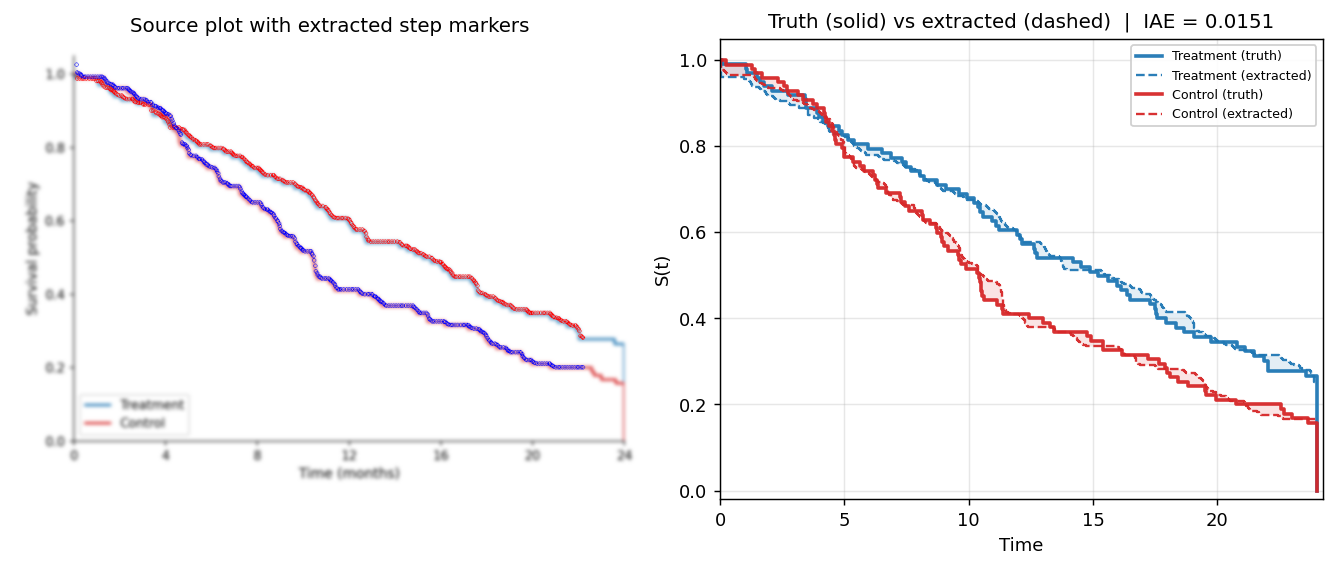}\\
\small \texttt{stress\_blurry} (IAE=0.0151, Med.\ AE=0.0134, Med.\ OS err=0.28~mo, 2 arms)
\end{minipage}
\caption{\textbf{stress\_stretched\_tall}: Stretched vertically; tick calibration resolves the bbox but column density is lower. \quad \textbf{stress\_blurry}: Gaussian blur widens the curve edge symmetrically; centroid strategy absorbs it.}
\end{figure}
\begin{figure}[htb]
\centering
\begin{minipage}[t]{\linewidth}
\centering
\includegraphics[width=\linewidth]{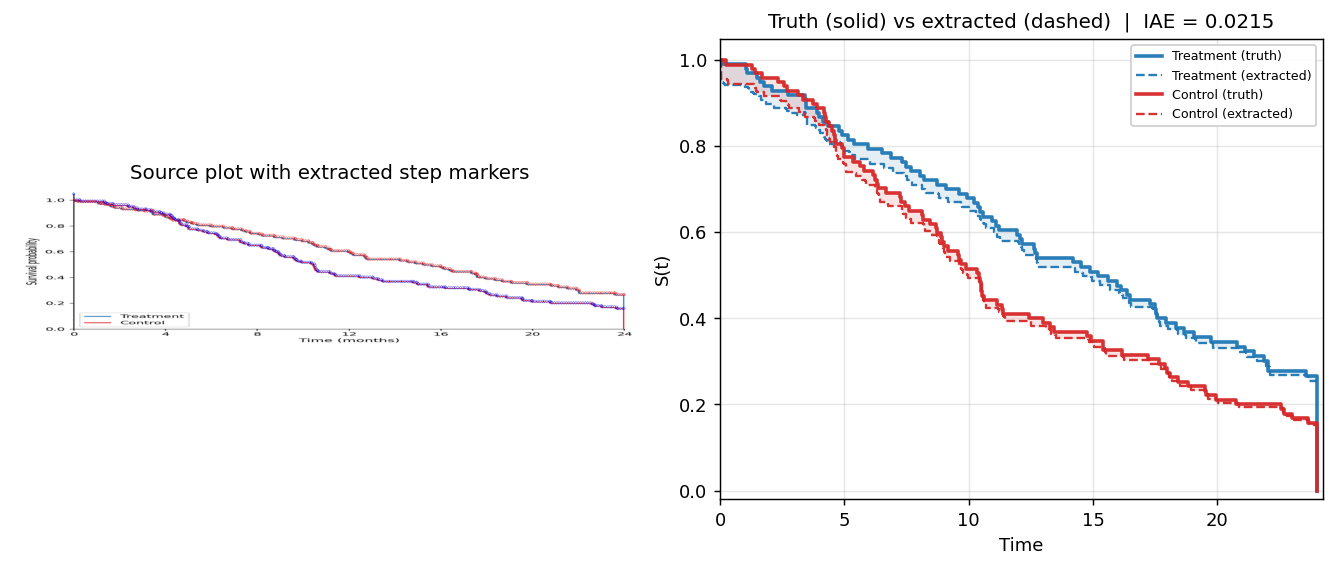}\\
\small \texttt{stress\_stretched\_wide} (IAE=0.0215, Med.\ AE=0.0187, Med.\ OS err=0.54~mo, 2 arms)
\end{minipage}
\hfill
\begin{minipage}[t]{\linewidth}
\centering
\includegraphics[width=\linewidth]{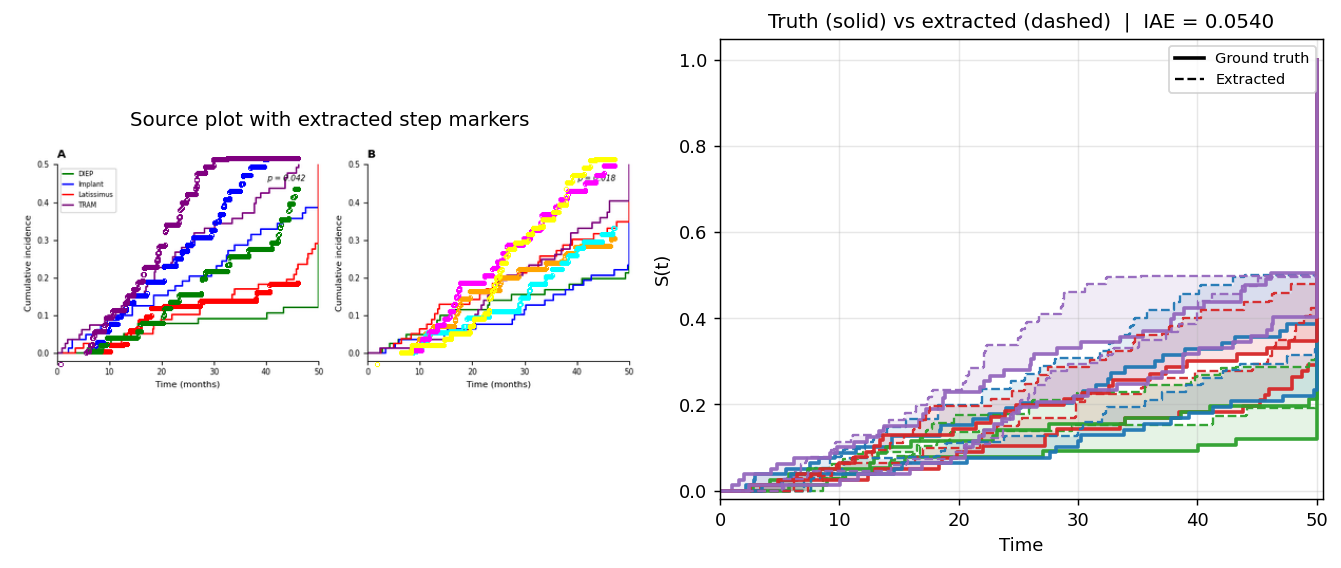}\\
\small \texttt{stress\_diep\_like} (IAE=0.0540, Med.\ AE=---, Med.\ OS err=---~mo, 8 arms)
\end{minipage}
\caption{\textbf{stress\_stretched\_wide}: Stretched horizontally; abundant columns, shallow $S$-axis resolution. \quad \textbf{stress\_diep\_like}: Two-panel cumulative-incidence plot with 8 arms total and heavy overlap. Per-arm IAE ranges 0.042--0.072; color attribution is the bottleneck---several arms share hue bands, so the tracer occasionally jumps between arms in dense regions.}
\end{figure}
\clearpage

\subsection{Combination Stress Tests (8)}
\begin{figure}[htb]
\centering
\begin{minipage}[t]{\linewidth}
\centering
\includegraphics[width=\linewidth]{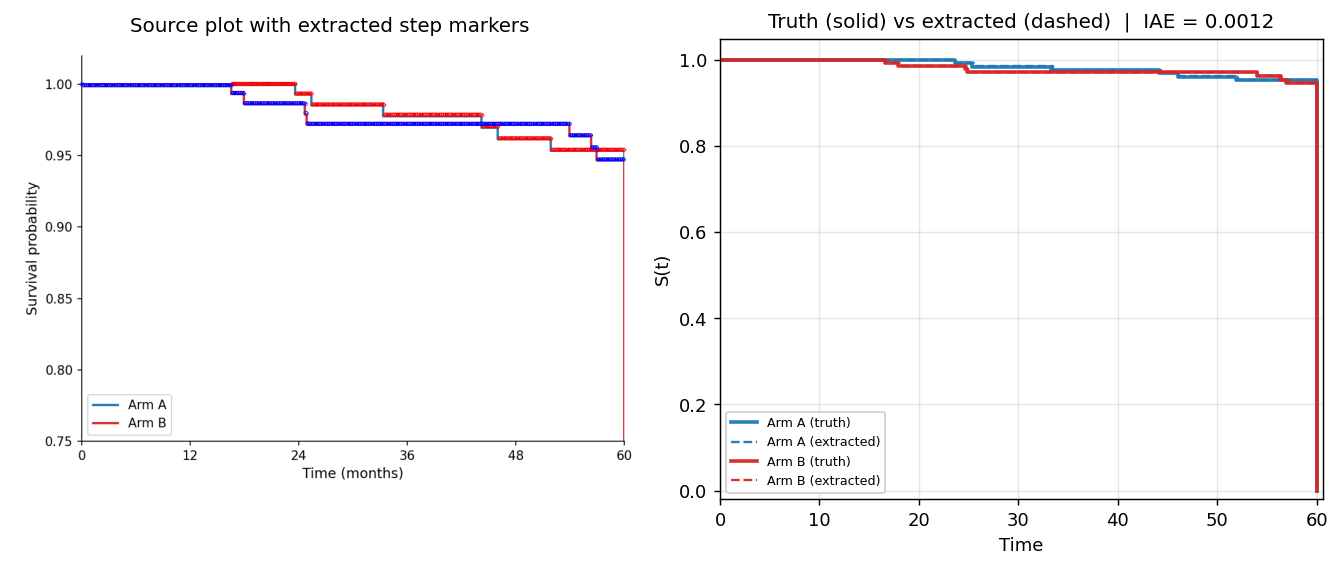}\\
\small \texttt{stress\_combo\_flat\_overlap} (IAE=0.0012, Med.\ AE=0.0005, Med.\ OS err=0.00~mo, 2 arms)
\end{minipage}
\hfill
\begin{minipage}[t]{\linewidth}
\centering
\includegraphics[width=\linewidth]{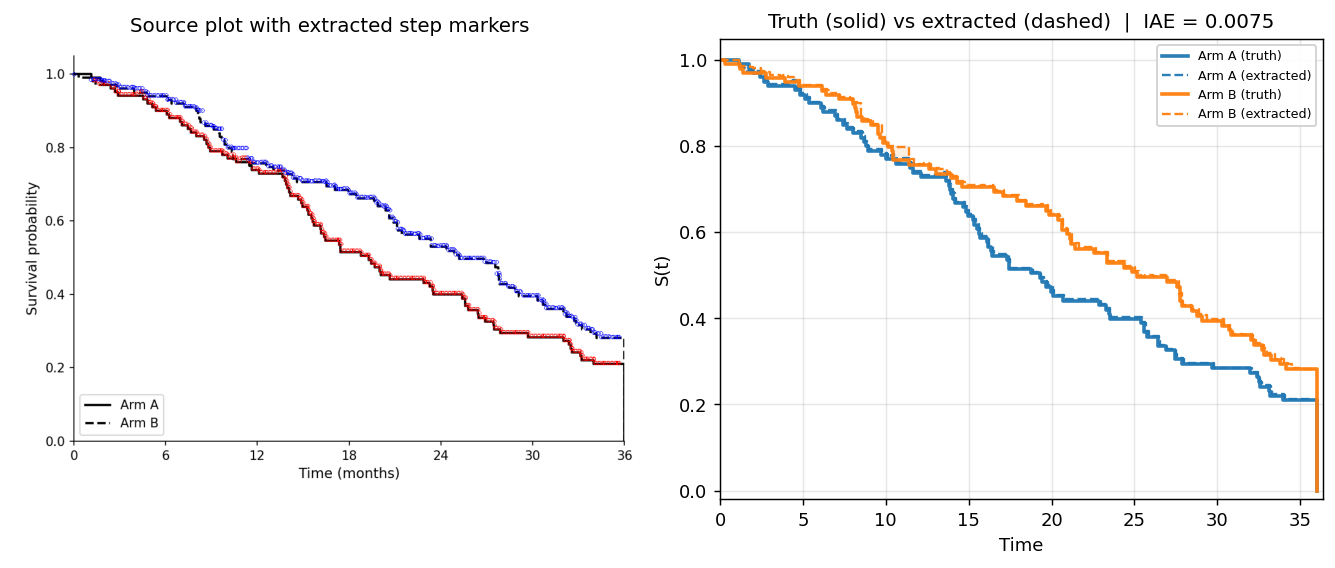}\\
\small \texttt{stress\_combo\_bw\_overlap} (IAE=0.0075, Med.\ AE=0.0037, Med.\ OS err=0.10~mo, 2 arms)
\end{minipage}
\caption{\textbf{stress\_combo\_flat\_overlap}: Both arms stay between 0.95--1.00 across 60~mo. Step markers land on the curves throughout; the combination of overlap and flatness defeats itself---any pixel near the curve gives near-zero error. \quad \textbf{stress\_combo\_bw\_overlap}: B\&W with overlapping solid and dashed curves; grayscale plus gap bridging resolves both.}
\end{figure}
\begin{figure}[htb]
\centering
\begin{minipage}[t]{\linewidth}
\centering
\includegraphics[width=\linewidth]{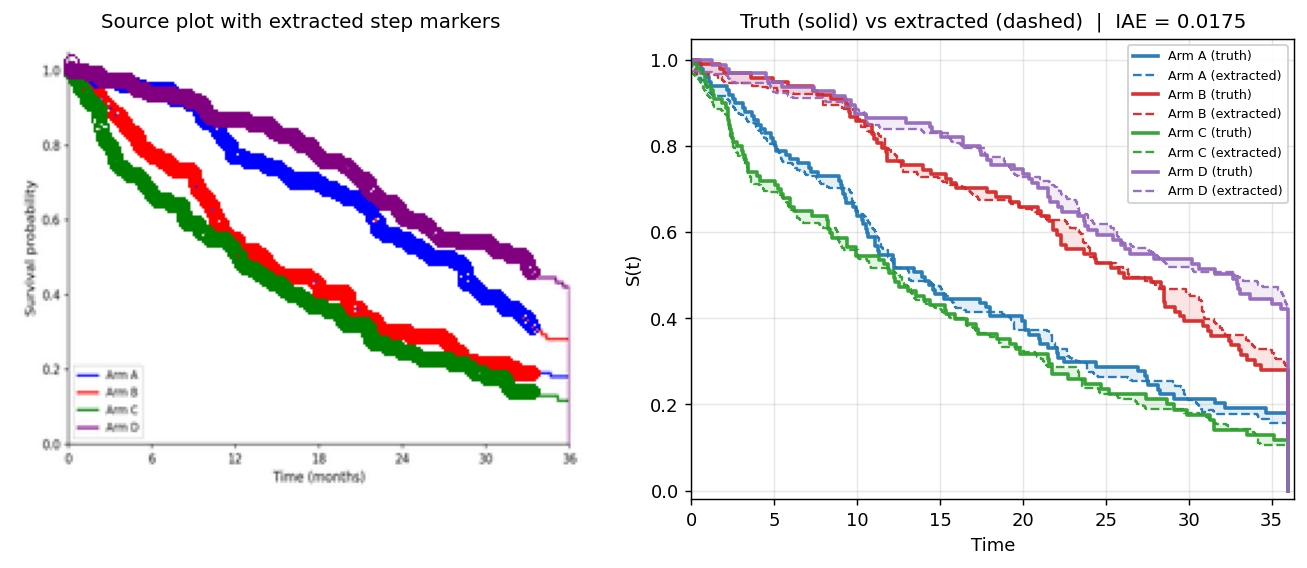}\\
\small \texttt{stress\_combo\_4arm\_tiny} (IAE=0.0175, Med.\ AE=0.0183, Med.\ OS err=0.41~mo, 4 arms)
\end{minipage}
\hfill
\begin{minipage}[t]{\linewidth}
\centering
\includegraphics[width=\linewidth]{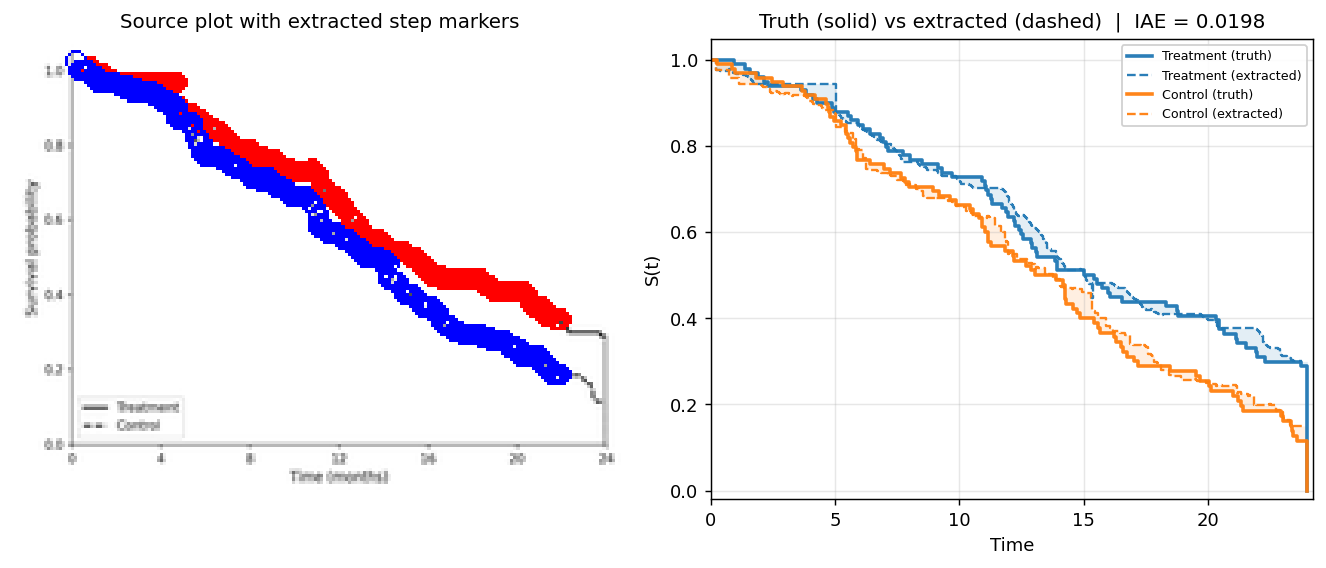}\\
\small \texttt{stress\_combo\_tiny\_bw} (IAE=0.0198, Med.\ AE=0.0161, Med.\ OS err=0.23~mo, 2 arms)
\end{minipage}
\caption{\textbf{stress\_combo\_4arm\_tiny}: 4 arms at tiny resolution; color separation holds but per-column sample count is low. \quad \textbf{stress\_combo\_tiny\_bw}: Tiny plus B\&W; low pixel count compounds grayscale ambiguity.}
\end{figure}
\begin{figure}[htb]
\centering
\begin{minipage}[t]{\linewidth}
\centering
\includegraphics[width=\linewidth]{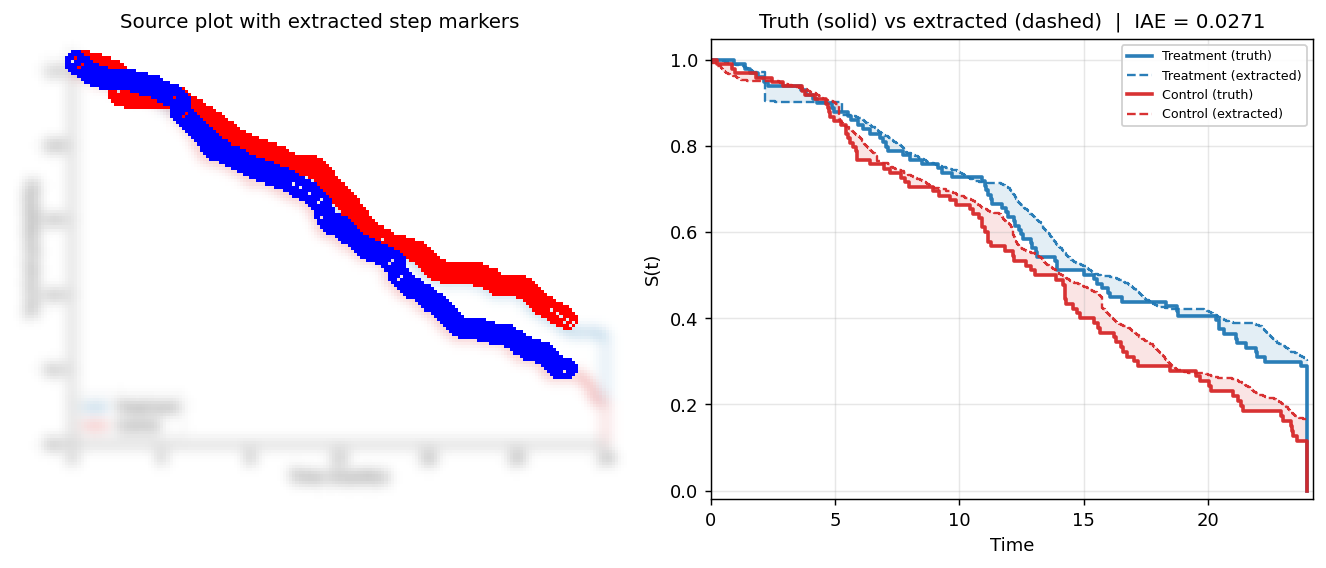}\\
\small \texttt{stress\_combo\_tiny\_blurry} (IAE=0.0271, Med.\ AE=0.0212, Med.\ OS err=0.20~mo, 2 arms)
\end{minipage}
\hfill
\begin{minipage}[t]{\linewidth}
\centering
\includegraphics[width=\linewidth]{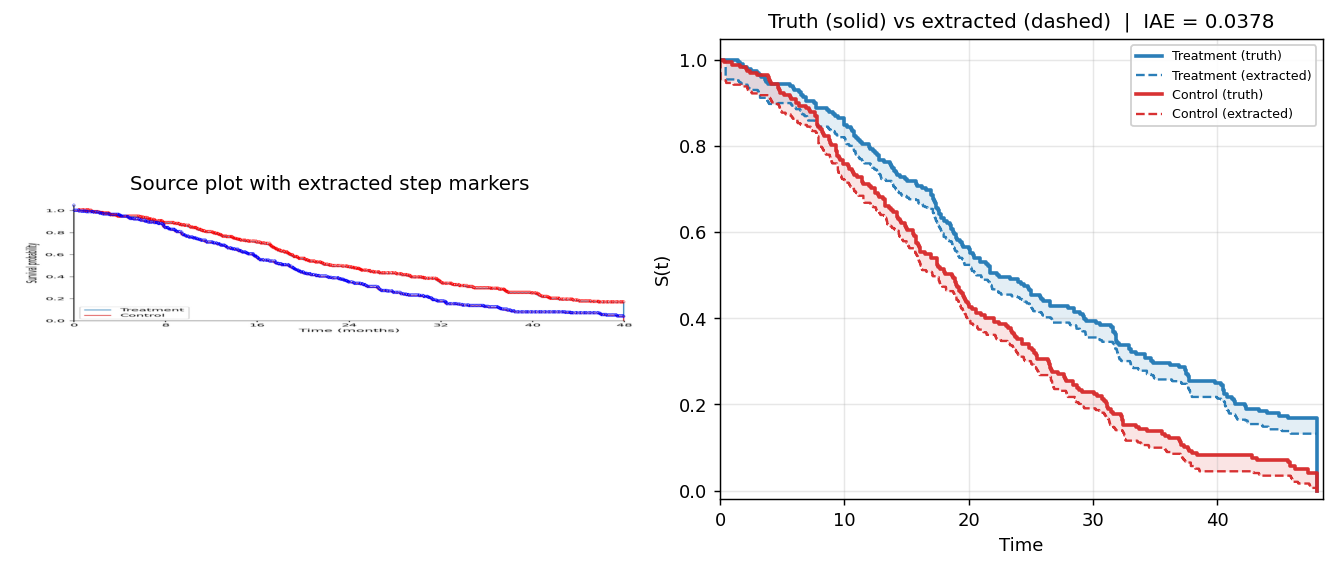}\\
\small \texttt{stress\_combo\_stretched\_dense} (IAE=0.0378, Med.\ AE=0.0375, Med.\ OS err=1.70~mo, 2 arms)
\end{minipage}
\caption{\textbf{stress\_combo\_tiny\_blurry}: Tiny plus blurry; edge widening dominates, yielding the highest IAE among these single-style combos. \quad \textbf{stress\_combo\_stretched\_dense}: Stretched aspect with dense step structure; occasional drift in densest regions.}
\end{figure}
\begin{figure}[htb]
\centering
\begin{minipage}[t]{\linewidth}
\centering
\includegraphics[width=\linewidth]{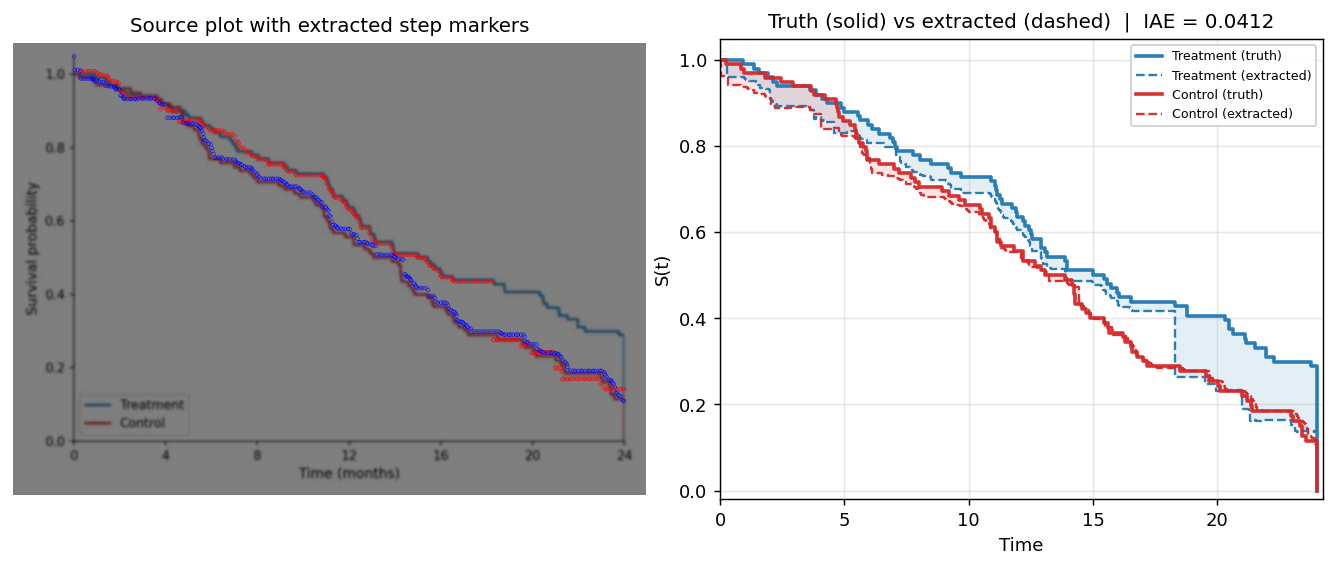}\\
\small \texttt{stress\_combo\_jpeg\_blurry\_dark} (IAE=0.0412, Med.\ AE=0.0256, Med.\ OS err=1.07~mo, 2 arms)
\end{minipage}
\hfill
\begin{minipage}[t]{\linewidth}
\centering
\includegraphics[width=\linewidth]{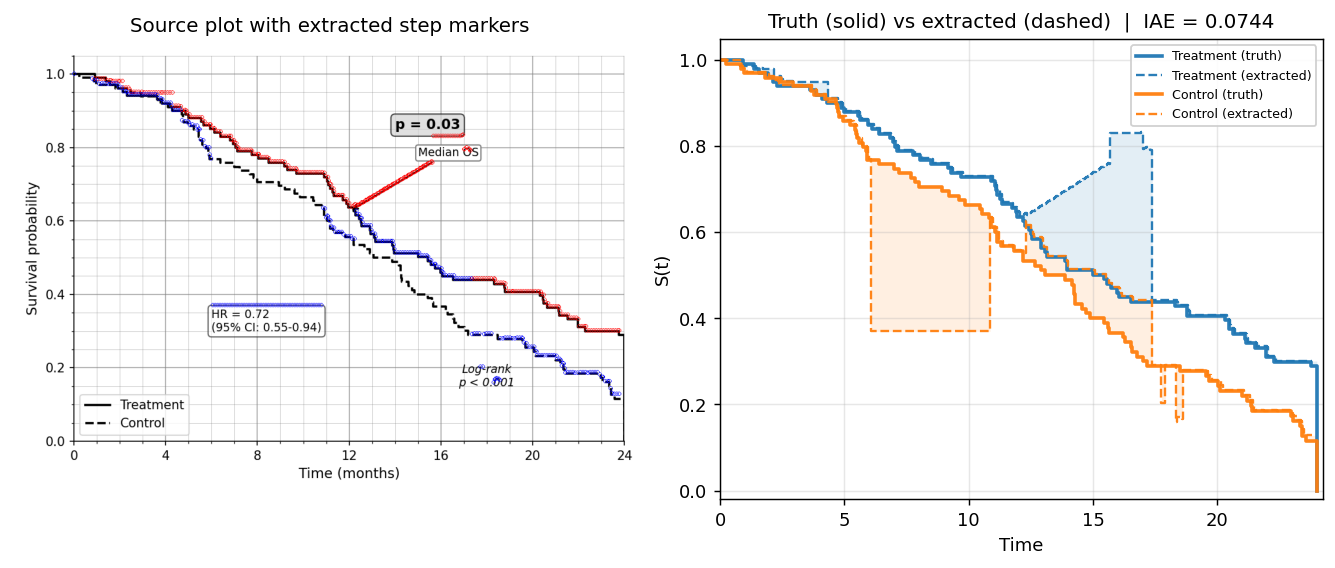}\\
\small \texttt{stress\_combo\_grid\_annotation\_bw} (IAE=0.0744, Med.\ AE=0.0058, Med.\ OS err=4.89~mo, 2 arms)
\end{minipage}
\caption{\textbf{stress\_combo\_jpeg\_blurry\_dark}: Heavy JPEG artifacts and blur widen the curve edge. Centroid tracking keeps the extraction within a few pixels of truth, but the smeared edge produces noisier step timing. \quad \textbf{stress\_combo\_grid\_annotation\_bw}: Adversarial combo: gridlines leak into the color mask, the ``Median OS'' diagonal leader line pulls the red markers off the true curve near $t=14$, and B\&W forces grayscale separation of solid vs dashed lines. Three failure modes stack.}
\end{figure}
\clearpage

\subsection{Best vs.\ Worst Case Analysis}
\label{sec:km-bestworst}

The easiest plots in the benchmark are not the ones with the fewest
adversarial features --- they are the ones where the adversarial
feature cancels itself out. \texttt{stress\_combo\_flat\_overlap} (IAE
$0.00118$, rank 1 of 32) stacks two nominal stressors: both arms are
nearly flat between $S \approx 0.95$ and $S \approx 1.00$, and the two
arms overlap visually throughout. Flatness means any reasonable pixel
assignment is within a few thousandths of truth in $S$-space; overlap
means even confused color attribution produces almost-correct values.
The stressor backfires. \texttt{edge\_high\_survival} (IAE $0.00124$)
shows the same effect for a different reason: 144~months of follow-up
with minimal events produces a near-horizontal curve that is tracked
cleanly.

The hardest plot is \texttt{stress\_combo\_grid\_annotation\_bw} (IAE
$0.07442$, rank 32 of 32). Three distinct failure modes compound: (i)
visible gridlines contaminate the color mask unless the
reference-line scrubber catches them; (ii) an on-plot ``Median OS''
annotation with a diagonal leader line physically crosses the curve
region around $t = 14$~mo, producing a magenta-colored stripe that
the red-channel mask follows for several columns --- the median OS
error of $4.89$~months reflects this jump; and (iii) black-and-white
rendering forces the extractor to disambiguate solid (Treatment) from
dashed (Control) by luminance rather than hue, and the annotation box
around ``HR = 0.72 (95\% CI: 0.55--0.94)'' adds further contamination.
Visually the extraction tracks the general shape of both curves but
exhibits a visible hop of ${\sim}15$~px near the median-OS annotation;
the rest of the plot is close.

Between the extremes, the failure pattern is informative.
\texttt{stress\_diep\_like} (IAE $0.054$, 8-arm cumulative-incidence
plot) fails at color attribution: several of the eight arms share
hue bands, so the continuity tracker occasionally snaps onto the wrong
arm in dense-overlap regions. \texttt{stress\_combo\_jpeg\_blurry\_dark}
(IAE $0.041$) fails at timing: the JPEG+blur combination widens the
curve edge to ${\sim}4$~px, and centroid tracking keeps spatial error
small but produces noisy step-onset timing (median OS error
$1.07$~mo). \texttt{stress\_combo\_stretched\_dense} (IAE $0.038$) and
\texttt{stress\_combo\_tiny\_blurry} (IAE $0.027$) similarly reflect
resolution-limited timing errors rather than systematic bias. The
accuracy ceiling in these cases is set by the input image, not by the
extractor. A full per-plot gallery with extraction overlays and
per-plot observations appears in Appendix~\ref{app:km-gallery}.

\section{Skill 1: KM Plot Extraction Pipeline}
\label{app:skill_1}

\begin{minted}{markdown}
# KMGen: Kaplan-Meier Curve Extraction Agent

You are given an image of a Kaplan-Meier survival plot. Your job is to extract the survival curve data as precisely as possible by writing and executing Python code tailored to this specific image.

## Your Process

### Step 1: Look at the image and strategize
Visually assess the image. Identify:
- Curve count, colors, line styles
- Axis ranges and scale (proportion 0-1.0 vs percentage 0-100%)
- Interference sources: reference lines, annotations, legends, gridlines, CI bands
- Patients-at-risk table (if present)

Then pick your techniques from the toolbox below. State what challenges you see and how you'll handle each one.

**Trust your eyes.** You can read axis labels, count curves, and see reference lines from the image directly. You may run **exactly one** short diagnostic script to confirm things you can't resolve visually (e.g., sampling pixel RGB values where curve colors look ambiguous). This is your only investigative code -- after it, go straight to writing the full extraction script in Step 2.

### Step 2: Write and run extraction code
Write ONE self-contained Python script that:
1. Loads the image (always `.convert("RGB")` first -- JPEG may be CMYK)
2. Detects the bounding box
3. Creates color masks for each arm
4. Traces each curve
5. Extracts coordinates at daily granularity (1/30 month)
6. Enforces monotonicity
7. Prints summary stats (median crossings, coordinate count, S(0) sanity check)

Run it. Check the output. If medians are way off or S(0) != 1.0, fix the bbox or masks and rerun.

### Step 3: Verify and save
Run the verification checklist:
- **PAR floor check**: `S(t) >= PAR(t) / PAR(0)` at every timepoint (if PAR available)
- **Published anchor check**: compare extracted medians to values printed on the plot
- **Monotonicity**: zero violations
- **Coverage**: coordinates span from x_min to near x_max

Save deliverables:
- `<name>.json` -- extraction data
- `<name>_report.md` -- your visual assessment, technique choices, verification results

After each extraction attempt, generate an annotation overlay and visually inspect it:
```python
import json
from benchmark_extract import annotate_image
with open('<name>.json') as f:
    extraction = json.load(f)
annotate_image('<image_path>', extraction, '<name>_annotation.png')
```
Then **read the annotation PNG** and compare the overlay dots to the underlying curves. Look for: dots diverging from curves, flat segments where curves decline, traces jumping to reference lines or the wrong arm. Fix these with targeted spot-fixes before saving final results. Do NOT write your own annotation code -- use `annotate_image()` only.

## Technique Toolbox

Pick what fits. Each has tradeoffs.

### Image Preprocessing
- **Color space**: Always `.convert("RGB")` first. JPEG may be CMYK.
- **Upscaling (2x LANCZOS)**: Helps when lines are thin (1-2px). Skip if already high-res.

### Bounding Box Detection
Bbox errors are the #1 source of systematic extraction error.

- **Tick mark detection**: Find evenly-spaced short segments on each axis. The outermost ticks define the bbox, NOT the frame border. Verify spacing is consistent (linear regression R^2 > 0.999).
- **Frame vs tick disambiguation**: Frame borders span the full axis length. Ticks are short perpendicular segments. Use tick spacing to tell them apart.
- **Sanity check**: `S(0)` at the leftmost curve pixel must be ~1.0. If it's 0.82 or 1.05, the bbox is wrong.
- **Axis scale**: Read y-axis labels -- "100, 80, 60..." = percentage, "1.0, 0.8, 0.6..." = proportion.

### Color Separation
- **HSL hue matching**: Best for distinguishing colors close in RGB but different in hue.
- **RGB thresholds**: Simple, works when colors are far apart. Sample actual pixels empirically.
- **Grayscale intensity (BW plots)**: When curves appear "both black," sample pixels where they're separated. One is often pure black (gray < 50), the other medium gray (140-170, sometimes with B > R). Use intensity thresholds.
- **Legend/text exclusion**: Mask out legend boxes, annotations, titles before filtering.

### Reference Line Detection (Critical)
Dashed horizontal reference lines (median at S=0.5, quartile lines) contaminate curve masks.

Scan each row for dark pixels spanning >30% of plot width -- that's a reference line, not a curve. Mask those rows (+/-2px) BEFORE tracing.

Vertical reference lines (median time markers): detect by scanning columns similarly.

### Curve Tracing
- **Continuity tracking**: Follow the curve column by column, picking the pixel cluster closest to the previous position.
- **Monotonicity constraint**: KM curves only go down. Reject upward jumps > anti-aliasing tolerance.
- **Gap handling**: Carry forward through gaps (dashed curves, reference line exclusions). Set max_gap based on dash pattern.
- **Centroid correction**: At every column, use `np.mean(cluster)` instead of topmost pixel. Measured -38% IAE improvement.

### Extraction
- **Daily granularity**: Sample at 1/30 month intervals. A 24-month curve = ~720 points per arm. Coarser sampling loses IPD precision.
- **Step detection**: Where steps are visually clear, detect vertical drops (min_drop_px=2-3).
- **Monotonicity enforcement**: Post-extraction, clip any upward moves in S(t).
- **Terminal drop detection**: At the last column, if the mask spans >10px vertically, append the bottom as a final step.
- **Fragmented extraction**: Your script is tailored to THIS image only -- don't generalize. If different regions of the curve need different techniques (e.g., color separation works for months 0-20 but the tail needs positional tracking), write separate extraction logic per region and stitch the coordinates together.

### Patients-at-Risk Table
- Crop once, zoom enough to read, move on. If unreadable after one attempt, skip it.
- Verify values are monotonically non-increasing.
- Use for PAR floor constraint: `S(t) >= PAR(t) / PAR(0)`.
- First PAR value = `n_total` for IPD reconstruction.

### Published Anchors
Printed medians, HR, landmark rates are free calibration checks. Compare and flag discrepancies > 0.5 months.

### Self-Correction (attempt 2+)
Generate diagnostics BEFORE rewriting:
1. Color mask overlay on original image (reveals contamination)
2. Zoomed crops of 3-4 worst regions
3. Perpendicular intensity profile at median crossing

Diagnose root cause, then make targeted fixes. Do NOT rewrite from scratch.

### Spot-Fixes
After verification, if a small localized artifact remains (e.g., trace sticking to a reference line in one region, a flat segment where the curve should decline), hardcode a local correction -- interpolate from clean neighbors or re-extract just that t-range. Tag patched coordinates with `"method": "interpolated"`. Don't re-run the full extraction for a 5-point fix.

## Extraction JSON Format

```json
{
  "image": "<filename>",
  "bbox": [left, top, right, bottom],
  "axis": {"x_min": 0, "x_max": 21, "y_min": 0.0, "y_max": 1.0},
  "arms": [
    {
      "label": "Arm A",
      "color": "blue",
      "coordinates": [
        {"t": 0.0, "s": 1.0, "method": "step"},
        {"t": 1.35, "s": 0.96, "method": "step"},
        {"t": 5.50, "s": 0.42, "method": "sample"}
      ]
    }
  ],
  "patients_at_risk": {
    "Arm A": {"0": 54, "1": 51, "2": 48},
    "Arm B": {"0": 53, "1": 51}
  }
}
```

## Available Libraries
numpy, Pillow (PIL), scipy, json, sys, pathlib. Do NOT use OpenCV.

## Key Principles
- **Trust your eyes, then code.** Visual assessment tells you the strategy. Code executes it. Don't use code to rediscover what you can already see.
- **Empirical over theoretical.** Sample actual pixel colors rather than guessing.
- **Verify with numbers, not vibes.** PAR floor check, published anchor comparison, S(0) sanity check.
- **Reference lines are poison.** Detect and mask before tracing.
- **Bbox precision matters most.** Calibrate from ticks, verify S(0) ~= 1.0.
- **Daily granularity.** ~30 coordinates per month per arm for IPD reconstruction.
- **Self-correction: diagnose, don't rewrite.** Use mask visualizations and zoomed crops.
\end{minted}

\section{Skill 2: IPD Generation}
\label{app:skill_2}

\begin{minted}{markdown}
# Clinical Trial Config Extraction

You are a clinical data abstraction agent. Your job: extract structured trial configs so the downstream `SyntheticPatientGenerator` can produce realistic patient-level data.

YOU CAN WRITE CODE. The ClinicalTrials.gov JSON is your primary source -- start by parsing `resultsSection`, `protocolSection`, and `baselineCharacteristicsModule` programmatically.

> [!CAUTION]
> **DATA LEAKAGE PREVENTION:** Extract ONLY from provided documents and the provided KM CSVs. Do NOT use external prior knowledge of the IPD.

---

## Source Priority (READ THIS FIRST)

Derive **as much as possible from the ClinicalTrials.gov JSON**, then fall back to the next source only when a field is missing or insufficiently granular. The strict priority order is:

1. **`NCTXXXXXXXX.json` (ClinicalTrials.gov)** -- primary source. Use it for arms, enrollment, demographics, AEs, regimen description, and anything else it reports. It is the most structured and the most consistent across trials.
2. **`*OS_km.csv` (reconstructed OS KM CSV)** -- ground-truth overall survival. Wire it into `os_km_csv` verbatim. Never re-fit it.
3. **Primary publication (+ supplementary appendix)** -- fall back here only for fields the CT.gov JSON omits or under-reports (e.g. ECOG breakdown not in `baselineCharacteristicsModule`, Details about AEs, finer race/region splits).
4. **SAP / protocol** -- last resort, used for fields the prior three sources cannot supply: cycle structure (`cycle_length_days`, `n_induction_cycles`), treatment phase boundaries, assessment schedule, etc.

For every field you populate, you should be able to name which of these four sources it came from. If a field isn't grounded in any of them, **omit it** and let `synthetic_patient_generator.py` use its default -- never guess.

---

## Inputs

For each trial `NCTXXXXXXXX`, you are given a directory under `datasets/<trial>/` containing:

- `NCTXXXXXXXX.json` -- ClinicalTrials.gov record (**primary source**)
- **One reconstructed OS KM CSV** -- a file matching `*km.csv` whose filename indicates overall survival (e.g. `*OS_km.csv`). Columns: `arm`, `time`, `event`. Ground-truth OS.
- Primary publication (PDF/text)
- Supplementary appendix (if available)
- SAP / protocol (if available)

### Reading publications, supplements, and SAPs

Do **not** try to OCR images or install PDF libraries. Use the `Read` tool directly -- it handles both formats natively:

- **PDF files** (`*.pdf`) -- `Read` extracts text and layout. For large PDFs (>10 pages), pass the `pages` parameter (e.g. `pages: "1-5"`) and page through the document; max 20 pages per call.
- **PNG / JPG images** (`*.png`, `*.jpg`) -- `Read` passes the image to the multimodal model, so you can visually inspect rendered SAP schedules, Table 1 screenshots, forest plots, etc. This is strictly better than `pytesseract` for figures and schedule-of-assessments diagrams.

Only fall back to Python if you specifically need programmatic table extraction that `Read` cannot give you.

> **Scope: OS only.** Even when other KM CSVs (PFS, DOR, TTR, ...) are present in the directory, **ignore them**. Do not wire PFS or any other endpoint into the config; the downstream generator only consumes OS as ground truth.

> [!CAUTION]
> **NEVER read or reference any `ipd/` subfolder inside `datasets/<trial>/`.** Those directories contain the real individual patient data and are reserved exclusively for downstream evaluation. Reading them -- or any file under them -- would leak ground truth into the extraction prompt and invalidate the entire benchmark. Treat `ipd/` as if it does not exist: do not glob it, do not list it, do not open any file beneath it.

Discover the OS KM CSV like this:

```python
import glob, os
# NOTE: only the trial's top-level files. Do NOT recurse into `ipd/` --
# that subfolder holds the held-out ground-truth IPD and is off-limits.
os_csvs = sorted(
    p for p in glob.glob(os.path.join(datasets_dir, trial, "*km.csv"))
    if "os" in os.path.basename(p).lower()
    and os.sep + "ipd" + os.sep not in p
)
```

If multiple `*km.csv` files exist, pick the one whose filename indicates OS and ignore the rest.

---

## Process

Work the sources **in priority order**. Don't open the publication until you've drained the CT.gov JSON; don't open the SAP until you've drained the publication.

### 1. Drain the ClinicalTrials.gov JSON (primary)

Parse the JSON programmatically. Pull out everything it offers before touching any other source:

- **Arms** -> `protocolSection.armsInterventionsModule.armGroups[*].label` and `.description`. These define the per-arm configs and the canonical `arm_name` strings (cross-check against the OS KM CSV `arm` column and reconcile).
- **Enrollment / `n_enrolled`** -> **per-arm count only** (each config is one arm; `n_enrolled` is THIS arm's enrollment, never the trial total). Pull from `resultsSection.baselineCharacteristicsModule.measures` denominators, `participantFlowModule`, or AE module `numAtRisk`. Cross-check against the OS KM CSV row count for that arm -- they should match. The downstream generator emits exactly `n_enrolled` synthetic patients per arm, so this number must reflect the analyzed arm size, not the protocol-section total.
- **Trial metadata** -> `protocolSection.identificationModule.officialTitle` -> `title`; `conditionsModule.conditions` -> `condition`; arm description -> `treatment`.
- **Demographics** -> `resultsSection.baselineCharacteristicsModule.measures` for age (mean/SD/min/max), sex, race, region, ECOG, weight, height. Read **per-arm** denominators where available; only use pooled values if the JSON only reports pooled.
- **AEs** -> `resultsSection.adverseEventsModule.seriousEvents` and `.otherEvents`. Each entry has `term`, `organSystem`, and per-arm `stats` with `numAffected` and `numAtRisk`. Compute `probability = numAffected / numAtRisk` per arm. Use the JSON's `organSystem` strings directly (they are MedDRA SOC). This is your AE backbone.
- **Eligibility / regimen hints** -> `eligibilityModule`, `armGroups[*].description`, `interventions[*].description` for treatment classification (chemo vs IO vs TKI), which informs `induction_ae_fraction` defaults and risk modifiers.

Whatever the JSON gives you, that is the value. Do not adjust or replace it from later sources unless the later source is strictly more granular for that field.

### 2. Wire Up the Ground-Truth OS KM CSV

For each arm, set `os_km_csv` to the path of the OS `*km.csv`. The `arm_name` in the config must match a value in the CSV's `arm` column **exactly** -- the generator filters by that name and bootstraps `(time, event)` pairs from those rows.

**The OS CSV is GROUND TRUTH. Do not fit anything.** No Weibull, no median extraction, no event-fraction tallying, no lifelines calls. Do not populate `median_os_months`, `death_probability`, `max_followup_months`, or `os_weibull_shape` -- those parametric fields are ignored by the generator whenever `os_km_csv` is set, and adding them only creates confusion. The CSV is the curve; the generator reproduces it exactly.

If no OS KM CSV exists for an arm, **only then** fall back to the legacy parametric fields (`median_os_months`, `death_probability`, `max_followup_months`, `os_weibull_shape`) -- and pull those from the publication, not from the SAP.

**Only OS.** If the directory also contains PFS, DOR, or other endpoint CSVs, do not wire them up anywhere.

### 3. Fill Gaps from the Publication (+ Supplement)

Open the primary publication only to fill what CT.gov could not:

- **Demographics CT.gov omitted** -- e.g. ECOG bucketing, finer race/region splits, weight/height if not in `baselineCharacteristicsModule`. Read Table 1 and the supplement before giving up. If the publication only reports KPS, the KPS->ECOG bucket mapping (100/90->0, 80/70->1, <=60->2) is acceptable but note in the config that ECOG was derived from KPS so the mismatch is traceable.
- **AEs below CT.gov's reporting threshold** -- supplementary appendix tables often list AEs at 1-10% that CT.gov truncates at 5%. Add these to the AE list.
- **Safety narrative AEs** -- case descriptions of rare events not tabulated anywhere else.
- **Subgroup OS** -- if the paper reports a subgroup OS materially different from the overall, that informs risk-modifier overrides.

**Never guess a demographic field.** Every value must be grounded in CT.gov, the publication, or the supplement -- or omitted entirely so the generator default applies. A guessed placeholder is worse than a default; past fidelity evaluations show real IPD distributions diverge sharply from generic guesses (e.g. some trials enroll only ECOG 1/2 with no zeros).

**Note: Merged buckets must be split, not collapsed.** When a source reports a categorical demographic as a merged bucket (e.g. ECOG "0-1" vs "2", age "<65" vs ">=65", race "White" vs "Other"), do **not** dump the entire bucket onto a single endpoint key
- **Split uniformly across the merged categories** by default
- **Or split with a documented prior** when one is available from the same source (e.g. a subgroup forest plot in the supplement reports ECOG 0 separately for a sub-analysis -- use that ratio).
- Add a brief note in the config (e.g. as an inline comment when generating, or in your extraction notes) that the split was uniform/derived, so the provenance is traceable.


### 4. Last-Resort Fields from the SAP / Protocol

The SAP is consulted **only** for what the prior three sources cannot supply:

- **`cycle_length_days` and `n_induction_cycles`** -- these MUST come from the SAP/protocol. CT.gov arm descriptions sometimes summarize the regimen but rarely give the precise induction-vs-maintenance boundary. Open the SAP, find the schedule of assessments and dose-modification rules, and use those to set both fields.
- **Treatment phases** -- induction / consolidation / maintenance boundaries, total planned treatment duration, max follow-up.
- **Assessment schedule** -- when labs are drawn (which determines when lab AEs get detected).
- **Class-effect AEs** -- known toxicities of the drug class that aren't observed in the trial-specific tables. Tier-4 (<1%, estimate 0.01-0.04). **Do not pad** the AE list with class-effect guesses to hit a target count; if CT.gov + publication + supplement genuinely produce fewer AEs, leave the list at whatever is grounded.

### 5. Identify Arms (cross-check)

Produce a **separate config JSON for each arm**. Arm names must match the `arm` column in the OS KM CSV verbatim (or, if the CSV uses different labels than CT.gov, reconcile to the CT.gov labels and verify the mapping is unambiguous). Each config gets its own `arm_name`, `n_enrolled` (this arm's count only -- never the trial total), survival source, and arm-specific demographics + AE probabilities. **Do NOT pool across arms.** **Do NOT add a separate `n_synthetic` field** -- the generator emits exactly `n_enrolled` synthetic patients for this arm.

### 6. Assign Risk Modifiers (if needed)

Read `synthetic_patient_generator.py` for the full set of defaults:
- `_RISK_FIELD_DEFAULTS` -- OS multipliers, death probability deltas, comorbidity fractions, anthropometric multipliers
- `_DEFAULT_AE_RISK_MODIFIERS` -- per-factor, per-organ-system AE risk multipliers
- `TrialConfig` dataclass -- all field defaults and documentation

The generator deep-merges your overrides with these defaults. **Only override fields where the documents provide disease- or treatment-specific evidence** that diverges from standard cytotoxic chemo. Omit risk modifier fields entirely to use defaults.

**When to override:**
- **Immunotherapy** -> boost endocrine/immune/skin AE modifiers
- **Platinum agents** -> boost renal/neuro
- **TKIs** -> boost skin/hepatobiliary/cardiac
- **Paper reports subgroup OS** -> compute ratio (subgroup median / overall median)
- **High-comorbidity disease** (pancreatic, HCC, elderly lung) -> increase `fraction_high_comorbidity`
- **Small trial** (<100 patients) -> decrease `patient_ae_propensity_sigma`

---

## Rules (from past fidelity failures)

### AE Term Naming
Use **exact MedDRA Preferred Term** spelling -- and CT.gov's `term` strings already are MedDRA PTs, so prefer them verbatim:
- `"Neuropathy peripheral"` not `"Peripheral neuropathy"`
- `"Upper respiratory tract infection"` not `"Upper respiratory infection"`

If you pull an AE from the publication or supplement instead, use the verbatim spelling from that table.

### Organ System Naming
Use **exact** full MedDRA SOC names from this list (no abbreviations). CT.gov's `organSystem` field is already canonical -- pass it through unchanged.
Blood and lymphatic system disorders
Cardiac disorders
Congenital, familial and genetic disorders
Ear and labyrinth disorders
Endocrine disorders
Eye disorders
Gastrointestinal disorders
General disorders and administration site conditions
Hepatobiliary disorders
Immune system disorders
Infections and infestations
Injury, poisoning and procedural complications
Investigations
Metabolism and nutrition disorders
Musculoskeletal and connective tissue disorders
Neoplasms benign, malignant and unspecified
Nervous system disorders
Pregnancy, puerperium and perinatal conditions
Psychiatric disorders
Renal and urinary disorders
Reproductive system and breast disorders
Respiratory, thoracic and mediastinal disorders
Skin and subcutaneous tissue disorders
Social circumstances
Surgical and medical procedures
Vascular disorders

### AE Timing
Most TEAEs cluster in the induction phase. Set `induction_ae_fraction` e.g.:
- **Cytotoxic chemo induction**: 0.85-0.92
- **Immunotherapy only**: 0.50-0.65
- **Targeted therapy**: 0.60-0.75
- **Maintenance-only**: 0.30-0.50

### AE Deduplication
One entry per AE term:
- `probability` = **any-grade** incidence fraction
- `is_serious` = `True` if grade >=3 >= 50% of any-grade

When the same term appears in both CT.gov json serious and other event lists for the same arm, combine them into a single entry (sum `numAffected`, use the larger `numAtRisk`).

---

## Output Format

One JSON per arm: `syn_datasets/NCTXXXXXXXX/<arm_label>_config.json`

```json
{
  "trial_id": "NCTXXXXXXXX",
  "arm_name": "Experimental Drug + Chemo",
  "title": "Full trial title",
  "condition": "Disease condition",
  "treatment": "Treatment regimen for THIS arm",
  "n_enrolled": 0,            // per-arm count for THIS arm -- also the synthetic generation target
  "cycle_length_days": 21,
  "n_induction_cycles": 4,
  "induction_ae_fraction": 0.85,
  "age_mean": 64.0,
  "age_sd": 8.3,
  "age_min": 40,
  "age_max": 82,
  "fraction_male": 0.70,
  "ecog_distribution": {"0": 0.43, "1": 0.43, "2": 0.14},
  "weight_mean": 75.8,
  "weight_sd": 16.25,
  "height_mean": 169.9,
  "height_sd": 9.04,
  "race_distribution": {"White": 0.972, "Black": 0.009, "Other": 0.019},
  "region_distribution": {"United States": 1.0},
  "os_km_csv": "datasets/NCTXXXXXXXX/os_km.csv",
  "adverse_events": [
    {"term": "Anaemia", "organ_system": "Blood and lymphatic system disorders", "probability": 0.60, "is_serious": true},
    {"term": "Neutropenia", "organ_system": "Blood and lymphatic system disorders", "probability": 0.58, "is_serious": true},
    {"term": "Alopecia", "organ_system": "Skin and subcutaneous tissue disorders", "probability": 0.34, "is_serious": false}
  ]
}
```

`ecog_distribution` keys must be strings. Risk modifier fields are optional -- omit to use defaults from `synthetic_patient_generator.py`. `is_serious` defaults to `false`.

`os_km_csv` is the ground-truth survival source for the arm -- the generator samples per-patient `(time, event)` directly from the matching `arm` rows in that file. Only fall back to the legacy parametric fields (`median_os_months`, `death_probability`, `max_followup_months`, `os_weibull_shape`) when no KM CSV is available for the arm, and source them from the publication.

---

## Quality Checklist

- [ ] CT.gov JSON parsed first; every field that JSON could supply was taken from JSON, not from a later source
- [ ] Publication / supplement used only to fill CT.gov gaps
- [ ] SAP touched only for `cycle_length_days`, `n_induction_cycles`, treatment phases, and class-effect AEs
- [ ] Separate config per arm with unique `arm_name` matching the KM CSV `arm` values exactly
- [ ] `os_km_csv` set for every arm that has a reconstructed OS KM CSV (no Weibull stand-in)
- [ ] No PFS / DOR / other endpoint CSVs wired into the config -- OS only
- [ ] CSV-derived QC median cross-checked against published OS median (discrepancy noted if material)
- [ ] Every `organ_system` matches canonical MedDRA SOC list
- [ ] No duplicate AE terms (CT.gov serious + other events combined per term)
- [ ] `induction_ae_fraction` set and justified
- [ ] Many unique AE terms (80+ for large trials)
- [ ] >=5 laboratory abnormality AE terms
- [ ] `cycle_length_days` and `n_induction_cycles` match SAP/protocol
- [ ] AE probabilities are arm-specific
- [ ] Risk modifier overrides (if any) justified by document evidence
- [ ] Valid JSON

**Begin.** Drain the ClinicalTrials.gov JSON first, then wire in the OS KM CSV, then fill demographic and AE gaps from the publication + supplement, then read the SAP only for cycle structure and class effects. Output a separate JSON config for each arm.

\end{minted}

\section{Limitations and Future Directions}
\label{app:limitations}

The remaining limitations of the current implementation span the adverse event, demographic, and survival components of the pipeline, plus two safeguards that are not yet in place.

The most quantitatively significant gap is \emph{adverse event burden underestimation}: synthetic cohorts systematically generate 4–6 fewer adverse events per patient compared to real IPD (Table \ref{tab:results}). Three factors contribute. First, the propensity ablation shows that increasing the lognormal propensity variance $\sigma_\eta$ monotonically improves burden KS, indicating that the default value ($\sigma_\eta = 0.4$) is conservative --- it was chosen for numerical stability in small cohorts, and the per-trial sweep in Table~\ref{tab:sigma-sweep} shows that raising it to $0.8$ recovers most of the gap at no cost to the survival or frequency channels. This is a recalibration, not a fix for the two structural causes that follow. Second, ClinicalTrials.gov reports patient-level cumulative incidences rather than per-cycle hazard rates, and the current Poisson rate formulation approximates but does not explicitly correct for this prevalence-to-rate distinction. Third, ClinicalTrials.gov truncates its non-serious AE results at a 5\% reporting threshold, so any adverse event occurring in fewer than 5\% of patients is absent from the JSON entirely; the pipeline partially recovers these low-frequency events from the primary publication and appendix via the source-priority hierarchy, but coverage depends on how exhaustively the source paper tabulates sub-threshold AEs, and the tail is systematically under-represented when it does not. A related—but less severe—issue affects the temporal structure of these events: deterministic cycle-based sampling produces visually identifiable periodic peaks in onset distributions (Figure \ref{fig:ae-onset}), even though mean onset error remains below 0.8 months on all trials, and even though visit-driven ascertainment makes some quantization of real onsets expected. Events are also sampled conditionally independent given patient archetype and propensity: the shared propensity $\eta_i$ induces general within-patient correlation, but term-specific co-occurrence (beyond a patient's overall toxicity level) is not represented, which understates clustering driven by shared organ-system stress. Fitting a copula or per-SOC frailty layer would address this but requires joint AE statistics that registry records do not publish.

Two safeguards are not yet implemented. The configuration is LLM-populated but not range-checked before sampling, so a contradictory field --- an induction fraction above one, a categorical distribution that does not sum to one --- would reach the sampler rather than being caught at the boundary; a deterministic validation pass that bounds fractions and probabilities, verifies normalization, and clamps or flags violations (including on values recovered from the publication and SAP fallback levels) is a straightforward addition and is planned. Relatedly, the sampler places no floor on cohort size: the marginal KM is preserved at any $n$ by construction, but conditional and subgroup structure is not estimable in very small arms, and we recommend a few dozen patients per arm as a practical minimum (Section~\ref{sec:survival}).

On the demographic side, race and region distributions achieve JSD = 1.0 on all trials due to non-overlapping label sets between ClinicalTrials.gov and sponsor IPD (e.g., ``White'' vs.\ ``Caucasian''). This artifact reflects taxonomy mismatch rather than sampling error and would be resolved by a label normalization layer applied during extraction.

Finally, median overall survival error ranges from 7\% to 19\% across trials. This error is inherited from upstream KM curve reconstruction rather than the sampling step: methods that incorporate human-in-the-loop annotation achieve lower reconstruction error by leveraging additional constraints from number-at-risk tables. The oracle-curve experiment of Section~\ref{sec:propagation} bounds how much of the end-to-end residual this accounts for: regenerating from a ground-truth curve rather than the extracted one lowers $\Delta_{\text{KM}}$ by only $0.006$--$0.016$, so extraction error is a real but minority contributor that does not amplify downstream, and the rest is finite-sample noise in the reference. KMGen prioritizes full automation at the cost of this residual error, providing a faster iteration cycle suitable for large-scale applications. Human review of the traced curve and of the extracted propensity fields is compatible with the pipeline and advisable in practice; we deliberately do not use it in this evaluation so that the reported numbers reflect the fully automated setting.

\paragraph{Future Directions}
The most immediate improvements map directly to the limitations above: raising $\sigma_\eta$ to $0.8$ (Table~\ref{tab:sigma-sweep}) and adding an explicit prevalence-to-rate correction would close the primary quantitative gap, and the configuration validation pass and onset-spreading fix described above are both small changes. A further question raised by the ablations is whether the two dispersion parameters, the rank-noise $\sigma$ and the propensity $\sigma_\eta$, should be \emph{learned} rather than fixed: both sweeps show clear optima for burden and onset, so per-trial fitting would likely improve fidelity. We do not pursue it here because fitting them against real IPD would breach the public-information-only setting that the evaluation depends on; fitting against expert-elicited priors, or across a corpus of trials with published toxicity summaries, is the leakage-free version of this idea. Beyond these targeted fixes, several broader directions warrant investigation given the limited scope of the current setting—for example, dynamic frailty updating based on early treatment response to improve late-phase AE prediction, a copula or hierarchical layer for term-specific AE co-occurrence, or extension to non-oncology indications with different visit-driven observation mechanisms.

\end{document}